%% file: main.tex
\documentclass[review]{elsarticle} 

\newcommand{\datacodeavailability}{%
    The \emph{Mangrove3D} dataset has been released on Zenodo (\doi{10.5281/zenodo.16933584}). The accompanying preprocessing, feature engineering, training, and evaluation code (including environment specification and reproducibility instructions) is available at \url{https://fz-rit.github.io/through-the-lidars-eye/}. %
}

\usepackage[T1]{fontenc}
\usepackage[utf8]{inputenc}
\usepackage{lmodern}

\usepackage[margin=1in]{geometry}
\usepackage{enumitem}
\usepackage{doi}
\usepackage{amsmath,amssymb}
\usepackage{graphicx}
\graphicspath{{./figures/}}
\usepackage{subcaption}
\usepackage{tikz}
\usetikzlibrary{shapes,positioning,fit,backgrounds,arrows.meta,calc}
\usepackage[section]{placeins}  

\usepackage{array}
\renewcommand{\arraystretch}{1.2}
\usepackage{booktabs}
\usepackage{multirow}
\usepackage{makecell}
\usepackage[table]{xcolor}
\usepackage{threeparttable}
\usepackage{colortbl}

\usepackage{microtype}
\usepackage{ragged2e}
\usepackage{csquotes}
\usepackage{url}  
\usepackage{tabularx}
\newcolumntype{Y}{>{\raggedright\arraybackslash}X}

\usepackage{siunitx}
\usepackage{lineno}
\modulolinenumbers[5]

\definecolor{rankOneBG}{HTML}{C6DBEF}
\definecolor{rankTwoBG}{HTML}{DEEBF7}
\definecolor{rankThreeBG}{HTML}{F1F7FC}
\definecolor{matchCyan}{HTML}{00BFFF}
\definecolor{mismatchPurp}{HTML}{C44DFF}

\newcommand{\first}[1]{\cellcolor{rankOneBG}\textbf{#1}}
\newcommand{\second}[1]{\cellcolor{rankTwoBG}\textbf{#1}}
\newcommand{\third}[1]{\cellcolor{rankThreeBG}\textbf{#1}}

\newcommand{\matchCyan}[1]{\textbf{\textcolor{matchCyan}{#1}}}
\newcommand{\mismatchPurp}[1]{\textbf{\textcolor{mismatchPurp}{#1}}}

\newcolumntype{C}[1]{>{\Centering\arraybackslash}m{#1}}
\newcolumntype{R}[1]{>{\RaggedRight\arraybackslash}m{#1}}
\newcolumntype{P}[1]{>{\raggedright\arraybackslash}p{#1}}
\newcolumntype{L}[1]{>{\RaggedRight\arraybackslash}p{#1}}

\usepackage{natbib}
\setcitestyle{numbers,square,comma,sort&compress}

\journal{ISPRS Journal of Photogrammetry and Remote Sensing}
\begin{document}

\begin{frontmatter}

    \title{Through the Perspective of LiDAR: A Feature-Enriched and Uncertainty-Aware Annotation Pipeline for Terrestrial Point Cloud Segmentation}

    \author[rit]{Fei Zhang\corref{cor1}}
    \ead{fzhcis@rit.edu}

    \author[rit]{Rob Chancia}
    \author[rit]{Josie Clapp}
    \author[rit]{Amirhossein Hassanzadeh}
    \author[rit]{Dimah Dera}
    \author[usfs]{Richard MacKenzie}
    \author[rit]{Jan van Aardt}

    \address[rit]{Chester F. Carlson Center for Imaging Science, Rochester Institute of Technology, Rochester, NY, USA}
    \address[usfs]{U.S. Forest Service, USA}

    \cortext[cor1]{Corresponding author.}


    \begin{abstract}
        Accurate semantic segmentation of terrestrial laser scanning (TLS) point clouds is limited by costly manual annotation. We propose a semi-automated, uncertainty-aware pipeline that integrates spherical projection, feature enrichment, ensemble learning, and targeted annotation to reduce labeling effort, while sustaining high accuracy. Our approach projects 3D points to a 2D spherical grid, enriches pixels with multi-source features, and trains an ensemble of segmentation networks to produce pseudo-labels and uncertainty maps, the latter guiding annotation of ambiguous regions. The 2D outputs are back-projected to 3D, yielding densely annotated point clouds supported by a three-tier visualization suite (2D feature maps, 3D colorized point clouds, and compact virtual spheres) for rapid triage and reviewer guidance.
        Using this pipeline, we build \textit{Mangrove3D}, a semantic segmentation TLS dataset for mangrove forests. We further evaluate data efficiency and feature importance to address two key questions: (1) how much annotated data are needed and (2) which features matter most. Results show that performance saturates after $\sim$12 annotated scans, geometric features contribute the most, and compact nine-channel stacks capture nearly all discriminative power, with the mean Intersection over Union (mIoU) plateauing at around 0.76. Finally, we confirm the generalization of our feature-enrichment strategy through cross-dataset tests on \textit{ForestSemantic} and \textit{Semantic3D}.
        Our contributions include: (i) a robust, uncertainty-aware TLS annotation pipeline with visualization tools; (ii) the \textit{Mangrove3D} dataset; and (iii) empirical guidance on data efficiency and feature importance, thus enabling scalable, high-quality segmentation of TLS point clouds for ecological monitoring and beyond. The dataset and processing scripts are publicly available at \url{https://fz-rit.github.io/through-the-lidars-eye/}.
    \end{abstract}

    \begin{keyword}
        Terrestrial laser scanning (TLS) \sep Semantic Segmentation \sep Uncertainty analysis \sep Spherical Projection \sep LiDAR Point Cloud \sep Forest Ecology
    \end{keyword}

\end{frontmatter}


\section{Introduction}
LiDAR (light detection and ranging) systems are deployed across platforms ranging from satellites to unmanned aerial systems (UAS), mobile units, and terrestrial scanners. Among these, terrestrial laser scanning (TLS) provides exceptionally dense 3D point clouds (mm-cm scale), enabling detailed structural analysis in ecological environments~\cite{calders2020terrestrial,disney2018weighing}. In forest ecosystems, TLS supports the retrieval of tree metrics, biomass, and habitat features~\cite{liang2016terrestrial,penman2023instructional}, and these applications fundamentally rely on accurate semantic segmentation to distinguish ground, stems, branches, foliage, and roots within complex scenes.

While deep learning architectures such as PointNet++ and KPConv have surpassed traditional classifiers (e.g., Random Forests, SVMs) in TLS segmentation~\cite{qi2017pointnet++,thomas2019kpconv}, their broader utility remains limited by two persistent bottlenecks. First, high-quality annotated datasets are scarce: manual labeling of full-resolution TLS scans is prohibitively labor-intensive, and ecological scenes are especially problematic due to severe occlusion, irregular geometry, and intertwined tree structures~\cite{kulicki2024artificial,bornand2024completing}. Second, the few open-source datasets that do exist are biased toward urban or indoor environments, where objects are simpler and more regularly shaped~\cite{hackel2017semantic3d,walczak2024inlut3d,dai2017scannet,merizette20253dses}. Consequently, current segmentation models often fail to generalize to ecologically complex scenes, slowing the adoption of automated analysis, despite urgent demand in forestry and environmental monitoring.

To overcome these obstacles, we introduce a human-in-the-loop annotation pipeline that reduces labeling burden, while improving segmentation fidelity in challenging forest environments. Our pipeline (i) leverages spherical projection to transform irregular 3D data into structured 2D maps, enabling efficient annotation and feature extraction in a lower-dimensional space; (ii) integrates feature-enriched segmentation to better capture radiometric, geometric, and statistical information; and (iii) incorporates uncertainty analysis to target regions of low model confidence for human correction. Together, these components accelerate annotation, enhance model robustness, and bridge the gap between small, high-effort ecological datasets and scalable, reproducible benchmarks. To this end, we release \textit{Mangrove3D}—to our knowledge the first TLS benchmark explicitly designed for structurally complex mangrove forests—and demonstrate the pipeline’s cross-domain effectiveness through evaluations on \textit{Mangrove3D}, \textit{ForestSemantic}\cite{liang2025forestsemantic,mspace_lab_2025_15193973}, and \textit{Semantic3D}\cite{hackel2017semantic3d}.

In summary, this paper aims to:
(a) alleviate the annotation bottleneck through a semi-automated pipeline,
(b) introduce the TLS dataset tailored to mangrove forests, and
(c) evaluate the effectiveness and cross-domain robustness of feature-enriched segmentation.

\subsection{Related Work}
\subsubsection{Manual Annotation Software and Methodologies}
\label{sec:manual-annotation-tools}
\input{tex_of_fig_tab/tab_annotation_tool_compare}
Semantic annotation of TLS point clouds remains a major bottleneck in ecological research. Unlike urban or autonomous driving scenes, TLS data contain dense occlusions, irregular branching, and intertwined roots, making manual labeling slow and error-prone. Existing 3D annotation tools (Table~\ref{tab:annotation-tools-comparison})—such as \textit{CloudCompare}, \textit{3D Slicer}, \textit{MathWorks LiDAR Labeler}, and \textit{Segments.ai}—offer functions like region clipping and polygonal selection, with some integrating deep neural networks (DNNs) for pre-annotation. However, most were developed for robotics or urban datasets and remain labor-intensive, requiring annotators to switch viewpoints, navigate dense 3D scenes, and often cross-reference RGB imagery that is rarely available in TLS.

To overcome these limitations, we propose a new annotation pipeline emphasizing efficiency, uncertainty-awareness, and adaptability for complex TLS datasets.

\subsubsection{Spherical (Equirectangular) Projection for TLS Point Clouds}
\label{sec:spherical-proj}
The concept of spherical projection originates from the equirectangular (plate carrée) projection, which linearly maps longitude and latitude to planar $x$-$y$ coordinates. First used in cartography as early as AD 400~\cite{snyder1997flattening}, this simple yet powerful mapping preserves angular relationships along parallels and meridians. Centuries later, the same principle finds renewed application in LiDAR point‐cloud processing, where unstructured 3D measurements are transformed into structured 2D ``range images.'' Each point’s azimuth and elevation (or zenith) are linearly mapped to pixel indices, creating a dense angular grid commonly referred to as a spherical projection~\cite{mao20233d}. In essence, this projection provides a way of viewing the world \textit{through the perspective of LiDAR}—capturing how the sensor perceives its surroundings across its field of view. Scalar attributes such as range, intensity, and return count are then stored as per‐pixel channels, providing a compact 2D representation that preserves the geometric structure of the original 3D scene.

Early works, such as Barnea et al.~\cite{barnea2008segmentation,BARNEA201333} and Mahmoudabadi et al.~\cite{MAHMOUDABADI2016135}, leveraged spherical intensity and range maps in combination with classical clustering methods such as mean-shift and graph-based segmentation. These approaches require manual parameter tuning, process each feature channel independently, and are constrained by the limited capacity of traditional machine learning algorithms to integrate heterogeneous features. More recently, DNN-based models such as RangeNet++\cite{milioto2019rangenet++}, SalsaNext\cite{cortinhal2020salsanext}, and SqueezeSeg\cite{wu2018squeezeseg} demonstrated the effectiveness of spherical projection for semantic segmentation, particularly in autonomous driving datasets. However, these approaches were primarily designed for automotive LiDAR, which typically has a narrow vertical field of view and limited feature inputs (often a single channel such as range or intensity), making them strongly dependent on synchronized multi-sensor data.

In this study, we adapt spherical projection and DNN to TLS, which is widely used in ecological, indoor, and construction environments where camera imagery is often unavailable or unreliable. We extend the traditional single-band representation to multi-channel feature stacks that generalize pseudo-color encodings on spherical maps, serving as an intuitive aid for both human annotators and DNN-based segmentation models.

\subsubsection{Active Learning and Self-Training for Efficient Annotation}
To further alleviate the burden of pixel-level semantic labeling on the spherical maps, we integrate two human-efficient annotation paradigms: active learning and self-training.

Active learning (AL) iteratively queries the annotator for the most informative samples, often those with the highest uncertainty, approximated via Monte-Carlo dropout~\cite{gal2016dropout} or ensembles~\cite{lakshminarayanan2017simple}. While well-studied for image classification, dense segmentation is more difficult, since annotation units (pixel, tile, or superpoint) must balance annotation cost against information gain. Recent studies~\cite{sener2018active,shao2022active,xu2023hierarchical} adapt AL to 2D and 3D domains, but mainly on RGB-rich indoor datasets, and still relies on slow, expertise-intensive 3D point editing.

Self-training is a semi-supervised strategy where a model trained on a seed set reuses its own predictions to expand the labeled pool. Typically, high-confidence predictions are promoted to pseudo-labels and reintroduced for training, sometimes with consistency regularization or noise injection to mitigate overfitting~\cite{xie2020noisy,zou2020pseudoseg}. This approach has shown success in natural images and range-image LiDAR~\cite{wu2019squeezesegv2}, but faces well-known limitations, e.g., pseudo-label errors easily accumulate, confidence does not always imply correctness, and most pipelines omit human oversight, allowing mistakes to cascade unchecked~\cite{zou2019confidence,el2025distribution}. These challenges are amplified on spherical projection maps of TLS scans, where clutter, occlusion, and class imbalance produce misleading confidence estimates.

We therefore propose an efficient, semi-automatic annotation framework that integrates active learning and self-training within the spherical projection domain to address these challenges. We use an ensemble of UNet++~\cite{zhou2018unet++}, DeepLabV3+~\cite{chen2018encoder}, and Segformer~\cite{xie2021segformer} to quantify pixel-wise epistemic uncertainty, directing annotators to the most informative regions (AL), while high-confidence pixels are automatically promoted to pseudo-labels (self-training). This integrated, uncertainty-aware pipeline reduces manual effort while still ensuring human verification, and— to our knowledge—represents the first scalable framework demonstrated on full-resolution TLS spherical projection maps.

\subsubsection{Feature fusion in semantic segmentation of point clouds}
\label{subsec:feature-fusion}

A closely related approach is PointPainting and its derivatives~\cite{vora2020pointpainting,dong2023pep}, which project point clouds onto segmented RGB images, attach semantic scores to points, and feed the enriched clouds into downstream networks. While effective for autonomous driving, these approaches depend on accurately registered multi-sensor data. In contrast, TLS generally operates without co-registered imagery, meaning that spectral cues must be inferred from intrinsic LiDAR signals such as intensity, range, and derived geometric attributes. Our framework is therefore designed to operate solely on LiDAR-derived features, eliminating dependence on external imagery while still capturing complementary structural and radiometric information.

Within LiDAR-only pipelines, feature fusion exploits complementary cues—geometric structure, radiometric response, and positional context—to enhance segmentation. Prior studies have shown that integrating geometric descriptors with intensity and elevation improves class separability~\cite{chehata2009airborne,rs5083749}, and benchmarks such as Semantic3D confirm gains from combining coordinates, intensity, and shape features~\cite{hackel2017semantic3d}. Accordingly, both projection-based and fully 3D networks (e.g., SqueezeSeg, PointNet++, RandLA-Net) often employ stacked multi-cue inputs~\cite{qi2017pointnet++,wu2018squeezeseg,hu2020randla}. Yet, despite broad agreement that multi-feature fusion helps, there is still no consensus on an optimal LiDAR-only feature set; redundant or correlated inputs can inflate dimensionality without consistent performance gains~\cite{weinmann2015semantic}.

To address this gap, we systematically expand and evaluate feature channels on TLS spherical projection maps—starting from single feature families and progressively integrating others—to quantify their complementarity and identify combinations that yield the greatest marginal improvements. This progressive analysis results in a compact yet robust feature stack that reduces redundancy, enhances segmentation efficiency, and provides practical guidance for TLS applications, where feature selection has traditionally been ad hoc.

\section{Methods and Materials}
In this section we introduce a new TLS dataset curated with our pipeline, detail the pipeline design, and describe the qualitative and quantitative approaches used to assess feature enrichment and data efficiency.
\subsection{A New TLS Dataset for Mangrove Forest}
\label{sec:mangrove3d-dataset}
We introduce the \emph{Mangrove3D} dataset. The data were collected in spring 2024 on
Babeldaob Island, Palau (\(7^{\circ}\,31' \,49''\mathrm{N},\,134^{\circ}\,33' \,53''\mathrm{E}\)), in coastal areas of Rhizophora mangrove characterized by dense prop-root networks
and multilayered canopies.

\input{tex_of_fig_tab/fig_Palau_mangrove_and_CBL_scanner}
\input{tex_of_fig_tab/tab_mangrove3d_noise.tex}
\input{tex_of_fig_tab/fig_mangrove3d_density_vs_range_hight.tex}
We use the Canopy Biomass LiDAR (Version 2.0; CBL), a TLS built around a SICK LMS-151 LiDAR unit (SICK AG, Waldkirch, Germany)~\cite{SICK_LMS151},
which employs a 905~nm wavelength laser and has an effective measurement range of 0.5--50~m at 90\% reflectivity. The LiDAR is attached to a rotation stage and the
CBL mounted on a modified surface elevation table (SET) arm---a portable leveling device attached to a benchmark pipe, as shown in
Fig.~\ref{fig:palau_rset}. For most scans, we invert the CBL from the standard tripod orientation to obtain the maximum potential ground points
surrounding the SET, orienting the \(90^{\circ}\) uncovered cone above the scanner position, thus enabling \(360^{\circ}\times270^{\circ}\) scans while
minimizing occlusion from below. Fig.~\ref{fig:mangrove3d_density} provides a compact validation of point density characteristics across all scans. 
Density decreases consistently with range across scans, while the vertical distributions confirm a strong near-ground concentration of points, consistent with the root-focused acquisition strategy.
To assess geometric noise characteristics, Table~\ref{tab:mangrove_noise_vs_class} summarizes PCA-based local surface roughness statistics of the Mangrove3D dataset across major semantic regions. Lower roughness is observed for ground and root surfaces, while higher variability is present in stem and canopy regions, consistent with reduced point density at larger ranges.

Each CBL scan is acquired at an angular resolution of \(0.25^{\circ}\) and completed in 33s. We collect eight scans at each of the benchmark sites, starting with the first scan position
with the SET-arm oriented directly north of the SET and the preceding scans at successive \(45^{\circ}\) intervals
rotating in a clockwise direction. Scans that suffer from obvious obstruction or operator error are eliminated to ensure data quality.

\input{tex_of_fig_tab/fig_mangrove3d_class_distribution}
The dataset comprises 39 TLS scans from seven mangrove plots, totaling 31.3 million points. 
Individual scans range from 0.51M to 0.94M points (averagely 0.80M per scan), depending on site conditions and preprocessing. Each point is assigned to one of five classes—\textit{Ground \& Water}, \textit{Stem}, \textit{Canopy}, \textit{Root}, and \textit{Object}—with a sixth label, \textit{Void}, marking empty pixels in the spherical projection image cube. 
For segmentation tasks, 30 scans from plots \#1-5 are used for training and validation, while 9 scans from plots \#6-7 are held out as a fixed test set. In our main experiment, we adopt a train/validation/test split of 27/3/9, yielding a validation-to-train ratio of approximately 0.1. The class distributions across these splits are shown in Fig.~\ref{fig:mangrove3d_class_dist}. A brief annotation guideline is provided in the~\ref{app:mangrove3d-dataset}.
The \emph{Mangrove3D} dataset is, to our knowledge, the first TLS benchmark for semantic segmentation in mangrove forests, providing centimeter-scale 3D geometry of tidally influenced ecosystems with dense root and canopy structures. The resulting significant occlusion poses a challenging benchmark for semantic segmentation, a key prerequisite for downstream applications such as biomass estimation and blue-carbon assessments.

\subsection{Semi-automated Annotation Pipeline}
\label{sec:annotation-pipeline}
\input{tex_of_fig_tab/fig_tls_2d3d_seg_pipeline}
As illustrated in Fig.~\ref{fig:tls_2d3d_seg_pipeline_E}, we design a three-stage pipeline to facilitate efficient and consistent annotation of TLS point clouds, detailed in the following subsections.

\subsubsection{Stage 1: Spherical Projection \& Visualization}\label{sec:stage1}
\input{tex_of_fig_tab/fig_spherical_projection}
\input{tex_of_fig_tab/fig_CBL_density_map}
We unwrap the TLS point cloud into a 2D spherical domain parameterized by zenith ($\theta$) and azimuth ($\phi$), yielding multi-channel spherical projection maps where each channel encodes aforementioned various features. Fig.~\ref{fig:spherical_projection} illustrates the process of spherical projection on a CBL LiDAR scan, point cloud and feature map were colorized from raw intensity values. For a 3D point $(x,y,z)$,
\[
    \theta = \arccos \left(\tfrac{z}{\sqrt{x^2+y^2+z^2}}\right), \quad
    \phi = \mathrm{mod}\left(\arctan2(y,x), 2\pi \right),
\]
which are mapped to pixel coordinates via the equirectangular projection
\[
    i=\Bigl\lfloor \tfrac{\theta-\theta_{\min}}{\Delta \theta}\Bigr\rfloor, \quad
    j=\Bigl\lfloor \tfrac{\phi-\phi_{\min}}{\Delta \phi}\Bigr\rfloor.
\]
For each CBL LiDAR scan, with an angular step of $0.25^\circ$ and a vertical field-of-view (FOV) = $135^\circ$ and a horizontal FoV = $360^\circ$, we obtain a $540\times1440$ spherical grid, providing a near one-to-one mapping between beam directions and pixels, as validated in Fig.~\ref{fig:CBL_density_map}.

This projection provides a stable canvas for stacking per-pixel attributes and descriptors. We organize the features into three groups: 1) basic properties: radiometric intensity, range, and inverted height; 2) geometric properties: normals, curvature, anisotropy, and planarity; and 3) statistical properties: low-dimensional features obtained via PCA, with MNF and ICA considered as alternatives. Local 3D structure is quantified using eigenvalue-based descriptors derived from the covariance of neighboring points. Specifically, curvature, anisotropy, and planarity are computed from the ordered eigenvalues, providing clear visual signatures and demonstrating segmentation utility.

\paragraph{Optimizations}
Two challenges are scale and density variation. Nearly one million points per scan can lead to excessive memory requirements, and scanner geometry leads to uneven point densities. We partition clouds into azimuth-elevation tiles processed in batches, padded with $k_\text{b}$ neighbors to mitigate edge artifacts. For density variation, we use an adaptive neighborhood radius $r_i=\text{clamp}(\lambda d_{(k)}, r_\text{min}, r_\text{max})$ with $d_{(k)}$ is the $k$-th nearest neighbor distance, $\lambda$ is a scaling factor (e.g., $\lambda = 1.5$), and $[r_\text{min},r_\text{max}]=[0.02,0.3]$ m. Together, these strategies enable a scalable, consistent feature computation.

Detailed preprocessing steps, additional feature map examples, correlation analyses, and alternative dimensionality-reduction methods are provided in \ref{app:stage1} (Table~\ref{tab:preprocessing_feat_map}, Figs.~\ref{fig:feature_maps_single_mangrove3d}-\ref{fig:feature_maps_stack_mangrove3d}).

\subsubsection{Stage 2: Hybrid Annotation with Semi-Supervised and Active Learning}
\input{tex_of_fig_tab/fig_pseudolabel-uncertainty}

In Stage~2, we adopt a human-in-the-loop strategy that combines self-training and active learning to reduce manual labeling effort, as shown in Fig.~\ref{fig:mangrove3d-pseudolabel-uncertainty}.

First, a small subset of spherical projection image cubes—one per TLS scan—is manually annotated using Adobe Photoshop (Adobe Inc., 2024)~\cite{adobe2024}. We use these seed labels for training an ensemble of three diverse 2D semantic segmentation models: UNet++, DeepLabV3+, and Segformer. Each model adopts a unique encoder backbone—i.e., ResNet‑34 \cite{he2016deep}, EfficientNet‑B3 \cite{tan2019efficientnet}, and MiT‑B1 \cite{yu2023mix}, respectively—to promote representational diversity and diminish correlated errors.
UNet++ is well-suited for boundary recovery~\cite{zhou2018unet++}, DeepLabV3+ captures multi-scale context through atrous spatial pyramid pooling~\cite{chen2018encoder}, and SegFormer leverages hierarchical Transformers for long-range dependencies~\cite{xie2021segformer}.

Each image is partitioned into five vertical tiles with buffer zones (paddings) to mitigate boundary artifacts in order to augment the training set and expose the model to a broader range of local contexts. Loss and evaluation are confined to the unbuffered cores, and tile dimensions are padded to multiples of 32 for architectural compatibility. This strategy ensures tractable training while preserving spatial continuity.

\input{tex_of_fig_tab/fig_ensemble_multiencoder}
\input{tex_of_fig_tab/tab_uncertainty_guided_workflow}
Because the input feature maps extend beyond standard RGB, we re-architect each backbone into a multi-encoder fusion design (Fig.~\ref{subfig:multiencoder-feature-fusion}). Each 3-channel feature group (e.g., intensity, range, Z-inv) is processed by a dedicated encoder (ResNet-34, EfficientNet-B3, or MiT-B1). These encoders are initialized with ImageNet pre-trained weights, ensuring transfer of learned low-level filters while allowing adaptation to non-RGB feature groups. Deep features are concatenated at the bottleneck and aligned by interpolation before decoding, a strategy shown to outperform naive early fusion~\cite{rs16203852, Lei9246289, ZHANG2020183}. This modular design accommodates any $3\times N$ channel configuration with minimal retraining.

The ensemble combines predictions to generate segmentation masks and epistemic uncertainty maps: high-uncertainty regions are refined manually (active learning), while high-confidence predictions are retained as pseudo-labels (self-training). Fusing logits allows the networks to complement one another~\cite{dietterich2000ensemble}, while deep ensembles provide more reliable uncertainty estimates than a Monte-Carlo dropout approach ~\cite{lakshminarayanan2017simple,kendall2017uncertainties}. The inference-refinement cycle is repeated until all scans are fully annotated. In practice, we display the pseudo-label masks, the uncertainty maps, and the input feature maps to guide annotators. As summarized in Table~\ref{tab:annotation_workflow_comparison},, this joint view accelerates review by steering attention to the genuinely ambiguous regions, while leaving well-segmented areas largely untouched. Technical details of the Dice-CrossEntropy joint loss~\cite{milletari2016v,sudre2017generalised,mao2023cross} and the ensemble-based epistemic uncertainty formulation are provided in ~\ref{app:stage2} (Eqs.~\ref{eq:loss_combined}-\ref{eq:uncertainty_epistemic}).

\subsubsection{Stage 3: Back Projection and Refinement in 3D Space}
\input{tex_of_fig_tab/fig_mangrove3d_back-projection}

\input{tex_of_fig_tab/fig_colorized_pcd_mangrove3d}
Stage 3 connects the 2D annotation workflow back to the 3D domain (Fig.~\ref{fig:back-projection-illus}). While spherical projection enables efficient processing, it may introduce localized ambiguities when multiple points map to the same pixel, particularly along object boundaries such as stem-canopy transitions. To compensate for this loss of explicit 3D neighborhood context, we apply a two-step refinement in 3D space.
First, geometric smoothing is performed using k-nearest neighbor (kNN) majority voting, which suppresses small, isolated label errors by enforcing local spatial consistency. Second, a Random Forest classifier trained on reliable core regions is used to correct systematic boundary ambiguities using geometric and radiometric features. Because both stages operate directly in 3D space, they re-embed each point within its local neighborhood and mitigate projection-induced inconsistencies.
Together, these steps yield crisp and reliable 3D semantic annotations, which we further verify through manual inspection.
Fig.~\ref{fig:colorized_pcd} illustrates examples of colorized 3D renderings from different feature groups, enabling visual inspection and cross-validation.
\input{tex_of_fig_tab/fig_mangrove3d_balls}
\paragraph{Compact Virtual Spheres}
While back-projected TLS point clouds provide detailed annotations, they remain cumbersome to inspect: large file sizes slow interaction, non-uniform angular sampling introduces distortions, and simple navigation tasks such as zooming or panning can be unintuitive. We thus introduce \emph{virtual spheres} to address these challenges: synthetically-defined spherical grids with preset and tunable angular resolution and radius, onto which 2D feature maps are re-projected. These constructs are compact, density-neutral, and resolution-controllable, while still preserving the global structure of each scan. In practice, they serve as lightweight three-dimensional ``thumbnails'' that enable rapid inspection of coverage, spatial proportions, and segmentation quality, without the overhead of full-resolution point clouds (Fig.~\ref{fig:mangrove_3d_balls}).

Technical details of the back-projection equations, kNN voting, and Random-Forest relabeling are provided in~\ref{app:stage3}(Eqs.~\ref{eq:backproj_simple}-\ref{eq:rf_update}), along with additional information and applications of the virtual spheres.

\subsection{Performance Evaluation and Analysis}
\label{subsec:perform_eval}
\paragraph{Evaluation metrics}
We evaluate each feature set and its epistemic-uncertainty map with the metrics in Table \ref{tab:eval-metrics}.
Segmentation quality is captured by overall accuracy (oAcc), mean class accuracy (mAcc), mean IoU (mIoU) and per-class IoU (IoU\(_c\)).
Pixel-wise errors and their uncertainty are quantified with \emph{Shannon Entropy}, while the uncertainty map’s ability to expose errors is measured by the \emph{area under the precision-recall curve (AUPRC)} between the uncertainty map and a binary error mask.
\input{tex_of_fig_tab/tab_eval_metrics}

\paragraph{Marginal Gains from Expanded Feature Sets}
Each handcrafted feature map highlights a specific geometric or radiometric property of the point cloud. These maps clearly enhance human interpretation by providing additional visual cues. However, it is less clear whether modern DNNs gain similar benefits from such explicit descriptors, or whether they can instead learn equivalent representations directly from simpler inputs. Furthermore, adding extra feature channels unavoidably raises memory requirements and computational costs on both CPU and GPU during training and inference.

We therefore compare segmentation performance in terms of accuracy and uncertainty across progressively enriched input configurations in order to quantify this accuracy-versus-cost trade-off and to assess the necessity of explicit feature engineering:
\begin{enumerate}[label=\alph*)]
    \item Basic 3-channel sets: raw \(\{\text{intensity},\text{range},z\}\) and their preprocessed  counterparts;
    \item Additional 3-channel sets: (i) geometric descriptors - curvature, anisotropy, planarity,
          (ii) pseudo-RGB from surface normals,
          and (iii) first three components of PCA, MNF, or ICA applied to the nine-channel set (combined from preprocessed basic set, i, and ii);
    \item Six-channel combinations: the basic set concatenated with any one additional trio, or two additional trios merged;
    \item Nine- and twelve-channel stacks: the core set joined with both normals and geometric descriptors (9 ch); and the same stack further augmented by the PCA trio (12 ch).
\end{enumerate}

\paragraph{Impact of Annotated Sample Size}
To assess the data efficiency of our annotation pipeline, we perform experiments with progressively larger subsets of annotated training data. Specifically, we randomly sample training sets of size \(n=\{4,\,8,\,12,\,16,\,20,\,24,\,28\}\) scans. For each subset size, we maintain a fixed validation ratio of 0.25, ensuring integral train/validation set splits (e.g., 3/1 for \(n=4\); 6/2 for \(n=8\), etc.). The test set remains 9 scans from plots \#6-7 across all experiments.

\section{Results}

\subsection{2D Segmentation Result}
\label{subsec:mangrove3d-seg2d-result}
\input{tex_of_fig_tab/fig_mangrove3d-eval-results-ch345}
Fig.~\ref{fig:mangrove3d-eval-results-ch345} shows that segmentation errors are primarily concentrated in cluttered transition regions, such as canopy--stem interfaces and root--ground boundaries. These error-prone areas are consistently highlighted by elevated epistemic uncertainty, with bright uncertainty responses largely overlapping misclassified pixels and yielding AUPRC values of 0.30--0.40 across scans. Notably, the highest-uncertainty pixels achieve near-unity precision at low recall, indicating that a small fraction of uncertain regions captures a large proportion of true errors, which is well suited for targeted human inspection in uncertainty-guided annotation workflows.
\subsection{Effect of feature enrichment}
\label{sec:feature-ablation-results}
\input{tex_of_fig_tab/fig_feature_ablation}
Fig.~\ref{fig:feat-ablation-mangrove3d} summarizes segmentation performance across 18 combinations of gradually expanded feature groups. Overall accuracy ranges between 0.80 - 0.89, mean accuracy between 0.78 - 0.87, and mean IoU between 0.68 - 0.76.

Across single-channel inputs—raw intensity, range, Z, their preprocessed versions, and per-point geometry (curvature, anisotropy, planarity)—mean IoU ranges between 0.702 - 0.731. Most three-channel combinations perform similarly, but the contrast-enhanced \textit{I.R.Z} stack (intensity, range, inverse Z) lifts mIoU to 0.745, confirming the value of the histogram stretch in preprocessing. Merging feature groups into a six-channel input pushes performance further (0.754-0.761), yet adding still more channels yields no real gain: results level off at \(\approx0.76\). The best configuration, \textit{IRZ\_N3\_CAP}, tops out at 0.768—three points better than raw intensity alone and six points better than raw range. In practice, therefore, a compact six-channel feature set captures nearly all available discriminative power on the \emph{Mangrove3D} dataset, whereas larger stacks add computation complexity with little return.

\input{tex_of_fig_tab/tab_model_complexity.tex}

Table~\ref{tab:model_comparison} reports model complexity, computational cost, and segmentation accuracy across feature groups. 
All backbone models were trained independently, and inference was performed at the level of individual TLS scans. 
Inference times range from 26--64~ms for single models and remain below 150~ms per scan for the ensemble, indicating that the additional cost of ensembling is modest in practical terms. 
While individual architectures exhibit different efficiency--accuracy tradeoffs, the ensemble provides consistently higher accuracy across feature groups, making it well suited for generating reliable pseudo-labels in offline processing pipelines.

\paragraph{Benchmarking with PointNet++}
\input{tex_of_fig_tab/tab_pointnet2_benchmark}
We benchmark PointNet++ on the \emph{Mangrove3D} dataset using 30 scans for training and nine scans for testing, following the split protocol in Section~\ref{sec:mangrove3d-dataset}. Models are trained for 40 epochs with the Adam optimizer (initial learning rate 0.01, cosine-annealing schedule), block size 2 m, batch size 32, and 4096 points per block; the checkpoint with the highest validation mIoU is retained.

As shown in Table~\ref{tab:pointnet2_benchmark}, the XYZ-only baseline achieves 0.634 mIoU, reflecting common \textit{Stem}-\textit{Canopy} confusion. Adding geometric descriptors—surface normals and \textit{C.A.P} (six additional channels)—boosts performance to 0.712 mIoU, the best configuration observed. Class-level trends mirror this pattern: \textit{Objects} show the largest gain (0.516 to 0.837), while \textit{Ground} is already strong and improves only marginally. Beyond this six-extra-channel setup, additional features do not yield further benefits, indicating feature saturation once orientation and local-shape cues are included.

\subsection{Data efficiency}
\label{sec:data-efficiency-ensemble}
\input{tex_of_fig_tab/fig_data_efficiency_accu}

\input{tex_of_fig_tab/fig_data_efficiency_uncert}
Fig.~\ref{fig:data_efficienty_accu} shows how training sample size influences ensemble model performance under four representative feature configurations. Accuracy improves steadily with additional samples, with performance largely saturating around 12 scans.
Fig.~\ref{fig:data_efficienty_uncert} depicts the corresponding changes in entropy and AUPRC. As training size increases, all feature configurations exhibit consistent reductions in the entropies of total uncertainty, epistemic uncertainty (as measured by mutual information), and error maps—indicating greater model confidence and stability. Higher-dimensional inputs (6-Ch and 12-Ch) maintain lower entropy and stronger alignment between uncertainty estimates and true errors, particularly once trained with 12 or more scans.

With fewer than 12 samples, models display elevated entropy and more variable AUPRC values. Beyond this threshold, AUPRC values converge across configurations, suggesting that all feature groups achieve comparably stable alignment between uncertainty and prediction error.

\subsection{Generalization to Open-Source TLS Datasets}
We apply our feature-enriched, uncertainty-aware pipeline to two public datasets to assess transferability beyond mangroves: \emph{ForestSemantic} (boreal forest scenes) and \emph{Semantic3D} (urban scenes).

\subsubsection{ForestSemantic}
The \emph{ForestSemantic} dataset comprises \(\sim\!720\) million points sampled from six nominal 32 m \(\times\) 32 m plots in Evo, Finland (61.19° N, 25.11° E).
Each plot is labeled into six classes: Ground, Trunk, First-order branch, Higher-order branch, Foliage, and Miscellany.
At the time of writing (Sept 2025) only three plots—\#1, \#3, and \#5—were publicly available on Zenodo~\cite{mspace_lab_2025_15193973}; our experiments therefore focused on these.

Visual inspection of the ground surface reveals five circular voids per plot, indicating that each plot is a registration of five individual scans. To prepare the dataset for our pipeline, we first detect the centroids of these voids and treat them as \emph{pseudo TLS positions}. We then randomly subsample the point cloud at 5\% density (e.g., from 20 million per plot to 1 million points per plot) for computational efficiency, then recenter each point by subtracting the nearest pseudo-scanner center in \((x,y)\) and the global mean \(z\). This yields five re-centered pseudo scans per plot and fifteen in total.

\paragraph{Qualitative Transfer}
\input{tex_of_fig_tab/fig_forestsemantic_2d}
\input{tex_of_fig_tab/fig_forestsemantic_pcd}
\input{tex_of_fig_tab/fig_forestsemantic_balls}
Figure~\ref{fig:forestsemantic_2d_visuals} shows different feature channels and trio stacks from the spherical projection of the preprocessed ForestSemantic dataset.
Each composite highlights complementary structural cues—from ground texture and mid-story complexity to fine-scale canopy patterns. The accompanying correlation matrix reflects moderately high positive correlations, driven largely by the substantial zero-valued (void) regions in the maps.

Figure~\ref{fig:forestsemantic_3d_pcd} renders the 3D point clouds colored by the same feature groups.
Compared with raw intensity alone, the \textit{I.R.Z} stack reveals clearer stratification between classes; normals and geometric descriptors (\textit{C.A.P}) accentuate stem boundaries and canopy surfaces; PCA produces smoother yet contextually meaningful color gradients, especially the colors of the stem and foliage, looking very similar to the manually assigned color of the ground truth labels. Figure~\ref{fig:forestsemantic_balls} visualizes virtual spheres back-projected from the colorized feature maps.

Together, these visualizations qualitatively confirm that the multi-group feature framework in our pipeline transfers effectively to new TLS forest environments.

\paragraph{Quantitative trend}
\input{tex_of_fig_tab/tab_forestsemantic_feature_compare}
Table~\ref{tab:forestsemantic-feature-comparison} shows that geometric features dominate performance on the \textit{ForestSemantic} plots, while radiometric preprocessing alone yields only marginal gains. Performance improves consistently with feature fusion, with the best results achieved by the IRZ\_N3\_CAP\_PCA configuration (mIoU = 0.511), indicating that combining complementary geometric and radiometric cues is most effective in complex forest scenes.
This performance level is broadly comparable in magnitude to results reported in the dataset reference paper for state-of-the-art methods evaluated on simpler 3-class (trunk–branch–foliage) settings~\cite{liang2025forestsemantic}.

\subsubsection{Semantic3D}
The \emph{Semantic3D} dataset comprises approximately 1.96 billion points annotated into eight semantic classes~\cite{hackel2017semantic3d}: 
(1) man-made terrain, (2) natural terrain, (3) high vegetation, (4) low vegetation, (5) buildings, 
(6) hardscape, (7) scanning artifacts, and (8) cars. 

Prior to spherical projection, we applied a two-stage hybrid subsampling strategy. 
First, stratified sampling was used to preferentially retain rare classes (frequency $\leq 1\%$), allocating 5\% of a 12M-point target budget to mitigate the loss of minority categories such as scanning artifacts and cars. 
Second, voxel grid downsampling with a resolution of 0.01~m was applied to enforce spatial uniformity while preserving the enriched class distribution. 
This design facilitates scalable processing of large-scale urban point clouds while maintaining semantic diversity. During training, 20\% of the data were held out for validation, and final evaluation was conducted on all 15 test scans.

\paragraph{Qualitative transfer}
\input{tex_of_fig_tab/fig_semantic3d_2d}
\input{tex_of_fig_tab/fig_semantic3d_pcd}
\input{tex_of_fig_tab/fig_semantic3d_balls}
Figure~\ref{fig:Semantic3D_spherical_maps} presents the 2D spherical-projection maps derived from a preprocessed \emph{Semantic3D} scan.  Stacking the basic LiDAR channels—intensity, range, and inverse height—already discriminates facades, ground, and hardscape. The surface-normal pseudo-color stack offers the most intuitive visual cue for human inspection, clearly distinguishing buildings and street furniture by orientation.  Geometric descriptors (anisotropy, curvature, planarity) further sharpen structural edges, while the statistical transforms (PCA, MNF, ICA) distill the dominant variance into three-band composites that reveal subtle material transitions not evident in the single channels.
Pearson correlation coefficients \(|\rho|\leq0.4\) demonstrate that the different feature maps convey largely complementary information, supporting their subsequent fusion.

Figure~\ref{fig:semantic3d-pcd} renders the 3D point clouds colored by the original RGB values and the proposed feature groups. Similar to the observations on 2D feature maps, the proposed feature maps provide additional insights into object characteristics compared to original RGB or Intensity alone.
Figure~\ref{fig:semantic3d_balls} visualizes virtual 3D ``rings'' back-projected from the colorized feature maps - with the top and bottom cropped (\(Zenith\in[45^\circ, 136^\circ]\))  from what used to be ``spheres''.

\paragraph{Quantitative trend}
\input{tex_of_fig_tab/tab_semantic3d_feature_compare}
Table~\ref{tab:semantic3d-feature-comparison} reports segmentation performance on the \textit{Semantic3D} benchmark for different feature-input combinations. Consistent with observations in forest scenes, multi-feature configurations outperform single-feature inputs, with IRZ\_N3\_CAP achieving the best overall performance across global metrics. When evaluated on the official Semantic3D test set under the standard protocol, this configuration attains an mIoU of 0.516, placing it within the performance range of mid-ranked methods on the official Semantic3D benchmark website~\cite{semantic3d_leaderboard}.

\section{Discussion}
\paragraph{High-quality 2D and 3D visualizations} A key outcome of this study is the demonstration that multi-channel visualizations—spanning 2D and 3D representations—provide an interpretable medium for TLS scene recognition. Our outputs form a three-tier suite: (i) \emph{2D spherical feature maps} that stack basic features, geometric features, and statistical descriptors to reveal vertical stratification, edges, and orientation patterns; (ii) \emph{3D colorized point clouds} that enable detailed inspection of boundaries, occlusions, and scan artifacts; and (iii) \emph{3D virtual spheres} that act as compact, visually striking ``thumbnails'' of entire scans. These products serve complementary purposes: 2D maps enable segmentation in 2D spherical space, 3D point clouds allow fine-grained error verification, and virtual spheres provide global summaries at a fraction of the data size. Together, this visualization strategy accelerates dataset triage, guides annotators, and supports cross-scene comparison, enhancing clarity in complex environments.

\paragraph{Data efficiency}
Our experiments on the \emph{Mangrove3D} dataset reveal that ensemble segmentation performance saturates after approximately 12 annotated scans, regardless of the specific feature configuration. This finding underscores the data efficiency of the proposed pipeline, particularly when combined with uncertainty-guided sample selection. By focusing annotation effort on the most uncertain regions—as identified by ensemble-derived mutual information maps—it is possible to approach peak performance with substantially fewer labeled samples. In practical terms, this reduces both the cost and time required to develop a high-performing TLS segmentation model, making the approach attractive for large-scale or time-sensitive monitoring campaigns. The observed saturation point also provides guidance for field data annotation strategies: beyond a certain threshold (e.g., 12 scans for \emph{Mangrove3D}), manually annotate additional scans may yield diminishing returns, unless the new samples introduce substantially novel scene geometries or conditions.

\paragraph{Feature importance}
Across all three benchmarked datasets—\emph{Mangrove3D}, \emph{ForestSemantic}, and \emph{Semantic3D}—feature enrichment consistently improves segmentation accuracy, but the gains are not uniform across feature types. Surface normals (\textit{N3}) emerge as the most consistently valuable, delivering substantial improvements in class-level IoUs for geometrically complex structures such as tree stems, roots, and small urban objects. Basic features (\textit{I.R.Z}) provide more modest but still meaningful gains, particularly for classes with strong vertical separation or height-dependent geometric-radiometric patterns. Importantly, the optimal trade-off between accuracy and computational cost is achieved with compact, nine-channel configurations (e.g., \textit{IRZ\_N3\_CAP}), which capture nearly all available discriminative power, while avoiding the redundancy and increased training time associated with larger feature stacks. This suggests that thoughtful feature selection, rather than maximal channel inclusion, is critical for efficient TLS segmentation pipelines.

\paragraph{Generalization across TLS datasets}
The proposed pipeline demonstrates consistent cross-domain robustness when transferred from the mangrove-dominated \emph{Mangrove3D} dataset to the structurally distinct \emph{ForestSemantic} (boreal forest) and \emph{Semantic3D} (urban) benchmarks. In all cases, feature enrichment improves segmentation accuracy, and the relative ranking of feature groups is broadly preserved despite differences in vegetation type, terrain complexity, and object composition.

Several design choices render the proposed framework both scalable and adaptable. First, all features are derived exclusively from LiDAR data, ensuring applicability to TLS scans regardless of the availability or reliability of RGB imagery. Second, feature fusion is performed at the bottleneck stage, enabling straightforward substitution of different encoder-decoder backbones (e.g., CNNs or Transformers) without altering the overall pipeline. Third, the patch size is determined adaptively, allowing the model to accommodate varying input resolutions and scene extents while maintaining computational efficiency. Fourth, the framework accommodates a flexible number of input channels under a triadic grouping scheme, thereby enabling seamless integration of diverse radiometric, geometric, or statistical descriptors without architectural modification. Finally, the \emph{pseudo-scanner center} concept introduced for \textit{ForestSemantic} can be extended to mobile or UAS-based LiDAR platforms that cover larger areas at lower point densities, though further optimization may be required to determine ideal center placements, spherical map resolutions, and truncation distances.

Together, the cross-dataset results and modular design show that the pipeline can be deployed in diverse TLS environments with minimal adjustment—an essential feature for large-scale monitoring programs where site conditions and sensor setups vary widely.

\paragraph{Limitations and future directions}
While the proposed multi-model ensemble improves segmentation accuracy and uncertainty estimation, its increased computational cost may limit applicability in real-time or resource-constrained settings. 
Future work will therefore investigate more robust and generic preprocessing strategies, as well as lighter backbone architectures or knowledge distillation approaches to reduce redundancy while preserving pseudo-label quality. 
Additionally, although uncertainty guidance was shown to restructure the annotation workflow and reduce manual labeling scope, this study does not report quantitative measures of annotation efficiency such as time or interaction counts; future work will incorporate systematic logging of annotation processes to enable controlled, quantitative evaluation of human annotation effort. 
Further efforts are also needed to narrow the gap between uncertainty estimates and actual error distributions by adopting improved uncertainty estimation methods.

\section{Conclusion}
We present a feature-enriched, uncertainty-aware pipeline for semantic segmentation of TLS point clouds, supported by a modular design that integrates diverse feature groups, flexible encoder backbones, and ensemble-based uncertainty estimation. Across mangrove, boreal forest, and urban benchmarks, the framework consistently improves segmentation accuracy, with compact nine-channel configurations capturing nearly all available discriminative power. The inclusion of complementary visualization products—2D spherical feature maps, 3D colorized point clouds, and compact virtual spheres—proves valuable for both annotator guidance and rapid scene assessment, enabling efficient, targeted annotation and dataset triage. The pipeline’s reliance solely on LiDAR-derived features, coupled with its demonstrated cross-domain robustness, positions it as a scalable and portable solution for large-scale monitoring in forestry, ecology, and urban environments. Future developments in preprocessing, backbone efficiency, uncertainty evaluation, and mobile deployment are expected to further expand its applicability and impact.

While sensor fusion is becoming increasingly popular, our feature-enrichment results highlight the untapped potential of LiDAR point clouds alone. Careful exploitation of LiDAR-derived features can, in some cases, surpass early (or \enquote{raw}) LiDAR-camera fusion in the spherical projection space, revealing a distinct and informative view \emph{through the perspective of LiDAR}. We therefore conclude that for downstream tasks such as semantic segmentation, rigorous feature design, and preprocessing of LiDAR data are no less important than integrating multi-source sensing.

\section*{Acknowledgments}
The authors are grateful to Dr. Nidhal Carla Bouaynaya and Dr. Bartosz Krawczyk for their valuable insights on uncertainty evaluation and semantic segmentation methods. We thank Mr. Brett Matzke for his assistance in setting up the computing resources. We also acknowledge RIT Research Computing~\cite{https://doi.org/10.34788/0s3g-qd15} for providing access to NVIDIA A100 computing resources.

\section*{CRediT authorship contribution statement}
\textbf{Fei Zhang}: Conceptualization, Methodology, Software, Validation, Formal Analysis, Investigation, Data Curation, Visualization, Writing - original draft, Project administration, Supervision.
\textbf{Fei Zhang} and \textbf{Rob Chancia}: Conceptualization, Experimental design.
\textbf{Fei Zhang}: Experiment execution.
\textbf{Fei Zhang} and \textbf{Josie Clapp}: Data Annotation (\emph{Mangrove3D} dataset).
\textbf{Rob Chancia}, \textbf{Richard MacKenzie}, and \textbf{Jan van Aardt}: Data Collection (\emph{Mangrove3D} dataset).
\textbf{Rob Chancia} and \textbf{Amirhossein Hassanzadeh}: Validation, Writing - review \& editing, Software.
\textbf{Jan van Aardt}: Writing - review \& editing, Supervision.
\textbf{Dimah Dera}: Writing - review \& editing.

\section*{Funding}
This work was supported by the USDA Forest Service [project numbers 20-JV-11272136-016].

\section*{Conflicts of Interest}
The authors declare that there is no conflict of interest regarding the publication of this article.

\section*{Data and Code Availability}
\datacodeavailability

\appendix
\section*{Supplementary Materials}
\input{Supple-Mater0}
\input{Supple-Mater1}
\input{Supple-Mater2}
\input{Supple-Mater3}
\input{Supple-Mater4}

\bibliographystyle{elsarticle-num}
\bibliography{references}

\end{document}

%% file: tex_of_fig_tab/tab_annotation_tool_compare.tex
\begin{table*}[htbp]
\centering
\caption{Comparison of commonly used 3D point cloud annotation software and their key features. 
Clip \& segment: region-based clipping and segmentation; Paintbrush: brush-like manual labeling; 
Multi-view: annotation from multiple perspectives; RGB cross-ref.: annotation aided by associated RGB/multispectral images; 
DL pre-annot.: integrated deep learning pre-annotation; Typical applications: primary research or industry usage areas.}
\label{tab:annotation-tools-comparison}
\resizebox{\textwidth}{!}{
\begin{tabular}{@{}l c c c c c c l@{}}
\toprule
\textbf{Software} & \textbf{Open} & \textbf{Clip \&} & \textbf{Paint-} & \textbf{Multi-} & \textbf{RGB} & \textbf{DL Pre-} & \textbf{Typical} \\
\textbf{Name} & \textbf{Source} & \textbf{Segment} & \textbf{brush} & \textbf{view} & \textbf{Cross-ref.} & \textbf{annot.} & \textbf{Applications} \\
\midrule
CloudCompare & Yes & Yes & No  & No  & No  & No  & General research, surveying, geology \\
3D Slicer    & Yes & Yes & Yes & Yes & Yes & No  & Medical imaging, biomedical applications \\
LabelCloud   & Yes & Yes & No  & No  & Yes & No  & Autonomous driving, urban scene annotation \\
Scalabel     & Yes & Yes & No  & Yes & Yes & No  & Autonomous driving, robotics, object detection \\
MathWorks LiDAR Labeler & No  & Yes & No  & Yes & Yes & Yes & Autonomous vehicles, robotics, industry \\
Segments.ai & No  & Yes & Yes & Yes & Yes & Yes & Autonomous driving, urban analytics \\
Autodesk ReCap\cite{autodesk_recap} & No  & Yes & No  & Yes & Yes & No  & Engineering, architecture, construction \\
Trimble RealWorks\cite{trimble_realworks} & No  & Yes & No  & Yes & Yes & No  & Infrastructure, surveying, civil engineering \\
Semantic Segmentation Editor\cite{semantic_segmentation_editor} & Yes & Yes & Yes & Yes & Yes & No  & Autonomous driving \\
LiDAR360\cite{semantic_segmentation_editor} & No  & Yes & Yes & Yes & Yes & Yes & Forestry, surveying \& mapping, general usage \\
\bottomrule
\end{tabular}
}

\end{table*}

%% file: tex_of_fig_tab/fig_Palau_mangrove_and_CBL_scanner.tex
\begin{figure}[!tbph]
    \centering
    \begin{minipage}[c]{0.35\textwidth}
        \includegraphics[width=\linewidth]{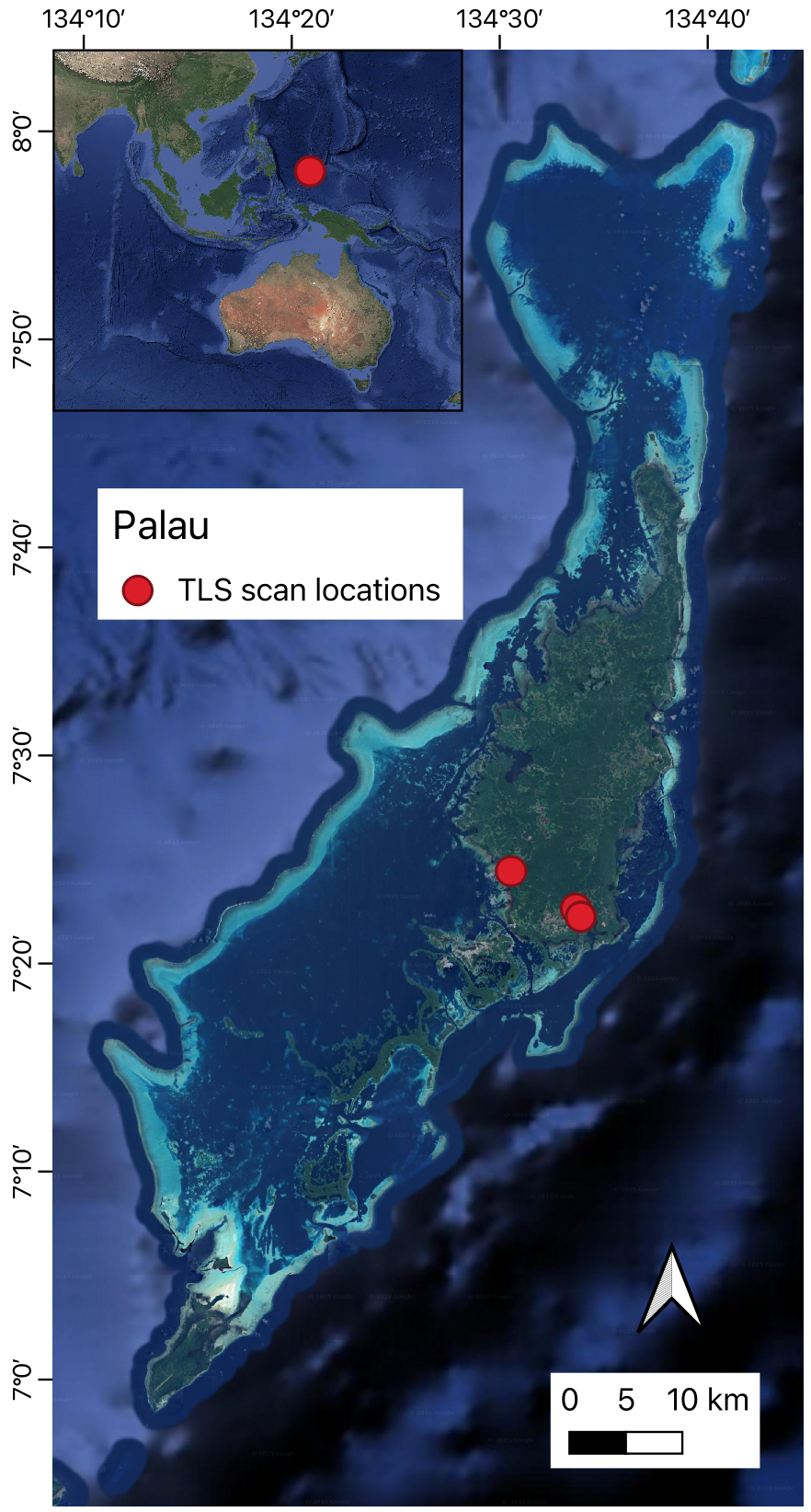}
        \caption*{(a) TLS scan sites in Palau}
    \end{minipage}\hfill
    \begin{minipage}[t]{0.64\textwidth}
        \includegraphics[width=\linewidth]{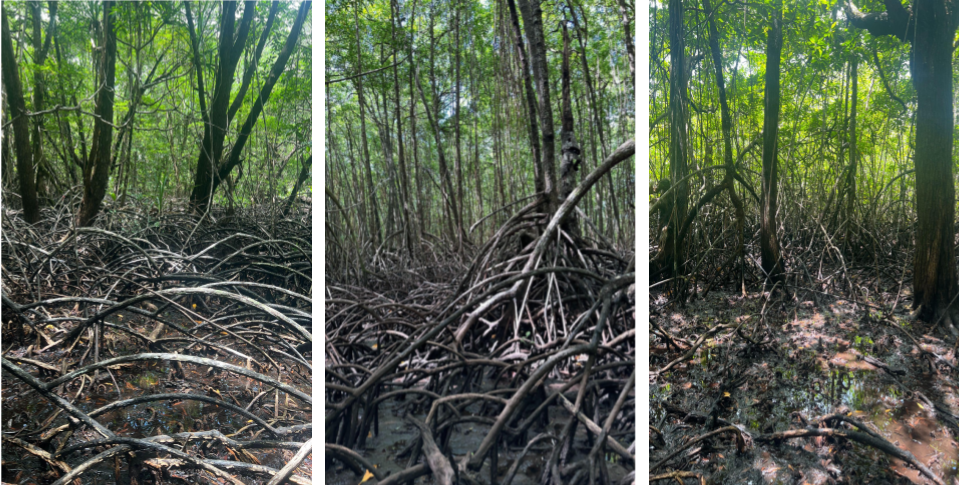}
        \caption*{(b) Field photos from three sites}
        
        \vspace{2mm}
        
        \begin{minipage}[t]{0.48\linewidth}
            \centering
            \includegraphics[width=\linewidth]{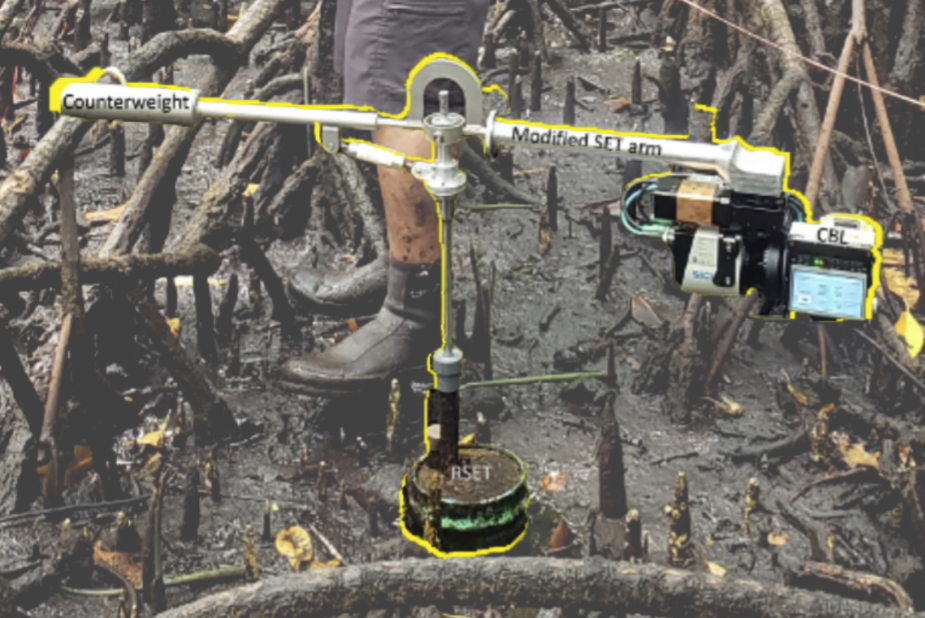}
            \caption*{(c) TLS setup}
        \end{minipage}\hfill
        \begin{minipage}[t]{0.48\linewidth}
            \centering
            \includegraphics[width=\linewidth]{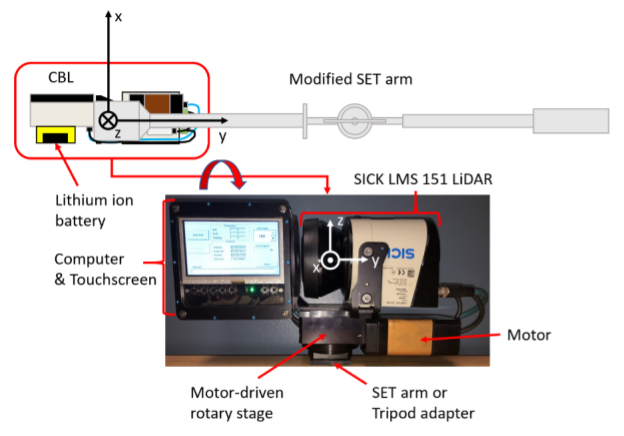}
            \caption*{(d) CBL and SET components}
        \end{minipage}
    \end{minipage}
    
    \caption{Map and field setup for TLS scans in Palau mangrove forests}
    \label{fig:palau_rset}
\end{figure}

%% file: tex_of_fig_tab/tab_mangrove3d_noise.tex
\begin{table}[t]
\centering
\caption{Local surface roughness statistics of the \emph{Mangrove3D} dataset.
Roughness is computed as $\sqrt{\lambda_3}$ from local PCA using a radius-based neighborhood
(radius = 0.06~m, capped at 60 neighbors) and summarized using robust statistics.
Points with insufficient neighbors were excluded from roughness computation.}
\label{tab:mangrove_noise_vs_class}
\begin{tabular}{clcccccc}
\toprule
Label & Region & Median (cm) & Q25--Q75 (cm) & P95 (cm) & Excluded (NaN) & Total points \\
\midrule
1 & Ground \& Water & 1.95 & 1.38--2.59 & 3.66 & 62 & 12,347,797 \\
2 & Stem & 8.44 & 5.21--10.37 & 11.67 & 1,570 & 1,788,806 \\
3 & Canopy & 9.24 & 7.13--10.59 & 12.20 & 49,534 & 6,096,210 \\
4 & Roots & 3.43 & 2.40--6.94 & 10.71 & 18,680 & 10,405,208 \\
5 & Objects & 2.19 & 1.67--2.75 & 3.48 & 0 & 373,016 \\
\bottomrule
\end{tabular}
\end{table}

%% file: tex_of_fig_tab/fig_mangrove3d_density_vs_range_hight.tex
\begin{figure}[!tbph]
    \centering
    \begin{minipage}[b]{0.49\textwidth}
        \begin{subfigure}[b]{\textwidth}
            \centering
            \includegraphics[width=\textwidth]{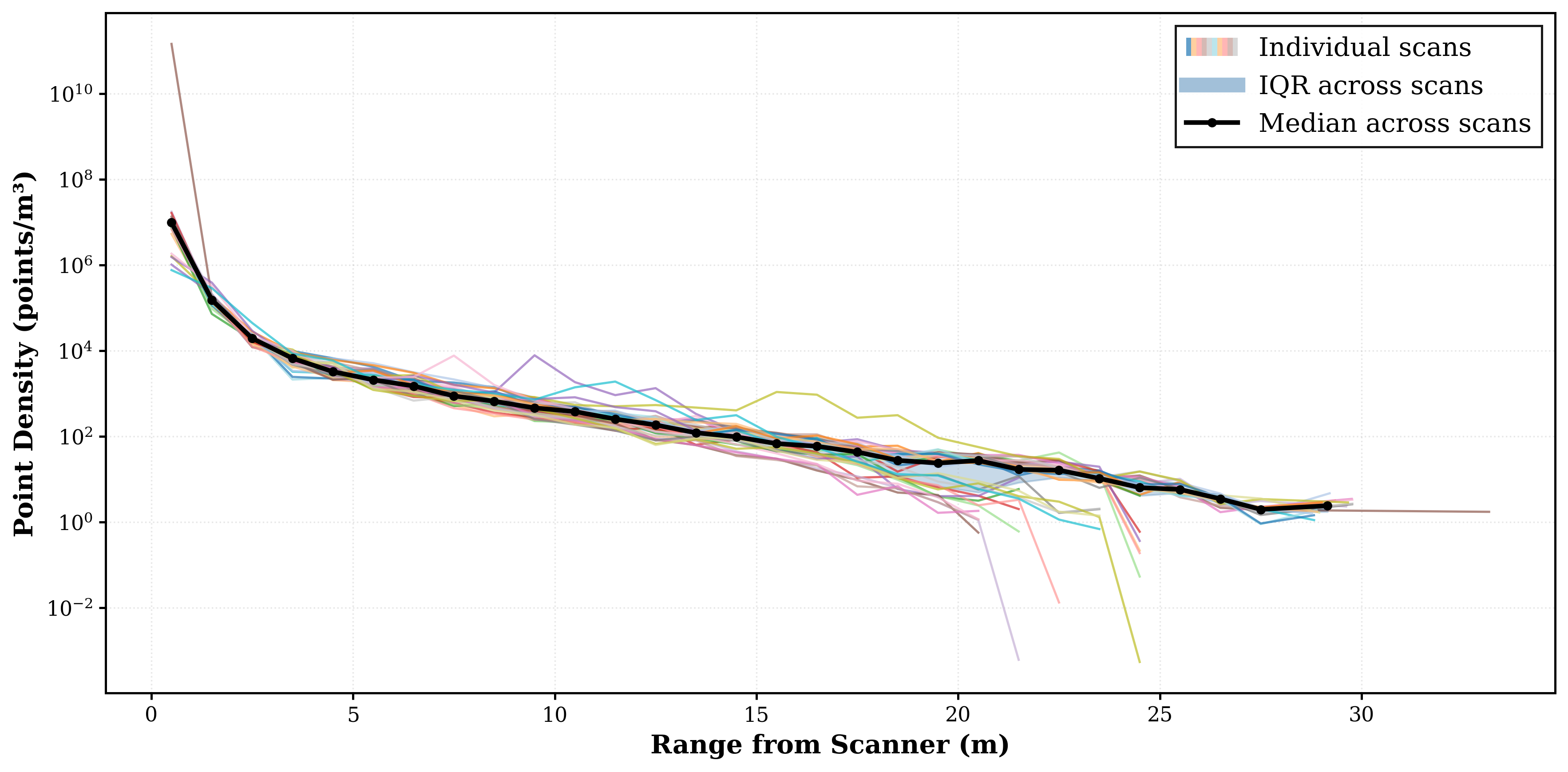}
            \caption*{(a) Point density vs range}
        \end{subfigure}
    \end{minipage}
    \begin{minipage}[b]{0.49\textwidth}
        \begin{subfigure}[b]{\textwidth}
            \centering
            \includegraphics[width=\textwidth]{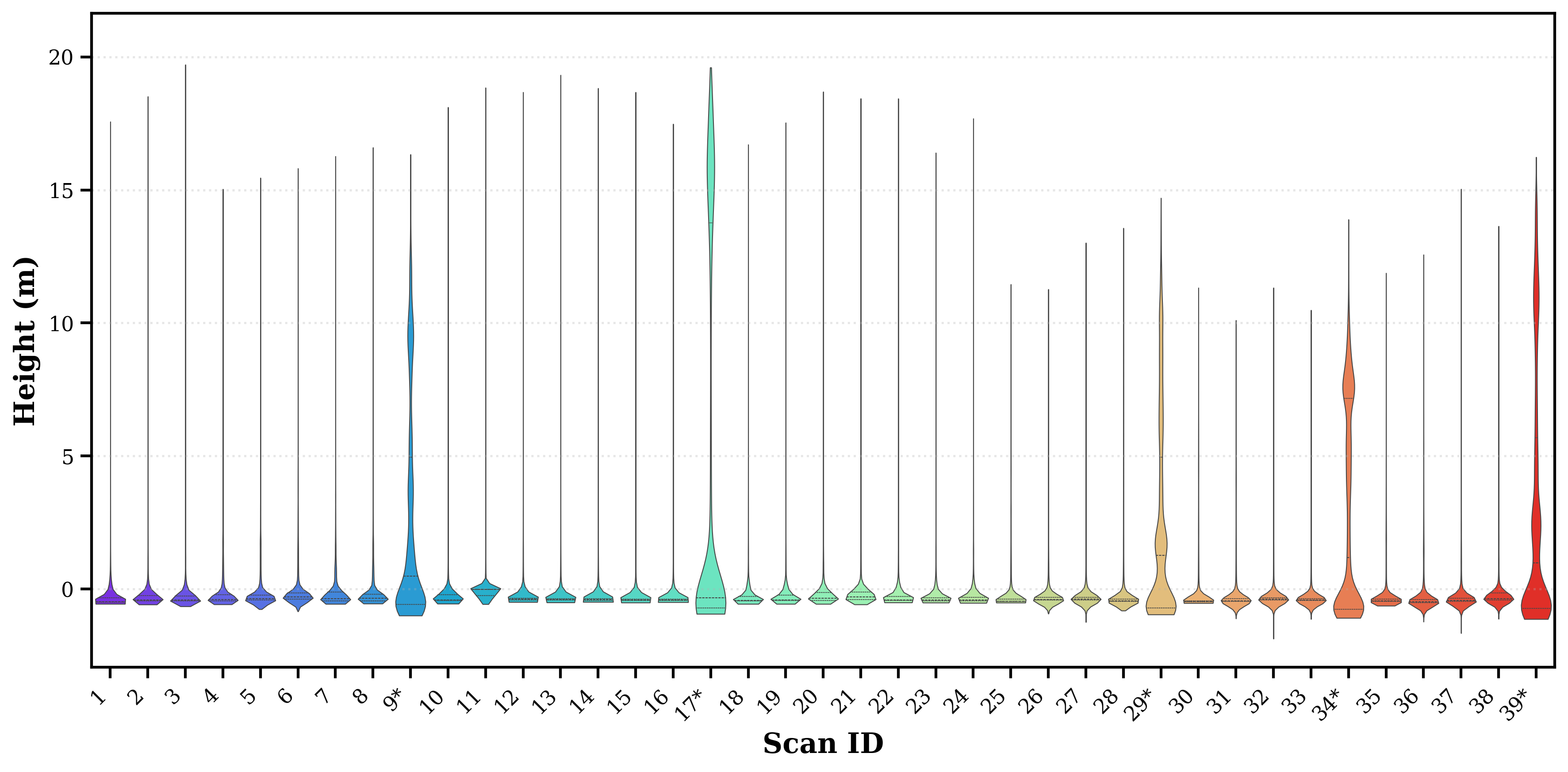}
            \caption*{(b) Point density along height}
        \end{subfigure}
    \end{minipage}
    
    \caption{Point density characteristics across \emph{Mangrove3D} scans.
(a) Point density as a function of range from the scanner, showing individual scans together with the median and interquartile range across scans.
(b) Vertical distribution of point returns per scan shown as violin plots, where violin width indicates the relative frequency of points at each height.
Scan IDs marked with * correspond to acquisitions with the scanner in an upright orientation; all other scans were acquired with the scanner positioned upside down.}

    \label{fig:mangrove3d_density}
\end{figure}

%% file: tex_of_fig_tab/fig_mangrove3d_class_distribution.tex
\begin{figure}[!tbph]
    \centering
    \begin{minipage}[b]{0.49\textwidth}
        \begin{subfigure}[b]{\textwidth}
            \centering
            \includegraphics[width=\textwidth]{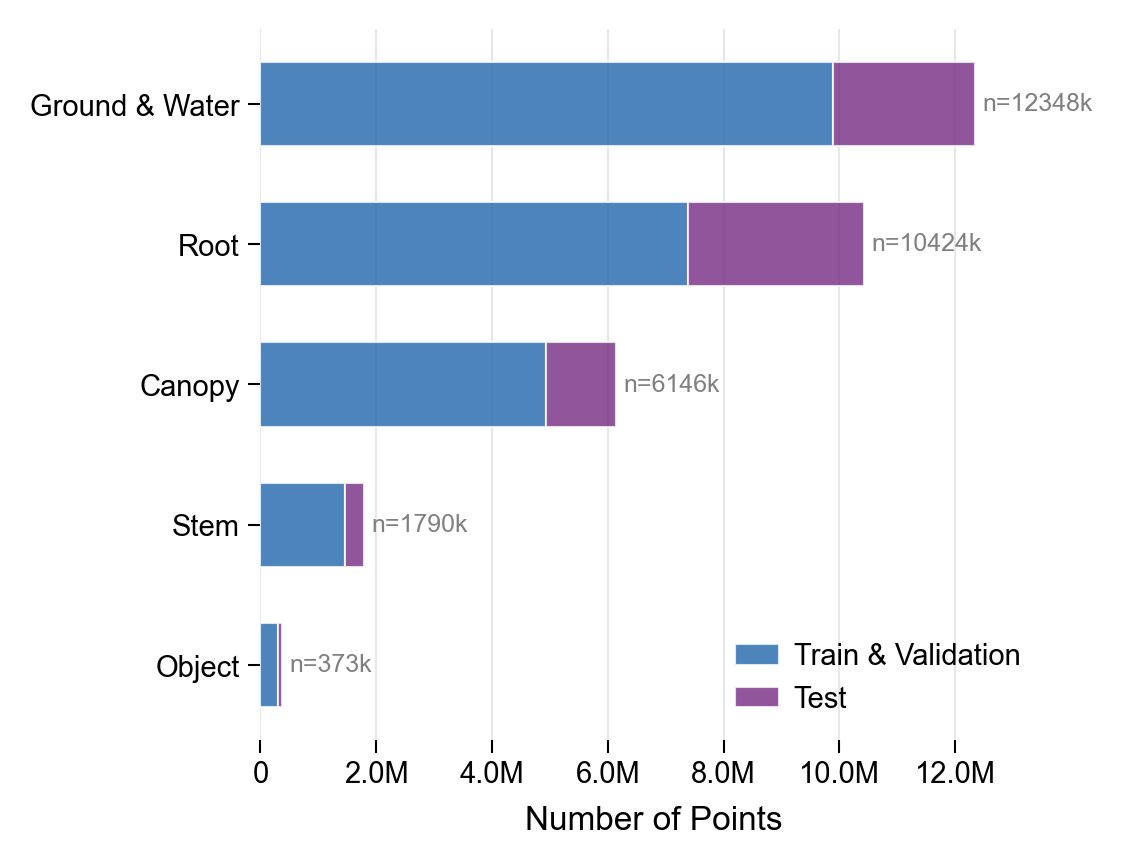}
            \caption*{(a) Class counts for point cloud labels}
        \end{subfigure}
    \end{minipage}
    \begin{minipage}[b]{0.49\textwidth}
        \begin{subfigure}[b]{\textwidth}
            \centering
            \includegraphics[width=\textwidth]{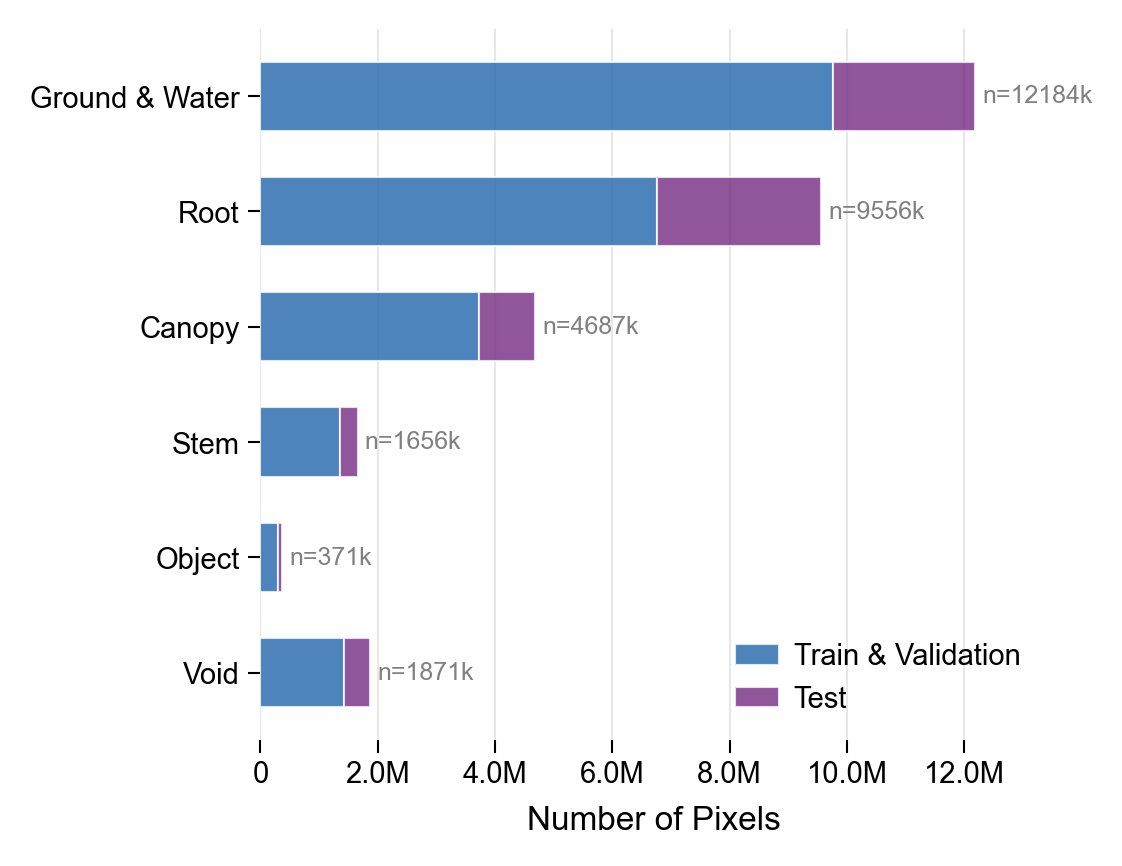}
            \caption*{(b) Class counts for pixel labels}
        \end{subfigure}
    \end{minipage}
    
    \caption{Class counts for the \emph{Mangrove3D} dataset.}
    \label{fig:mangrove3d_class_dist}
\end{figure}

%% file: tex_of_fig_tab/fig_tls_2d3d_seg_pipeline.tex
\begin{figure}[htbp]
    \centering
    \includegraphics[width=\textwidth]{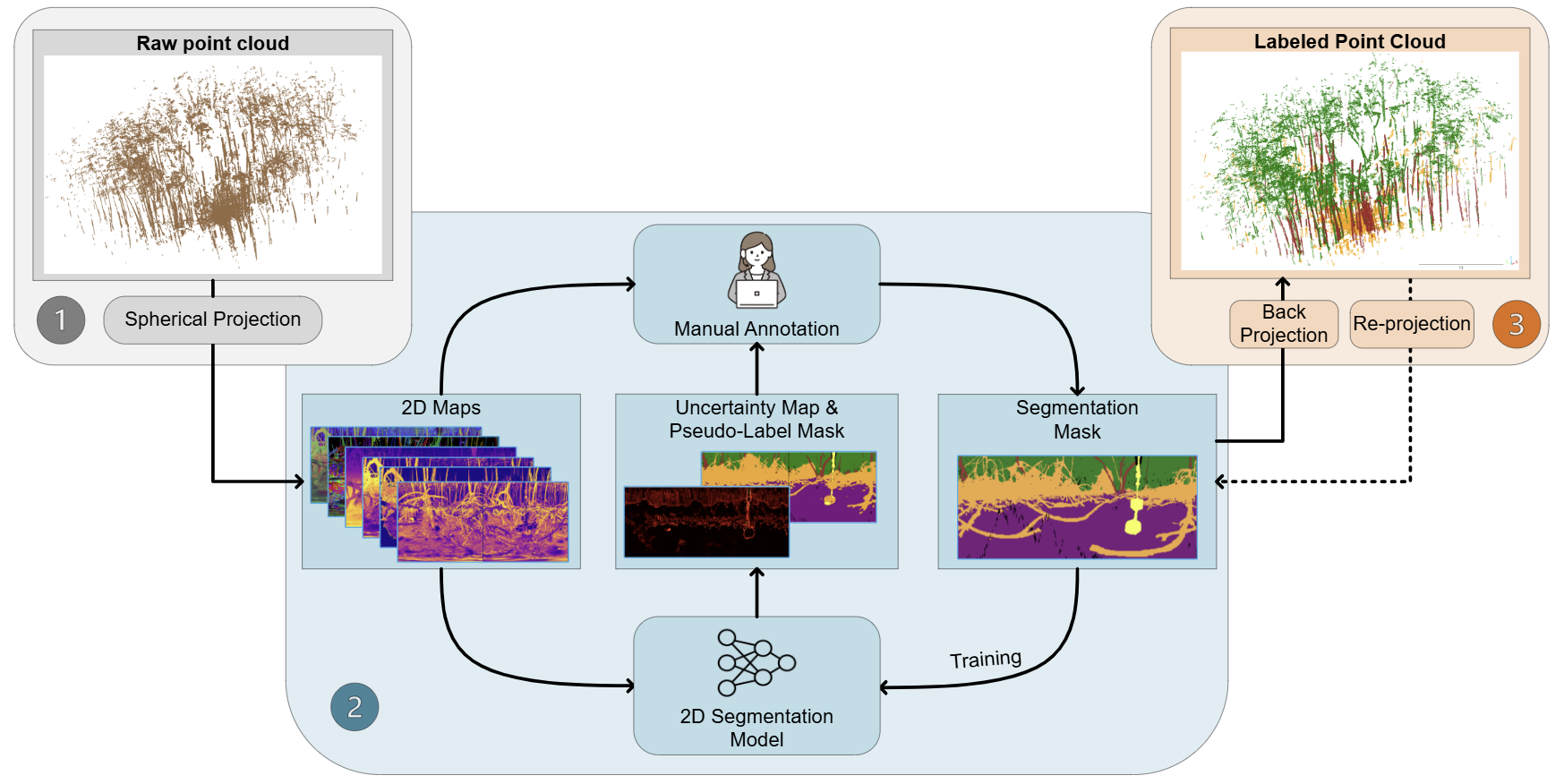}
    \caption{Three-stage workflow for annotating terrestrial-LiDAR scans.
        \textbf{Stage\,1:} Spherical projection converts raw TLS points into
        two-dimensional feature maps and pseudo-RGB images.
        \textbf{Stage\,2:} An iterative loop combines active learning and
        self-training: an emsemble segmentation model is repeatedly refined using
        uncertainty-guided queries and high-confidence pseudo-labels.
        \textbf{Stage\,3:} The resulting 2-D segmentation masks are
        back-projected, followed by label refinement in 3D space and then reproject back to 2D, to yield a fully
        annotated point cloud and refined 2D segmentation mask.}
    \label{fig:tls_2d3d_seg_pipeline_E}
\end{figure}

%% file: tex_of_fig_tab/fig_spherical_projection.tex
    \begin{figure}[htbp]
        \centering
            \begin{subfigure}[c]{0.26\textwidth}
                \centering
                \includegraphics[width=\textwidth]{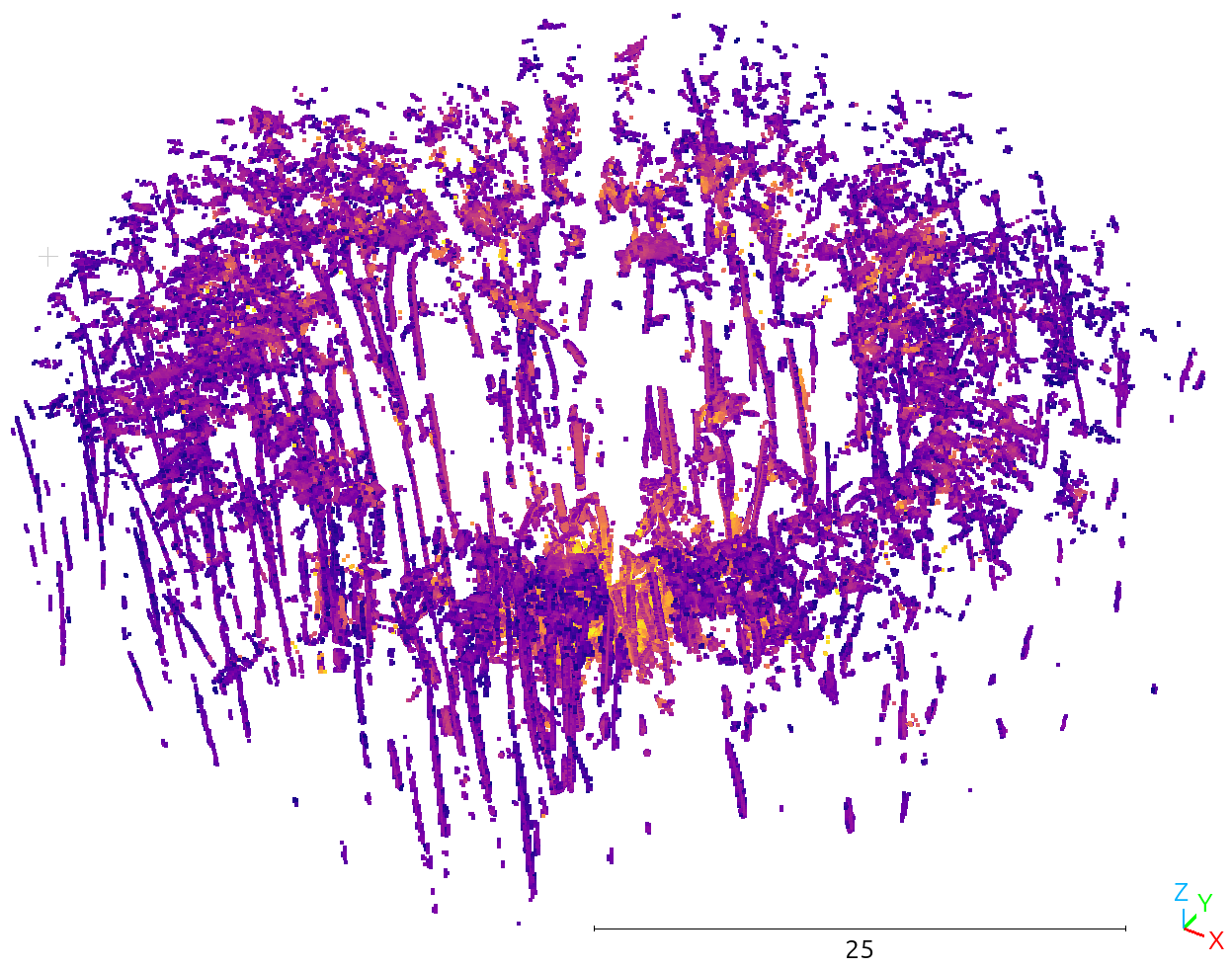}
                \caption*{(a)}
            \end{subfigure}
            \begin{subfigure}[c]{0.2\textwidth}
                \centering
                \includegraphics[width=\textwidth]{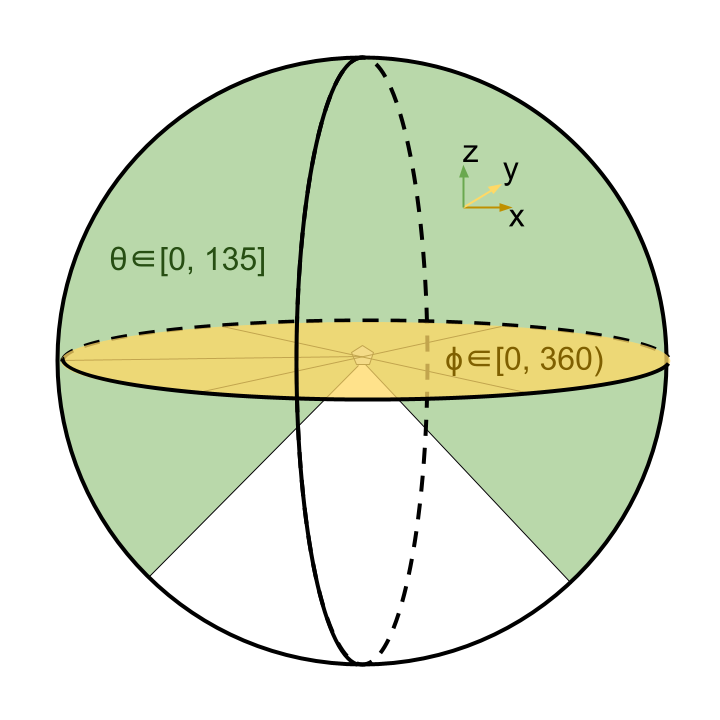}
                \caption*{(b)}
            \end{subfigure}
            \begin{subfigure}[c]{0.5\textwidth}
                \centering
                \includegraphics[width=\textwidth]{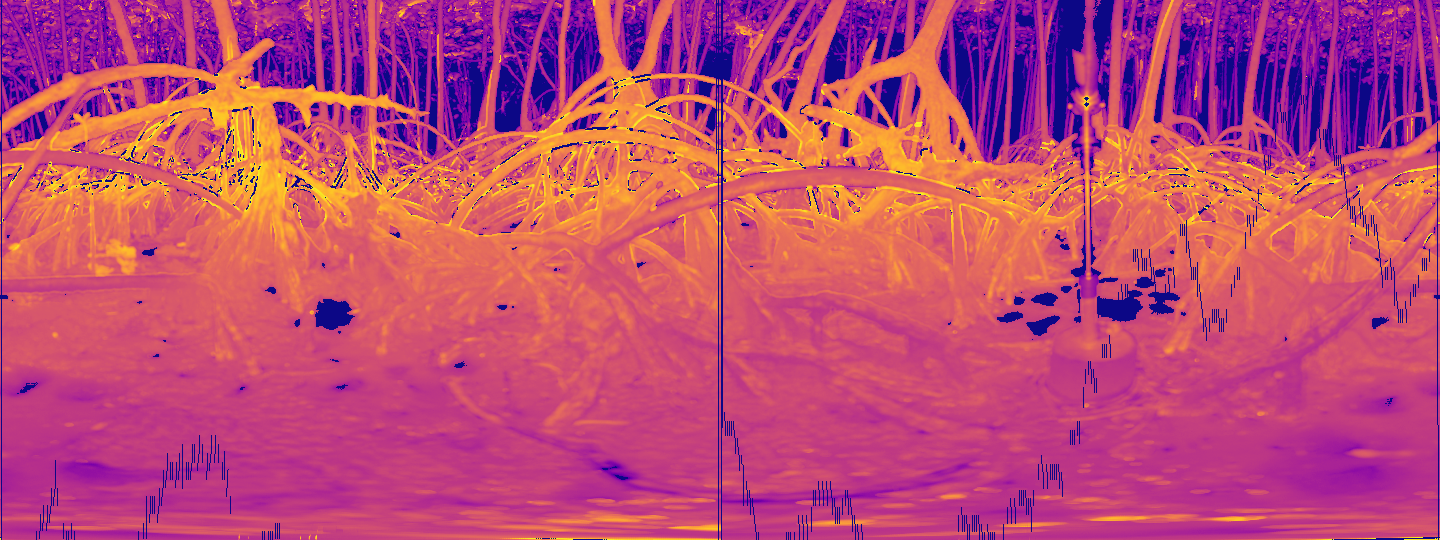}
                \caption*{(c)}
            \end{subfigure}

        \caption{Spherical projection workflow illustrated with a scan of the CBL LiDAR.
        (a) Original 3D point cloud visualized by intensity with plasma color scale. 
        (b) Geometric illustration of the spherical projection. 
        (c) Spherical projection map of raw intensity values. }
        \label{fig:spherical_projection}
    \end{figure}

%% file: tex_of_fig_tab/fig_CBL_density_map.tex
    \begin{figure}[htbp]
        \centering
        \begin{subfigure}[c]{0.95\textwidth}
            \centering
            \includegraphics[width=\textwidth]{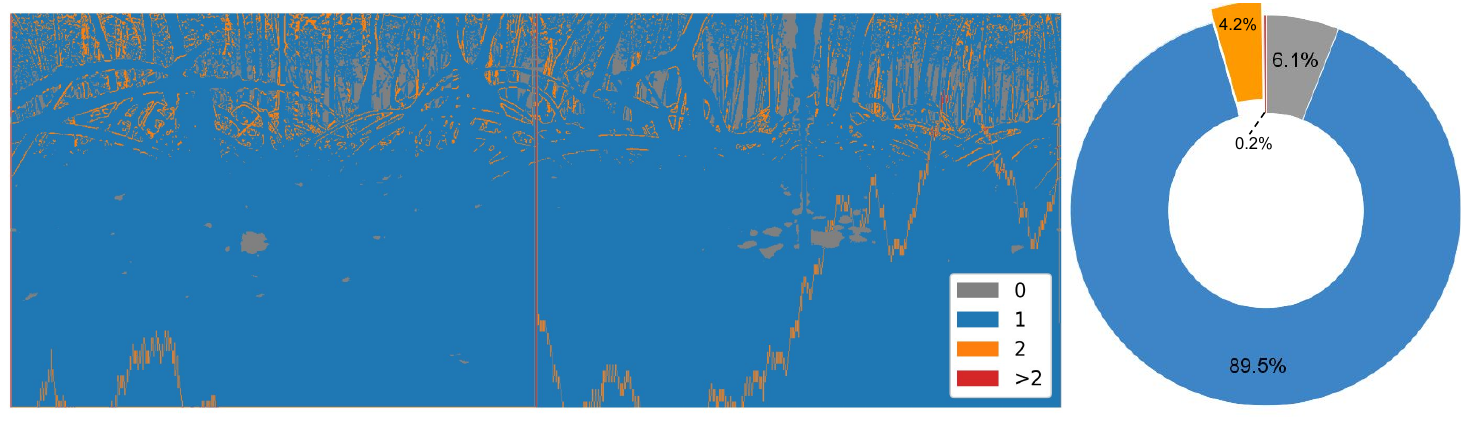}
            \caption*{(a) Spherical projection point-density map with corresponding percentage distribution for a representative scan (same as in Figure~\ref{fig:spherical_projection}.)}
            \label{fig:d}
        \end{subfigure}
        \begin{subfigure}[c]{0.95\textwidth}
            \centering
            \includegraphics[width=\textwidth]{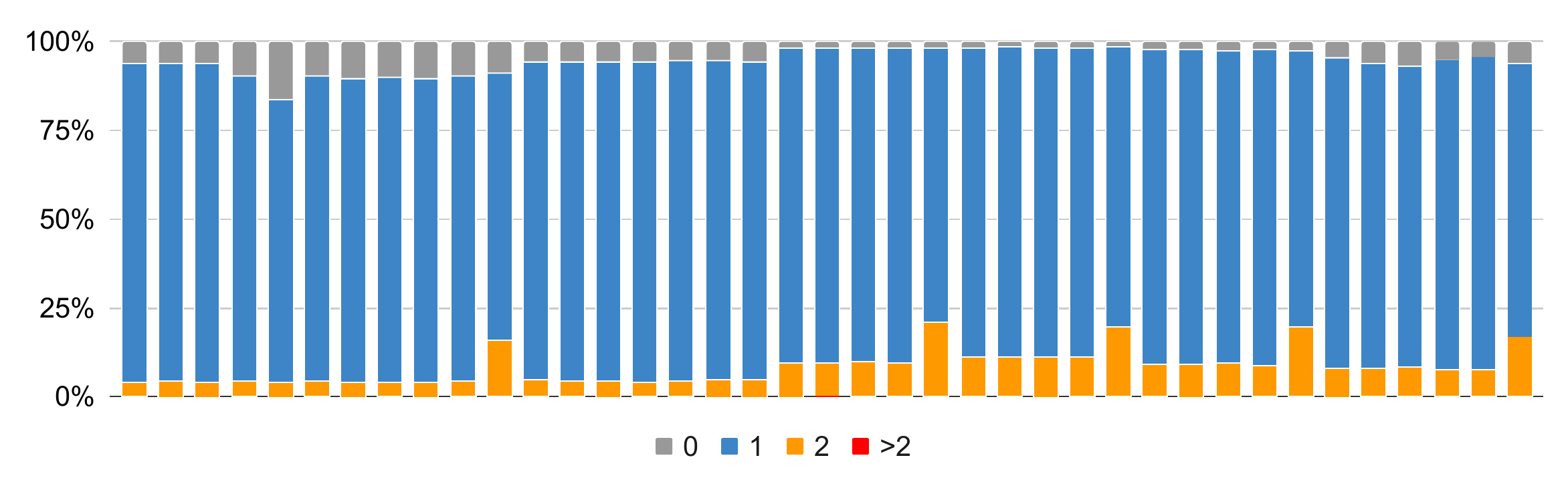}
            \caption*{(b) Scan-wise point-density distribution across the Mangrove3D dataset (39 scans).}
        \end{subfigure}
    
        \caption{Point-density characteristics of the spherical projection map. When the map resolution matches the TLS angular resolution, 80\% - 90\% of pixels contain exactly one point. Pixels with a value of 0 indicate no return from the corresponding angular direction, values of 2 represent dual-return measurements (mostly <10\%), and values greater than 2 occur rarely, typically due to systematic noise.}
        \label{fig:CBL_density_map}
    \end{figure}

%% file: tex_of_fig_tab/fig_pseudolabel-uncertainty.tex
\begin{figure}[htbp]
\centering
\begin{minipage}[c]{0.48\textwidth}
    \begin{subfigure}[c]{\textwidth}
        \centering
        \includegraphics[width=\textwidth]{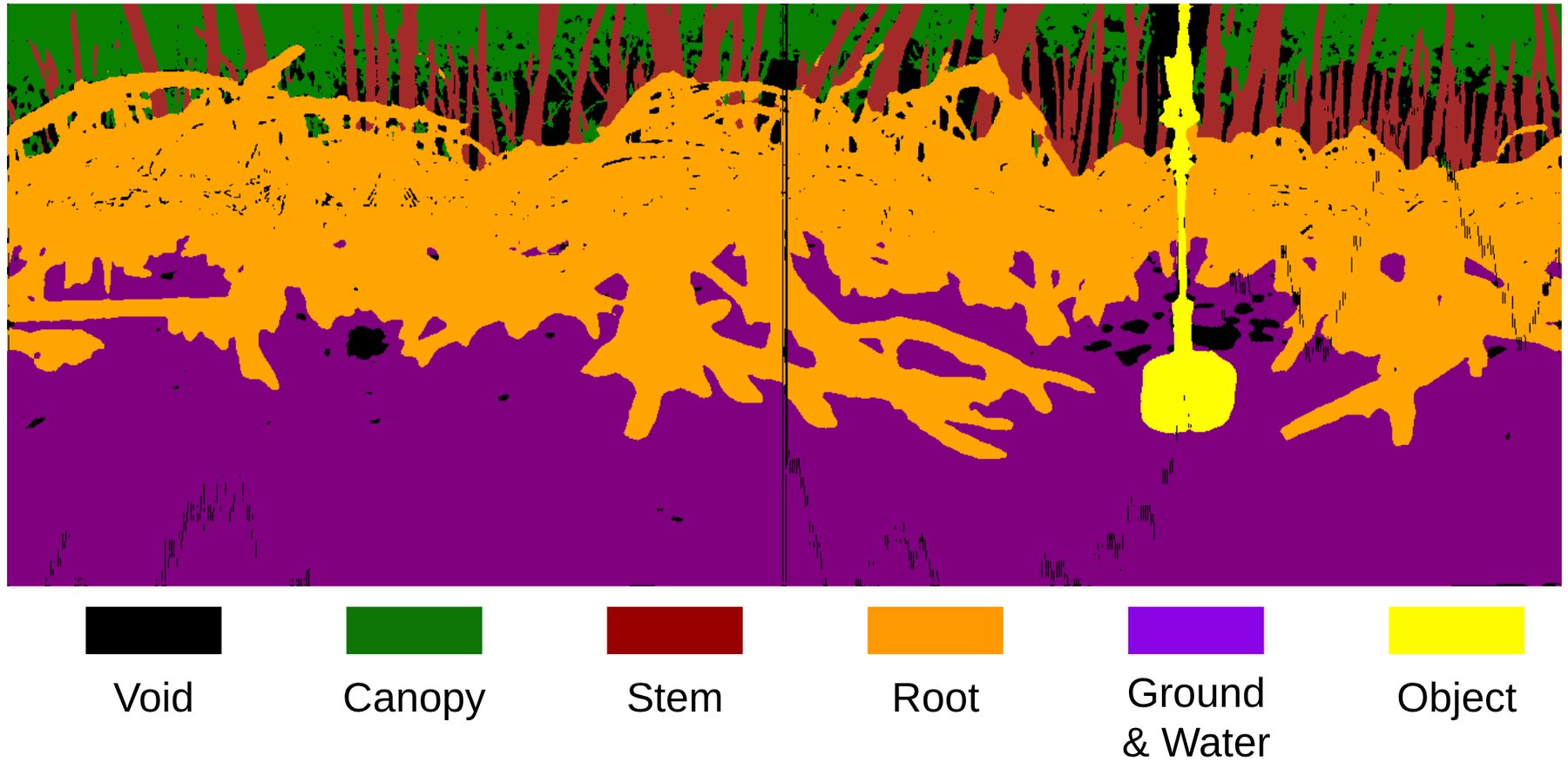}
        \caption{Predicted Segmentation Map}
        \label{subfig:pu-pred-seg}
    \end{subfigure}
\end{minipage}
\begin{minipage}[c]{0.48\textwidth}
    \begin{subfigure}[c]{\textwidth}
        \centering
        \includegraphics[width=\textwidth]{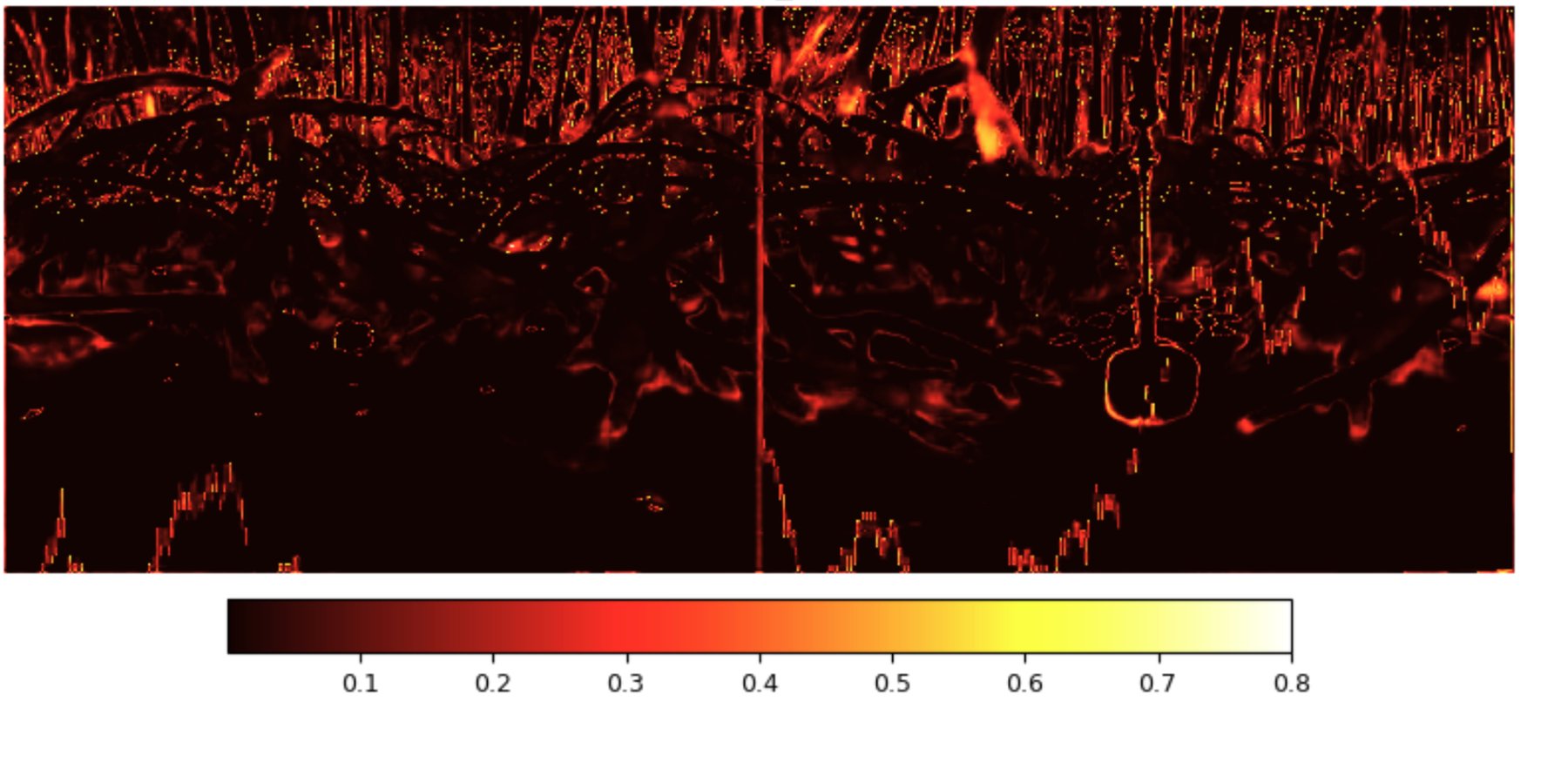}
        \caption{Epistemic uncertainty map}
        \label{subfig:pu-eps-uncert}
    \end{subfigure}
\end{minipage}    
\caption{Examples of (a) a pseudo-label map and (b) epistemic uncertainty calculated from mutual information.}
\label{fig:mangrove3d-pseudolabel-uncertainty}
\end{figure}

%% file: tex_of_fig_tab/fig_ensemble_multiencoder.tex
\begin{figure}[htbp]
\centering
\begin{minipage}[t]{0.48\textwidth}
    \begin{subfigure}[t]{\textwidth}
        \centering
        \includegraphics[width=\textwidth]{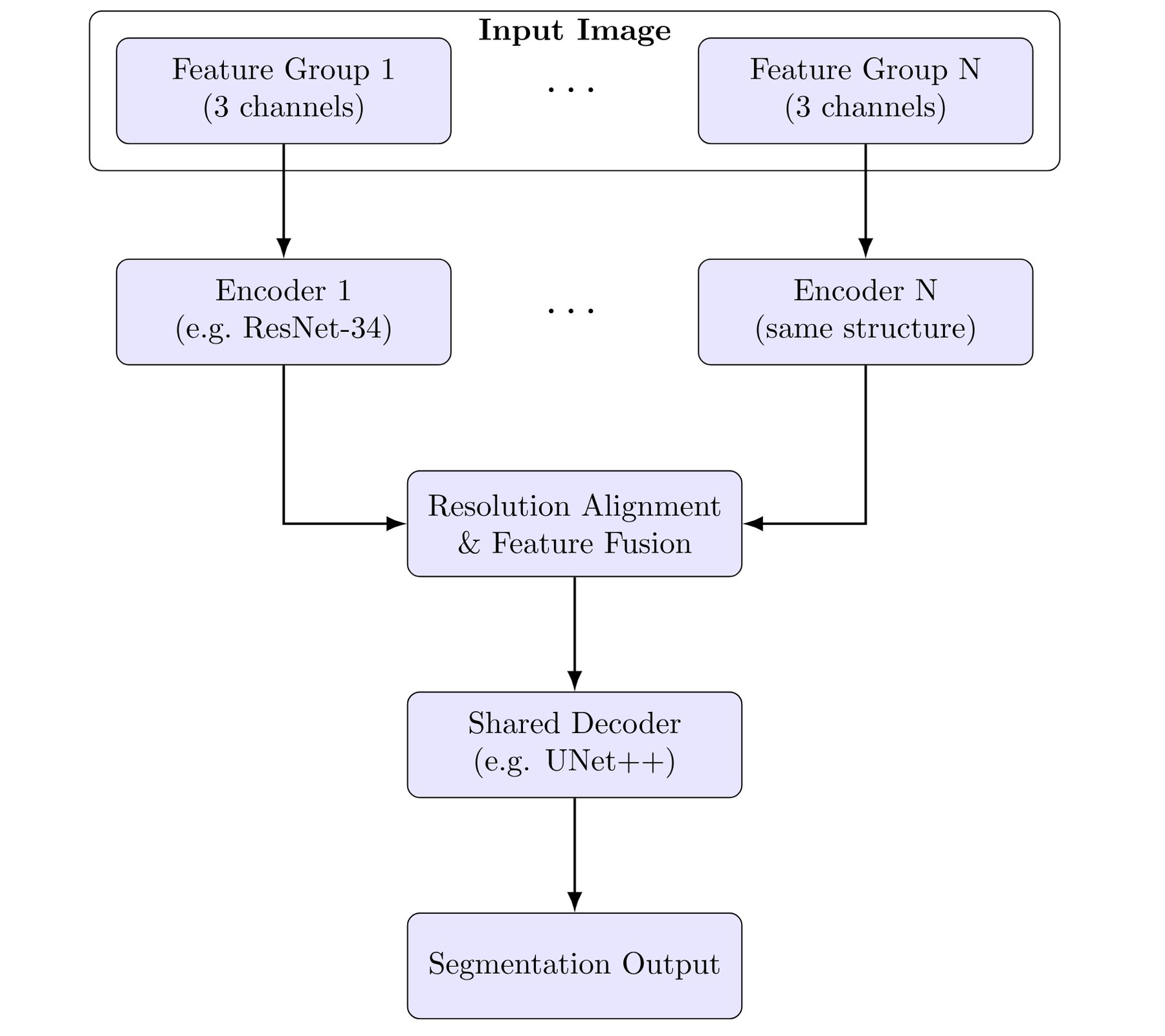}
        \caption{Multi-encoder fusion for $3\times N$-channel heterogeneous feature groups. Dedicated encoders preserve pretrained weights and are fused before decoding.}
        \label{subfig:multiencoder-feature-fusion}
    \end{subfigure}
\end{minipage}    
\hfill
\begin{minipage}[t]{0.48\textwidth}
    \begin{subfigure}[t]{\textwidth}
        \centering
        \includegraphics[width=\textwidth]{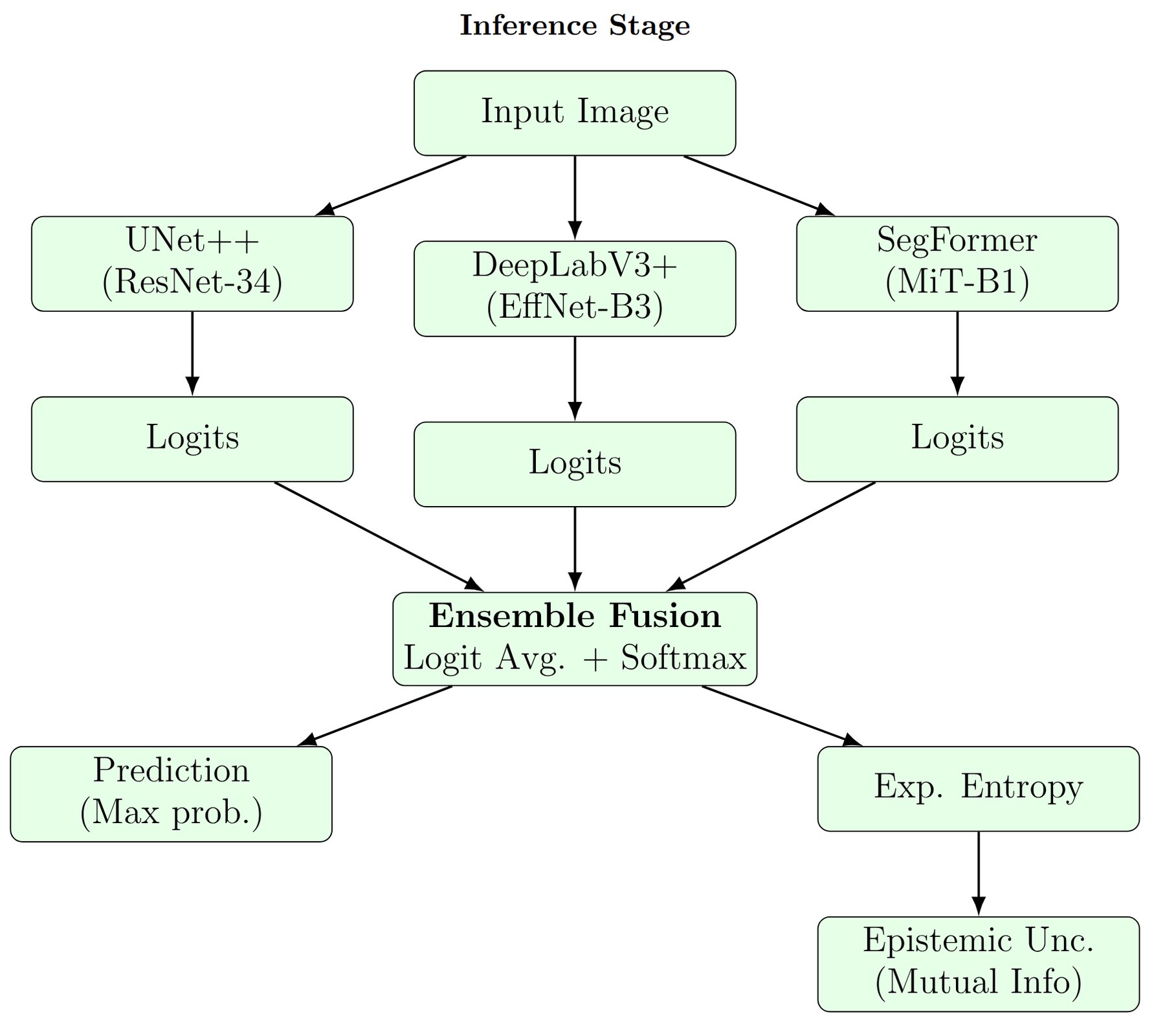}
        \caption{Schematic of the ensemble-based inference pipeline. Each model is trained independently and later fused at inference using logit averaging and softmax.}
        \label{subfig:ensemble-inference}
    \end{subfigure}
\end{minipage}
\caption{Ensemble inference pipeline and model architecture for feature fusion.}
\label{fig:combined_ensemble_fusion}
\end{figure}

%% file: tex_of_fig_tab/tab_uncertainty_guided_workflow.tex
\begin{table}[t]
\centering
\caption{Qualitative comparison of annotation workflows with and without uncertainty guidance.}
\label{tab:annotation_workflow_comparison}
\footnotesize
\setlength{\tabcolsep}{4pt}
\renewcommand{\arraystretch}{1.1}
\begin{tabular}{p{3.2cm} p{4.2cm} p{4.2cm}}
\toprule
\textbf{Aspect} & \textbf{Manual} & \textbf{Uncertainty-assisted} \\
\midrule
Initial state
& Blank or uniform mask
& Model-generated pseudo-label \\

Annotation mode
& Labeling from scratch
& Correcting predictions \\

Edit focus
& Entire scene
& High-uncertainty regions \\

Error localization
& Visual inspection only
& Uncertainty map assisted \\

Low-uncertainty areas
& Manually verified
& Conditionally accepted \\

Annotator load
& High
& Reduced \\
\bottomrule
\end{tabular}
\end{table}

%% file: tex_of_fig_tab/fig_mangrove3d_back-projection.tex
\begin{figure}[htpb]
    \centering

    \begin{subfigure}{\textwidth}
        \centering
        \includegraphics[width=\textwidth]{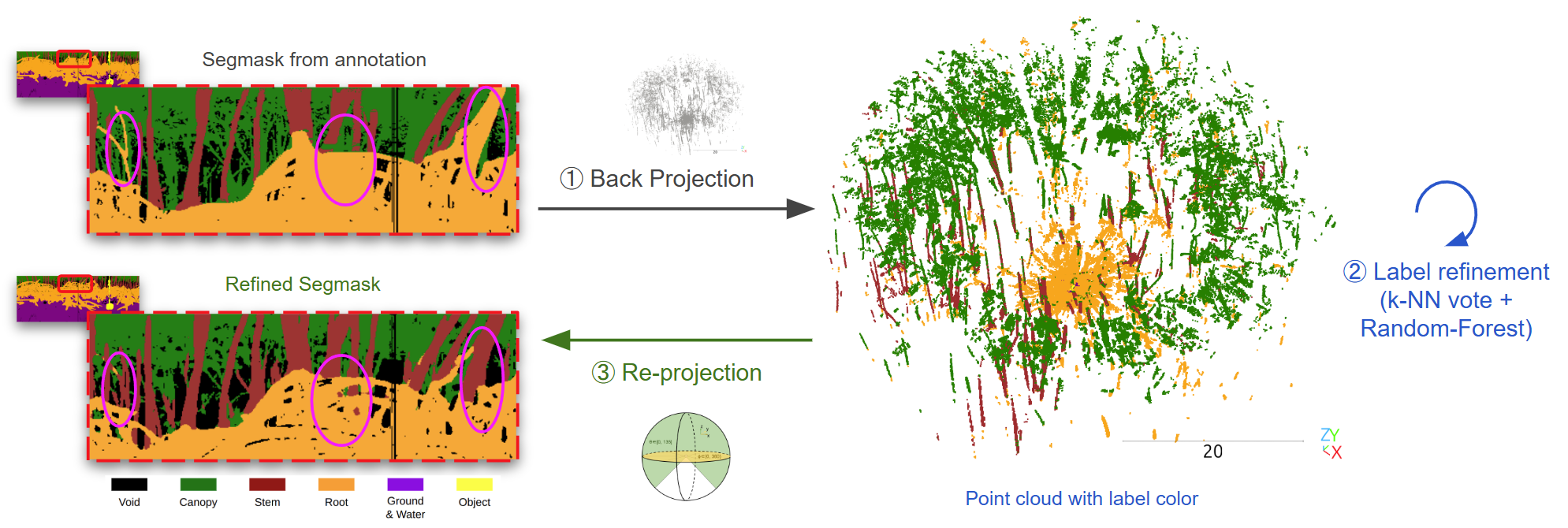}
        \caption{Workflow overview of stage 3.}
        \label{fig:backproj-a}
    \end{subfigure}

    \vspace{0.8em} 

    \begin{subfigure}{0.32\textwidth}
        \centering
        \includegraphics[width=\textwidth]{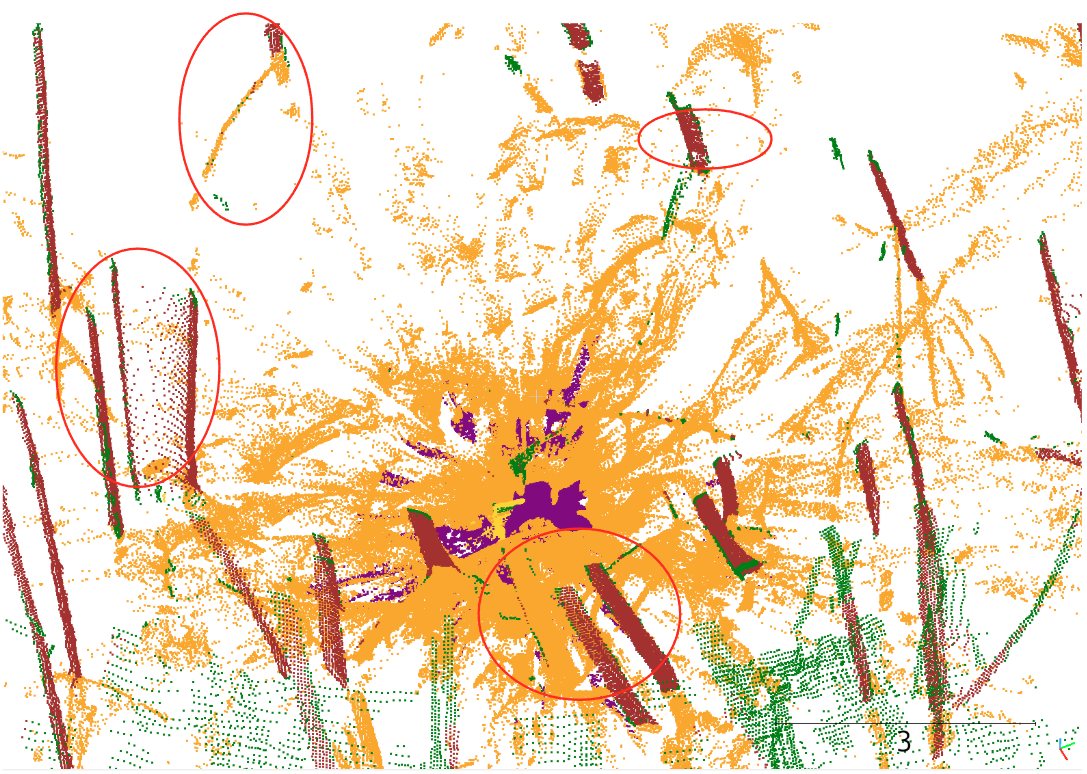}
        \caption{Before refinement.}
        \label{fig:backproj-b}
    \end{subfigure}
    \hfill
    \begin{subfigure}{0.32\textwidth}
        \centering
        \includegraphics[width=\textwidth]{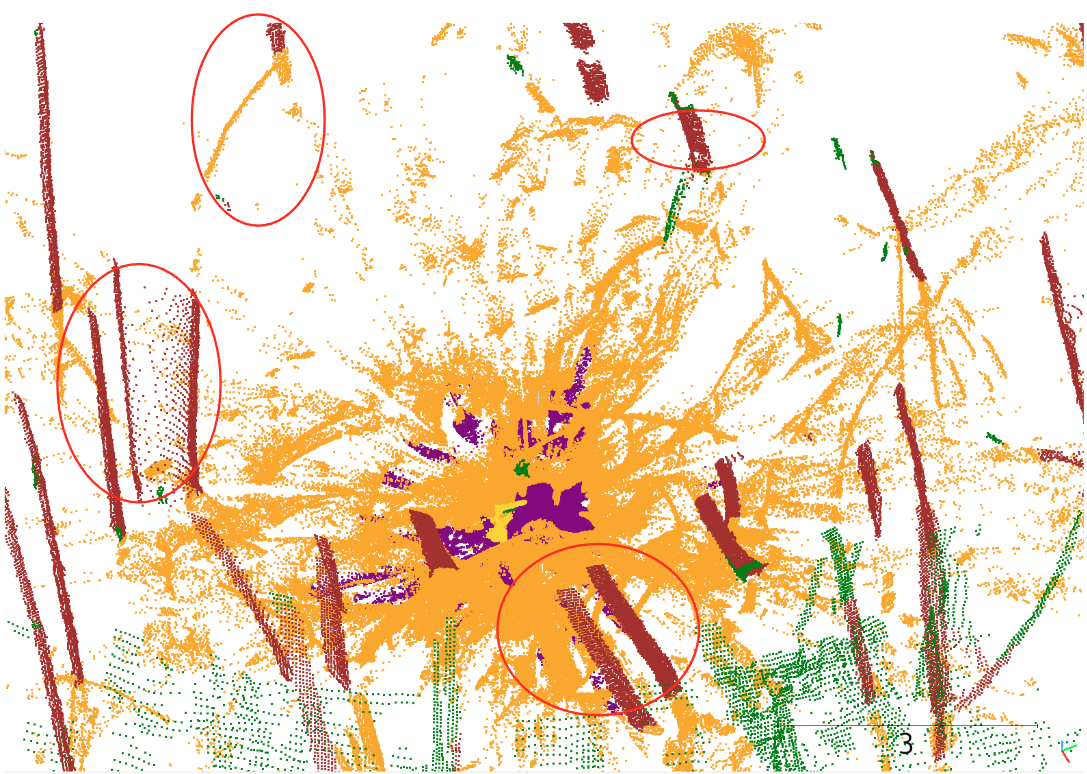}
        \caption{After refinement.}
        \label{fig:backproj-c}
    \end{subfigure}
    \hfill
    \begin{subfigure}{0.32\textwidth}
        \centering
        \includegraphics[width=\textwidth]{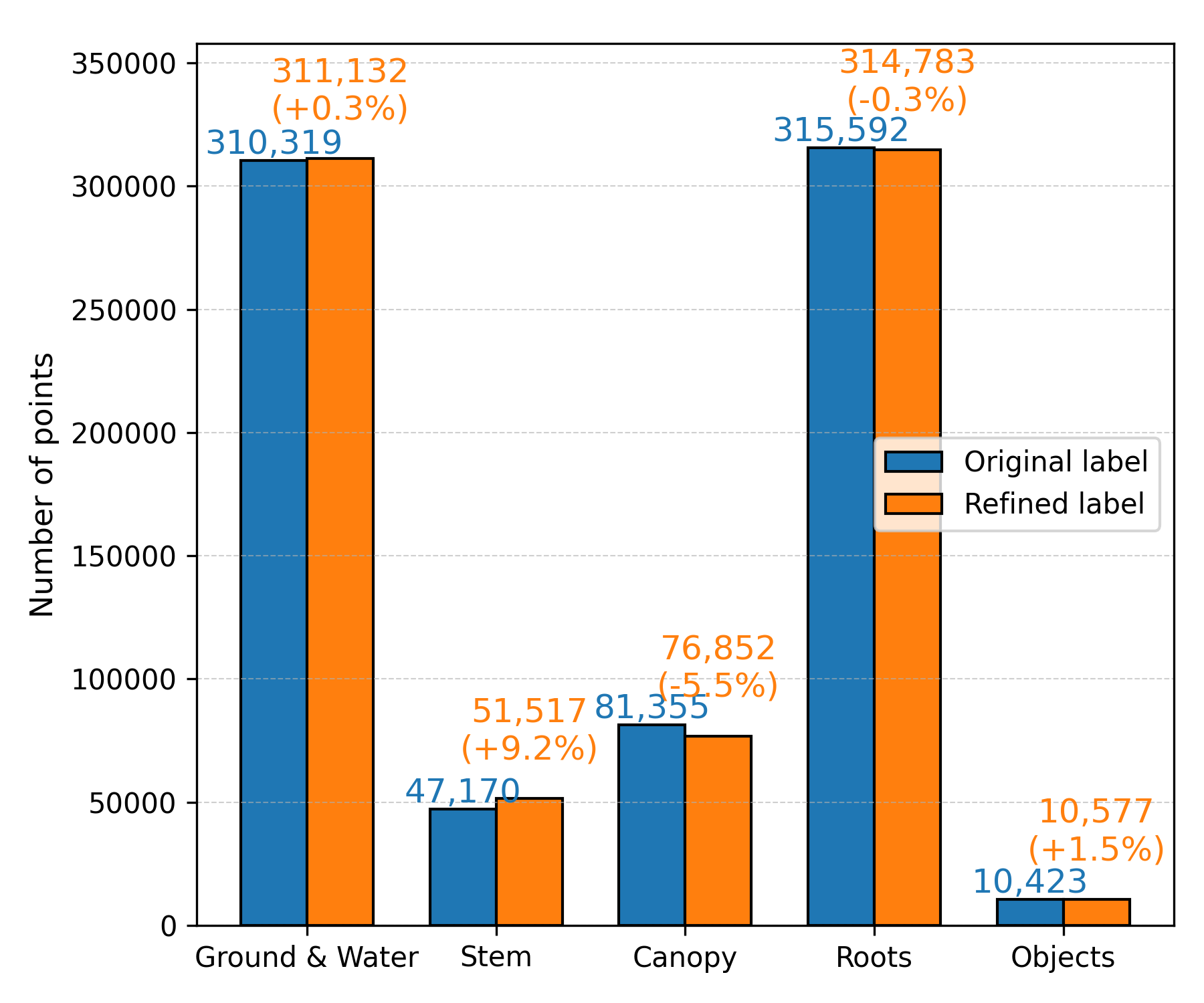}
        \caption{Label histograms.}
        \label{fig:backproj-d}
    \end{subfigure}

    \caption{Stage-3 back-projection and 3D label refinement. (a) Overall workflow connecting 2D spherical annotations to the 3D point cloud via back-projection and re-projection.
    (b-c) Example point clouds before and after 3D refinement, showing reduced isolated label noise and improved boundary consistency.
    (d) Class-wise label distributions before and after refinement, indicating that local corrections are achieved while the global class balance is preserved.}
    \label{fig:back-projection-illus}
\end{figure}

%% file: tex_of_fig_tab/fig_colorized_pcd_mangrove3d.tex
\begin{figure}[htbp]
    \centering
    \begin{minipage}[b]{0.32\textwidth}
        \begin{subfigure}[b]{\textwidth}
            \centering
            \includegraphics[width=\textwidth]{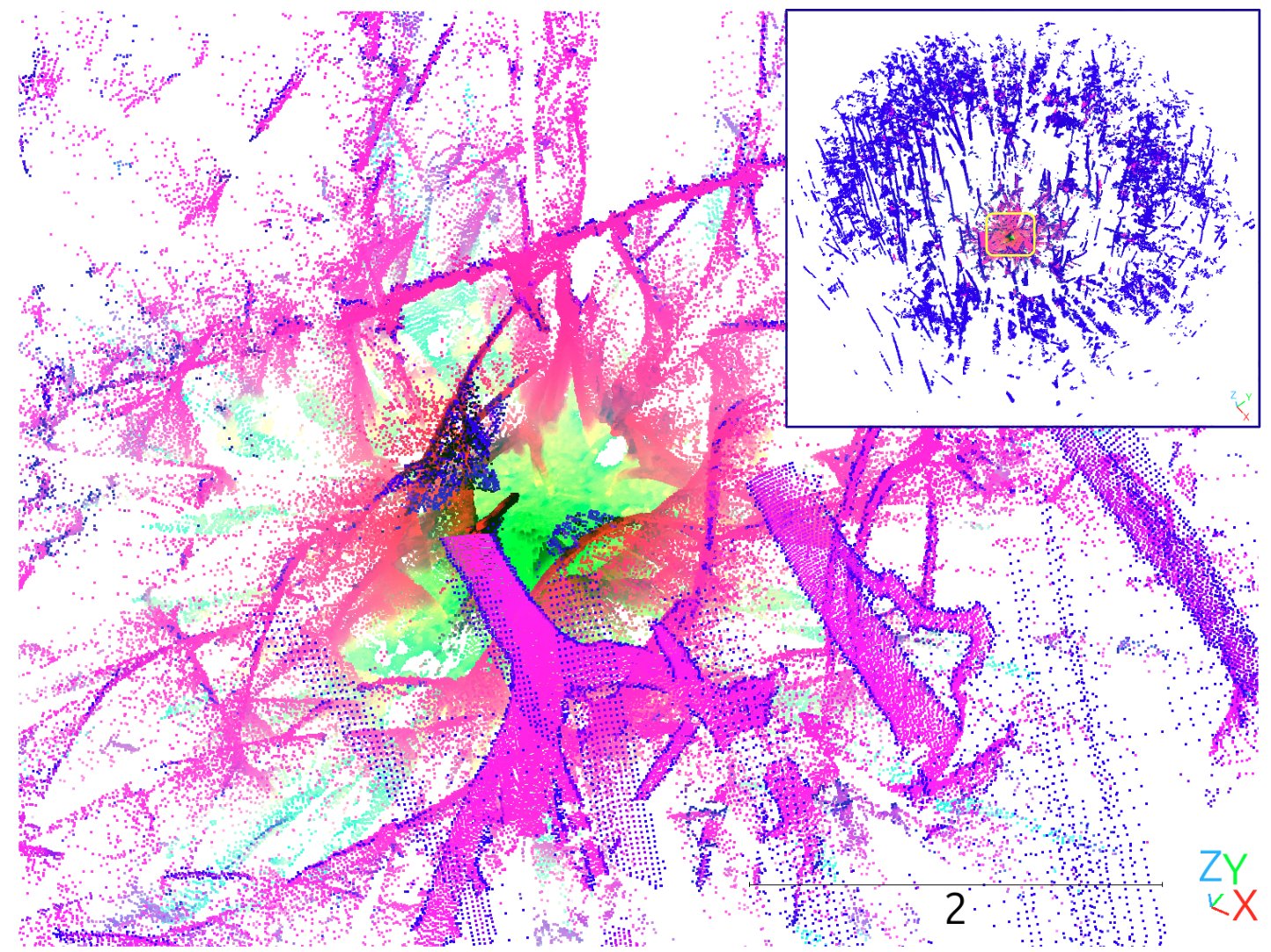}
            \caption*{(a) I.R.Z stack}
        \end{subfigure}
    \end{minipage}
    \hfill
    \begin{minipage}[b]{0.32\textwidth}
        \begin{subfigure}[b]{\textwidth}
            \centering
            \includegraphics[width=\textwidth]{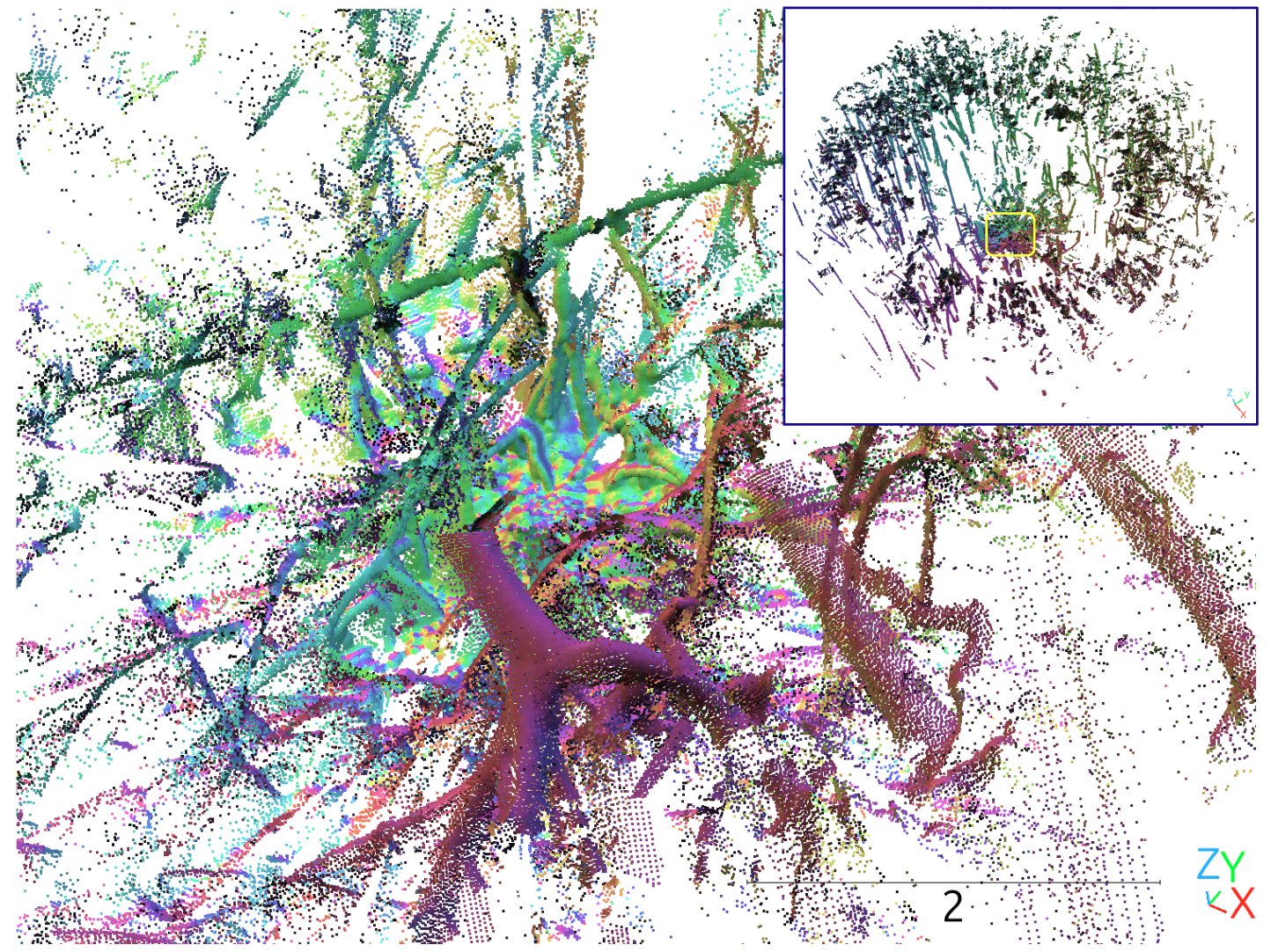}
            \caption*{(b) Normals}
        \end{subfigure}
    \end{minipage}
    \begin{minipage}[b]{0.32\textwidth}
        \begin{subfigure}[b]{\textwidth}
            \centering
            \includegraphics[width=\textwidth]{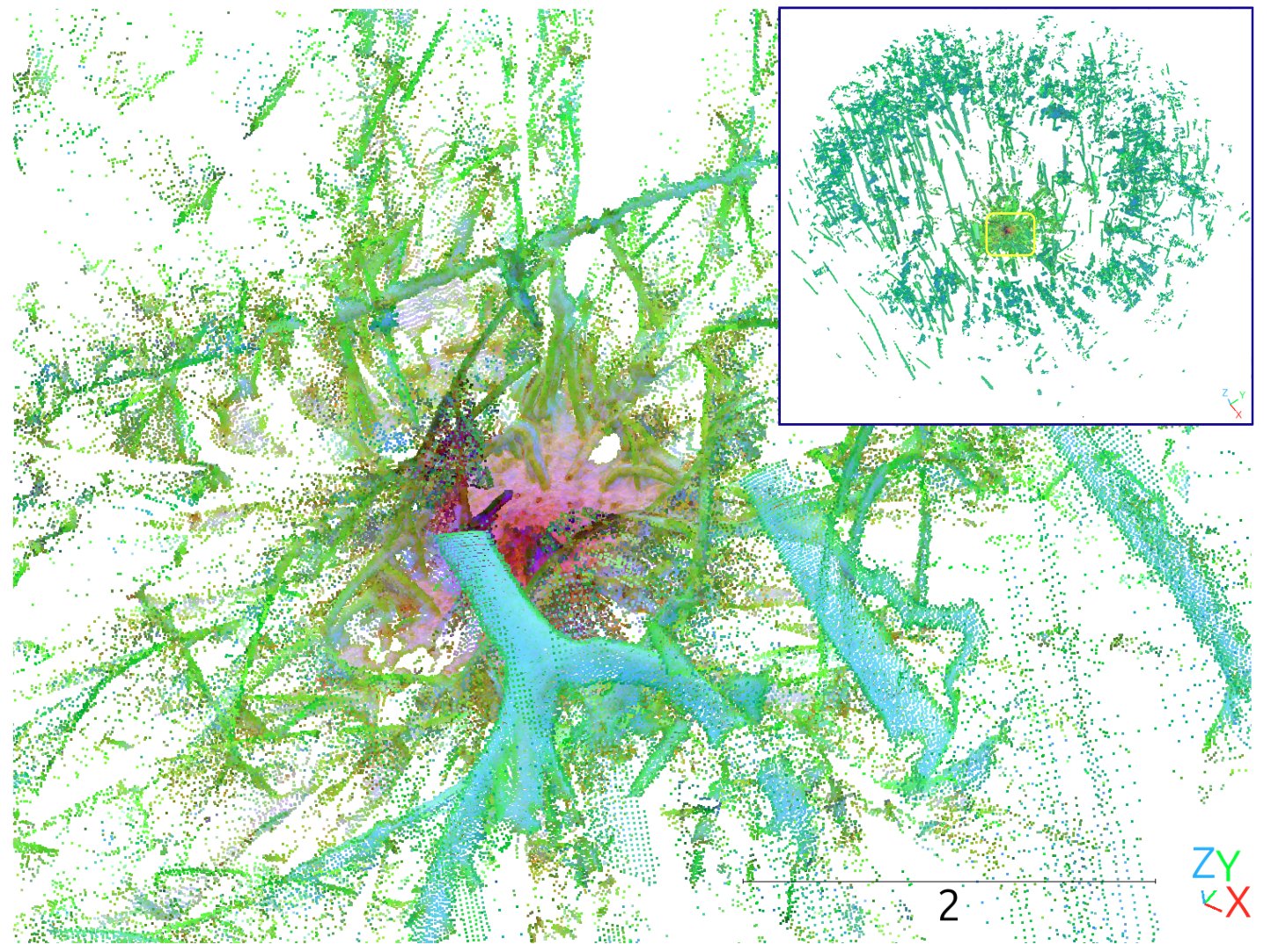}
            \caption*{(c) PCA}
        \end{subfigure}
    \end{minipage}

    \caption{Colorized point cloud from (a) stack of preprocessed intensity-range-z value; (b) pseudo-color from normals; and (c) first three components of PCA. Animation available in~\ref{app:animations}-Table~\ref{tab:animations}}
    \label{fig:colorized_pcd}
\end{figure}

%% file: tex_of_fig_tab/fig_mangrove3d_balls.tex
\begin{figure}[!htbp]
    \centering
    \begin{minipage}[b]{0.32\textwidth}
        \begin{subfigure}[b]{\textwidth}
            \centering
            \includegraphics[width=\textwidth]{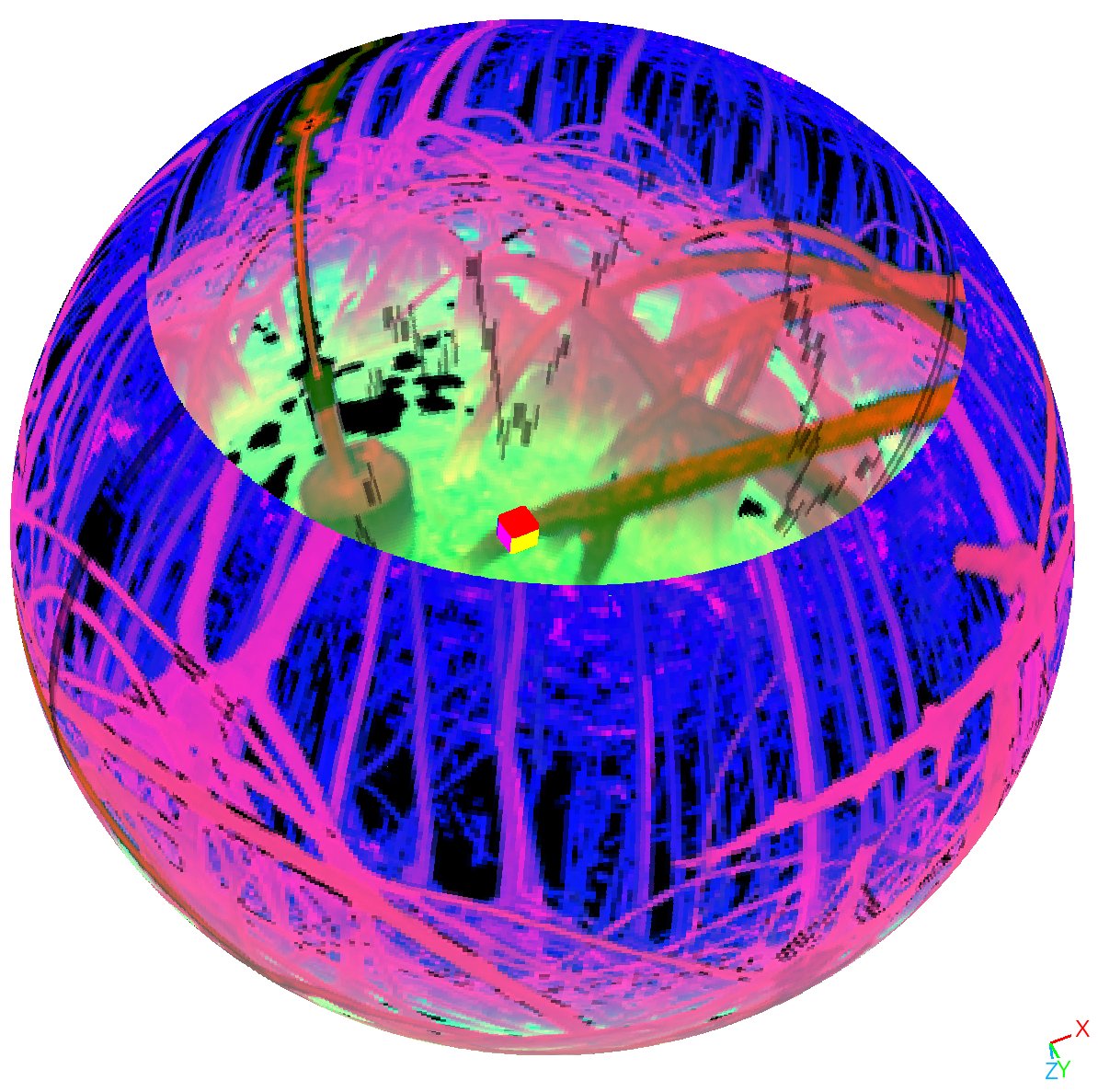}
            \caption*{(a) I.R.Z stack}
        \end{subfigure}
    \end{minipage}
    \hfill
    \begin{minipage}[b]{0.32\textwidth}
        \begin{subfigure}[b]{\textwidth}
            \centering
            \includegraphics[width=\textwidth]{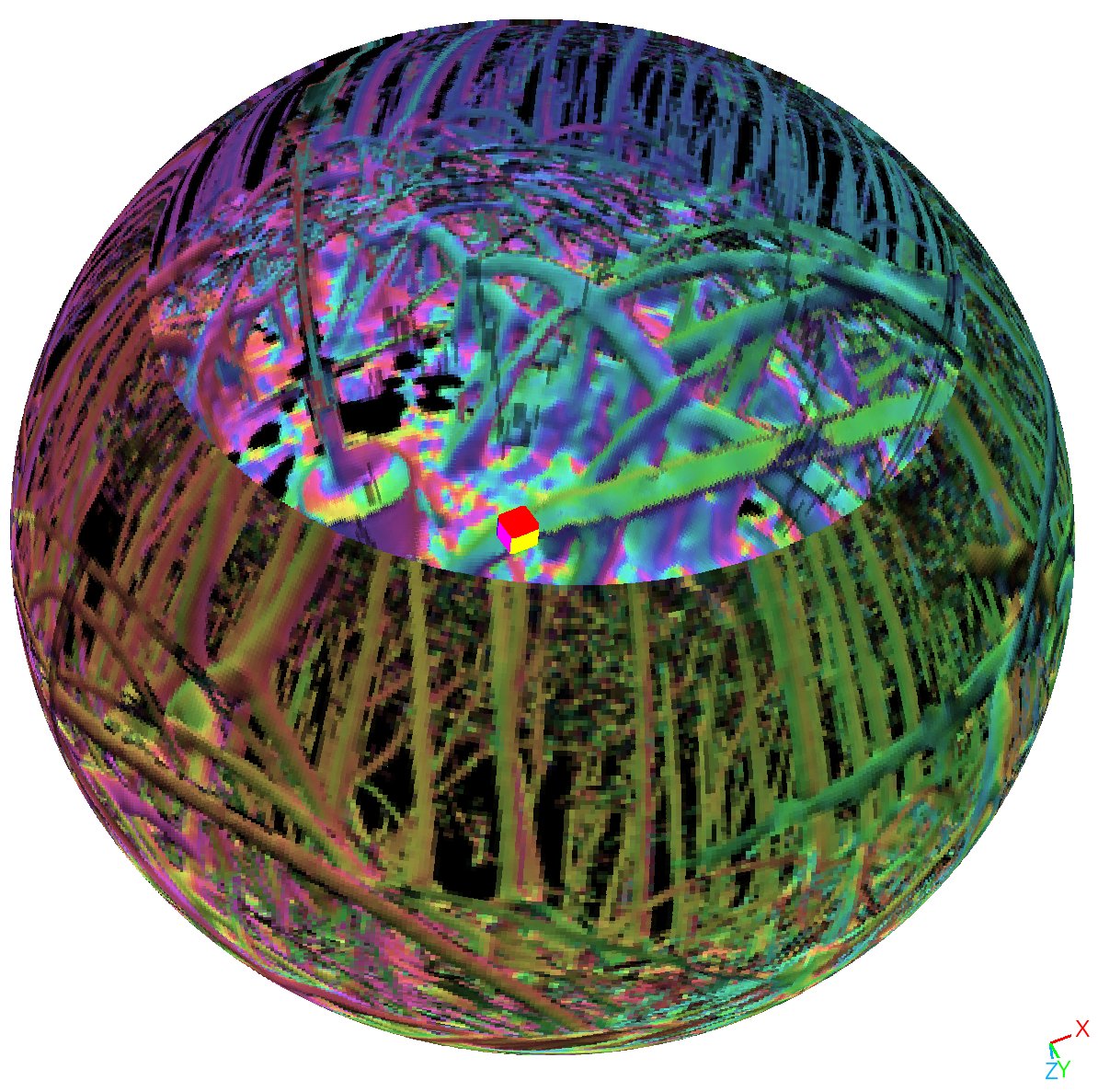}
            \caption*{(b) pseudo-color from normals}
        \end{subfigure}
    \end{minipage}
    \begin{minipage}[b]{0.32\textwidth}
        \begin{subfigure}[b]{\textwidth}
            \centering
            \includegraphics[width=\textwidth]{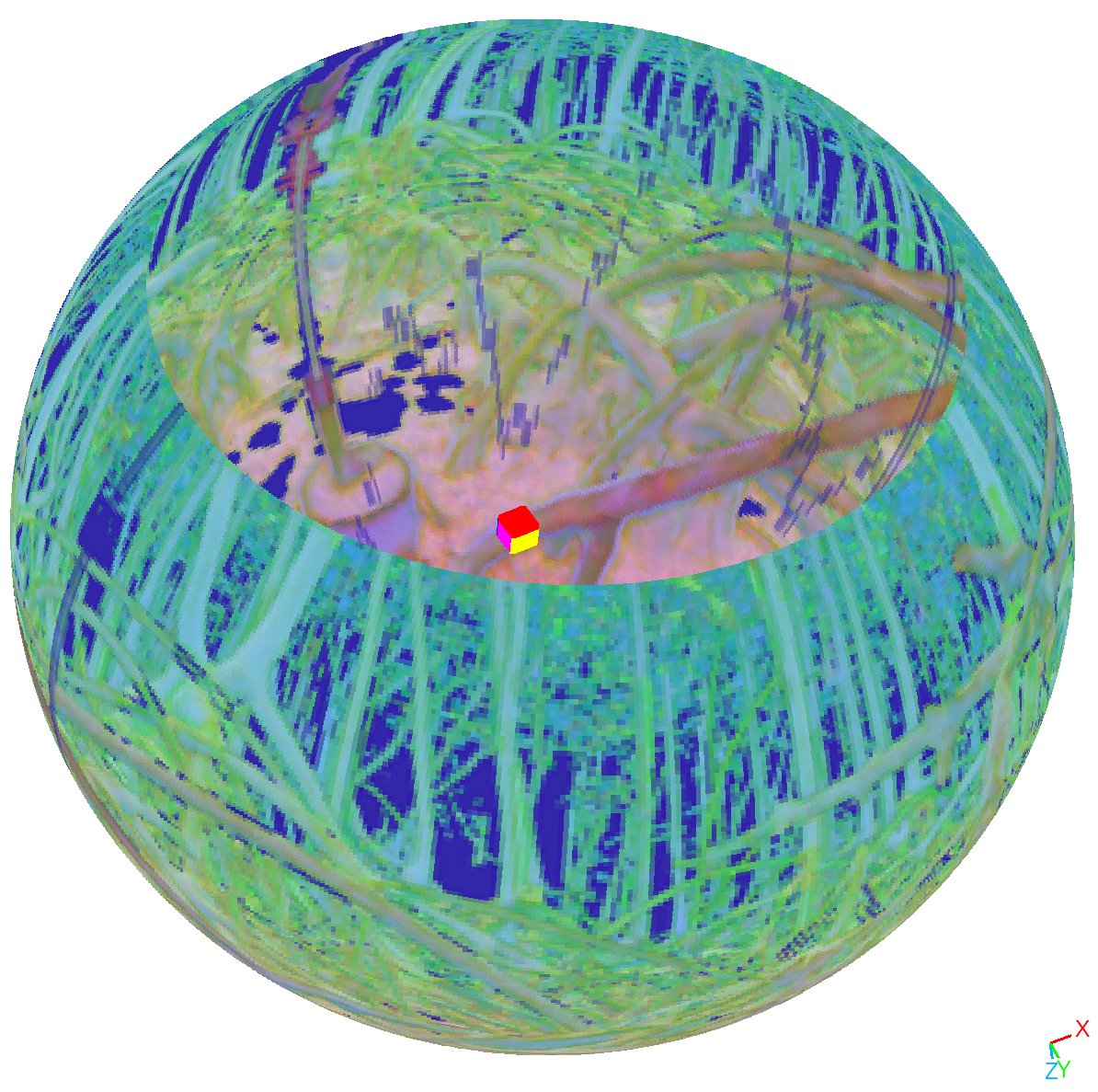}
            \caption*{(c) PCA}
        \end{subfigure}
    \end{minipage}
    
    \caption{Colorized virtual spheres from different feature stacks. Animation available in~\ref{app:animations}-Table~\ref{tab:animations}}
    \label{fig:mangrove_3d_balls}
\end{figure}

%% file: tex_of_fig_tab/tab_eval_metrics.tex
\begin{table}[htbp]
    \centering
    \scriptsize 
    \setlength{\tabcolsep}{3pt} 
    \renewcommand{\arraystretch}{1.1} 
    \begin{threeparttable}
    \caption{Segmentation accuracy and uncertainty evaluation metrics.\tnote{1}}
    \label{tab:eval-metrics}
    \begin{tabular}{@{}P{2.2cm}P{2.8cm}P{4.2cm}P{4.5cm}@{}}
    \toprule
    \textbf{Group} & \textbf{Metric} & \textbf{Definition / Equation} & \textbf{Interpretation} \\
    \midrule
    \multirow{4}{*}{\parbox{2cm}{\centering\textbf{Pseudo-label accuracy}}}
     & Overall Accuracy (oAcc) &
       \(\tfrac{\sum_c \mathrm{TP}_c}{\sum_c (\mathrm{TP}_c+\mathrm{FP}_c+\mathrm{FN}_c)}\) &
       Fraction of correctly labelled pixels. \\
     & Mean Accuracy (mAcc) &
       \(\tfrac{1}{C}\sum_{c=1}^{C}\tfrac{\mathrm{TP}_c}{\mathrm{TP}_c+\mathrm{FN}_c}\) &
       Average per-class recall (balanced). \\
     & Per-class IoU (\(\mathrm{IoU}_c\)) &
       \(\tfrac{\mathrm{TP}_c}{\mathrm{TP}_c+\mathrm{FP}_c+\mathrm{FN}_c}\) &
       Overlap between prediction and truth for class \(c\). \\
     & Mean IoU (mIoU) &
       \(\tfrac{1}{C}\sum_{c=1}^{C}\mathrm{IoU}_c\) &
       Mean overlap across classes. \\
    \midrule
    \multirow{2}{*}{\parbox{2cm}{\centering\textbf{Uncertainty maps}}}
     & Shannon Entropy (\(H\)) &
       \(H(\mathbf{p})=-\sum_{k=1}^{K} p_k\log p_k\) &
       Info content of error/uncertainty map; higher \(H\) = more dispersed. \\
     & Area Under PR Curve (AUPRC) &
       \(\sum_{j=1}^{J-1} (\mathcal{R}_{j+1}-\mathcal{R}_{j})\,\mathcal{P}_{j+1}\) &
       Ability of an uncertainty map to highlight errors  
       (\(1=\) perfect, \(0=\) random). \\
    \bottomrule
    \end{tabular}
    
    \vspace{2pt}
    \begin{tablenotes}[flushleft]
        \item[1] \textbf{Notation.}  
        \(\mathrm{TP}_c\), \(\mathrm{FP}_c\), \(\mathrm{FN}_c\): true-, false-positive, and false-negative pixels; \(C\): number of classes.  
        \(\mathbf{p}=[p_1,\dots,p_K]\): histogram of map values binned into \(K\) bins.  
        \(\mathcal{P}_j\), \(\mathcal{R}_j\): precision and recall at the \(j^{\text{th}}\) threshold.
    \end{tablenotes}
    \end{threeparttable}
\end{table}

%% file: tex_of_fig_tab/fig_mangrove3d-eval-results-ch345.tex
    \begin{figure}
        \centering
        \includegraphics[width=\linewidth]{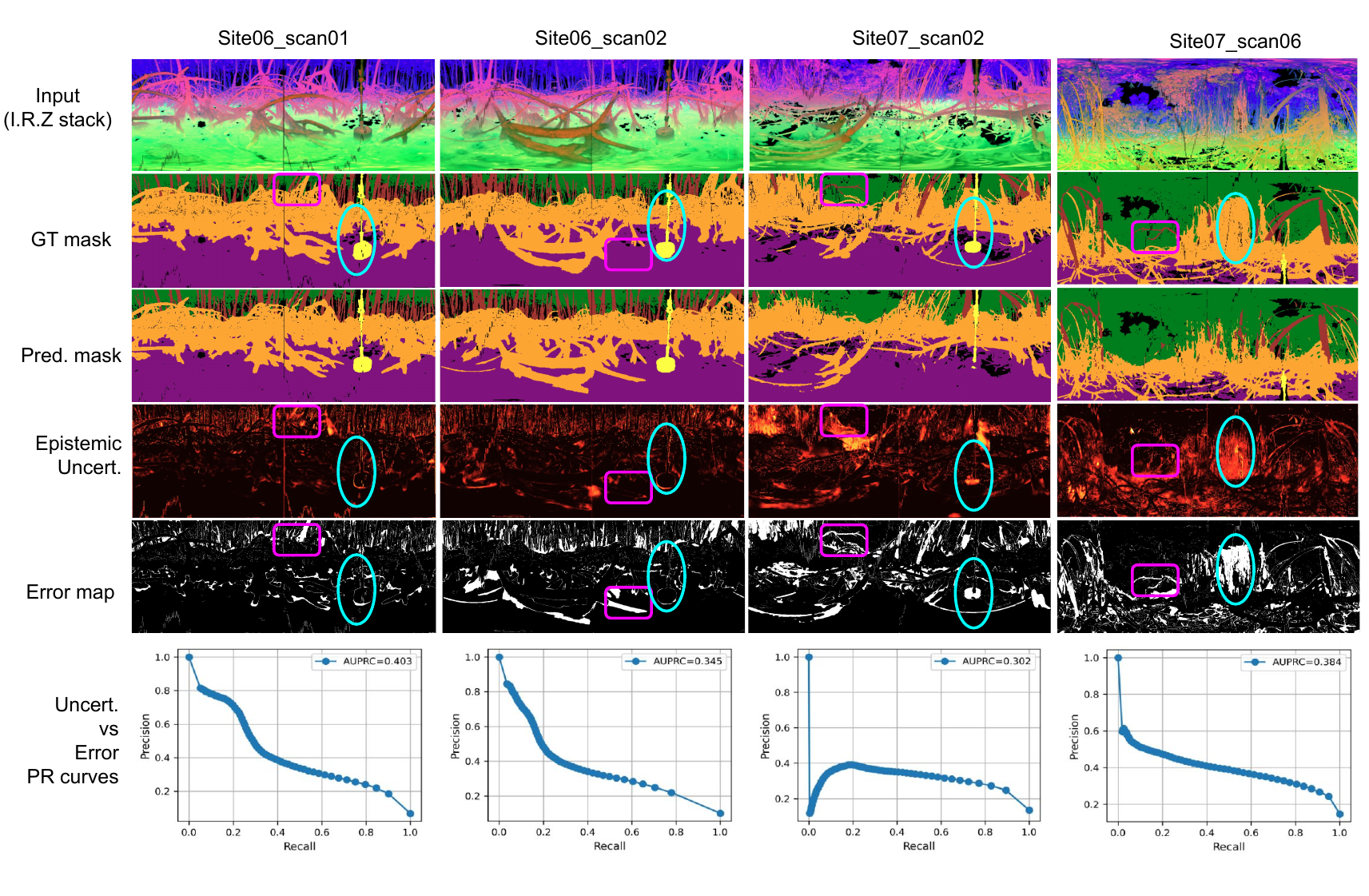}
        \caption{\textbf{Qualitative comparison on four representative test scans.}
        Each column shows the input and prediction outputs of a scan. From top to bottom: (1) preprocessed Intensity–Range–Z (I.R.Z) input stack, (2) ground-truth mask, (3) predicted mask, (4) epistemic-uncertainty map (values rescaled to [0, 1] and rendered with the same \textit{hot} color map as Fig. \ref{subfig:pu-eps-uncert}; brighter red = higher uncertainty), (5) binary error map obtained by the pixel-wise XOR between GT and prediction, and (6) precision–recall (PR) curve measuring how well the uncertainty map localizes errors (AUPRC reported in legend). \mismatchPurp{Purple} rectangles highlight regions where high uncertainty does not coincide with errors;  \matchCyan{cyan} circles mark areas of good agreement between uncertainty and error maps.}
        \label{fig:mangrove3d-eval-results-ch345}
    \end{figure}

%% file: tex_of_fig_tab/fig_feature_ablation.tex
\begin{figure}[ht]
  \centering
  \begin{subfigure}[b]{0.44\textwidth}
      \centering
      \includegraphics[width=\linewidth]{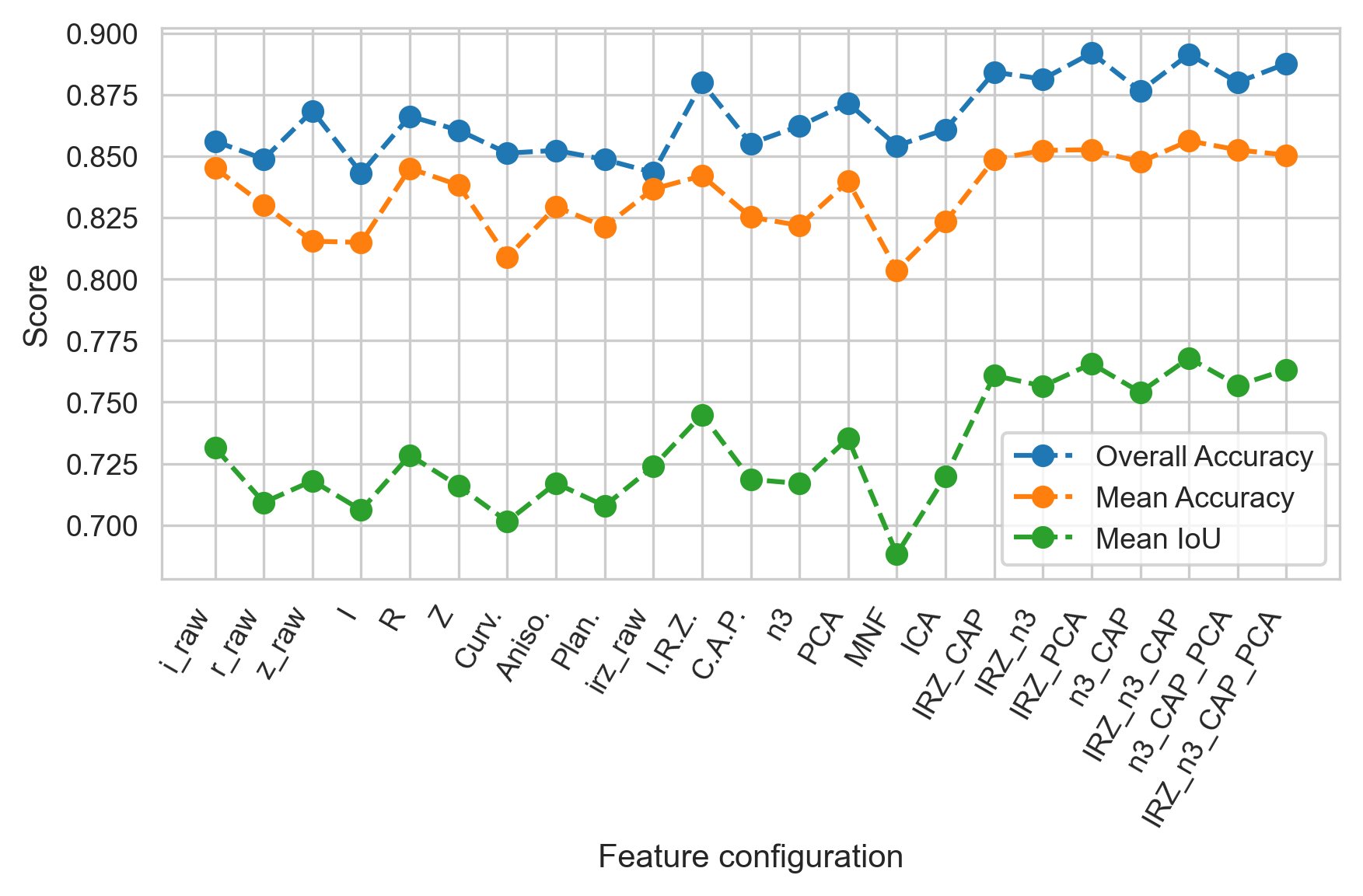}
      \caption{Global segmentation metrics across feature configurations.}
      \label{subfig:feat-ablation-global}
  \end{subfigure}
  \hfill
  \begin{subfigure}[b]{0.54\textwidth}
      \centering
      \includegraphics[width=\linewidth]{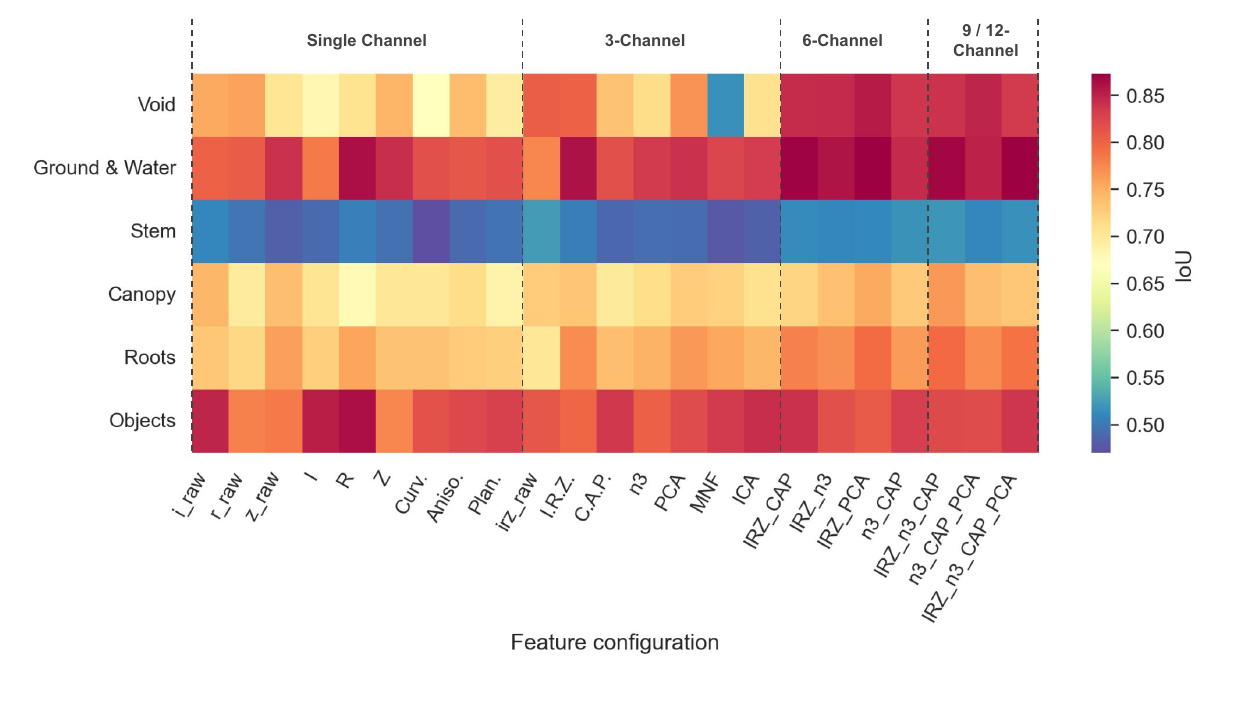}
      \caption{Per-class IoU heat-map.}
      \label{subfig:feat-ablation-heatmap}
  \end{subfigure}

  \caption{Effect of feature enrichment on segmentation performance. (a) Evolution of global accuracy metrics. (b) Class-specific IoU across the same feature sets, with dashed lines separating single-channel, 3-channel, 6-channel, and 9/12-channel groups.}
  \label{fig:feat-ablation-mangrove3d}
\end{figure}

%% file: tex_of_fig_tab/tab_model_complexity.tex
\begin{table*}[htbp]
\centering
\caption{Computational cost and segmentation performance of evaluated models using the Intensity--Range--Z (IRZ; 3-channel) and Intensity--Range--Z with surface normals (IRZ\_N3; 6-channel) feature groups.}
\label{tab:model_comparison}
\resizebox{\textwidth}{!}{
\begin{tabular}{llcccccccc}
\toprule
Feature group & Model & Params (M) & GFLOPs & Train time / epoch (s) & Peak GPU mem (MB) & Best epoch & Peak GPU mem (GB) & Inference time (s) & mIoU \\
\midrule
\multirow{4}{*}{IRZ}
& UNet++          & 26.08 & 53.97 & 16.04 & 7313.7  & 37 & 7.14 & 0.034 & 0.7448 \\
& DeepLabV3+      & 11.68 & 8.85  & 22.77 & 9645.9  & 67 & 9.42 & 0.045 & 0.6937 \\
& SegFormer       & 13.68 & 10.85 & 15.84 & 4649.5  & 52 & 4.54 & 0.026 & 0.6893 \\
& \textbf{Ensemble}        & \textbf{--}    & \textbf{--}    & \textbf{--}    & \textbf{--}      & \textbf{--} & \textbf{--}   & \textbf{0.082} & \textbf{0.7599} \\
\midrule
\multirow{4}{*}{IRZ\_N3}
& UNet++          & 53.70 & 155.69 & 20.55 & 13960.0 & 66 & 13.63 & 0.064 & 0.7646 \\
& DeepLabV3+      & 22.88 & 9.07   & 19.92 & 18420.8 & 55 & 17.99 & 0.055 & 0.7175 \\
& SegFormer       & 27.09 & 18.49  & 19.52 & 8452.4  & 28 & 8.25  & 0.043 & 0.7029 \\
& \textbf{Ensemble}        & \textbf{--}    & \textbf{--}     & \textbf{--}    & \textbf{--}      & \textbf{--} & \textbf{--}    & \textbf{0.144} & \textbf{0.7766} \\
\bottomrule
\end{tabular}
}
\vspace{2pt}
\begin{minipage}{\textwidth}
\fontsize{7}{8}\selectfont
\raggedright
\textbf{Notes:} GFLOPs denote the theoretical number of floating-point operations per forward pass, estimated at an input resolution of $544 \times 352 \times ch$. Runtime measurements use a batch size of 10.
Inference time is reported per TLS scan.
All models were trained and evaluated on an HPC system using a single NVIDIA A100 GPU, with 20 GB of system memory requested per SLURM job.
Peak GPU memory refers to the maximum allocated memory observed during training.
Best epoch corresponds to early stopping triggered with a patience of five epochs based on validation performance.
\end{minipage}
\end{table*}

%% file: tex_of_fig_tab/tab_pointnet2_benchmark.tex
\begin{table*}[t]
\centering
\caption{PointNet++ segmentation performance on the \textit{Mangrove3D} test set for ten incremental feature–channel configurations. Within each metric column, the top three scores are shaded (dark = best, medium = second, light = third). The \textit{XYZ} baseline is shown in light gray. “Extra feat.” = number of channels added on top of XYZ.}
\label{tab:pointnet2_benchmark}
\resizebox{0.8\textwidth}{!}{%
\setlength{\tabcolsep}{4.5pt}
\small
\begin{tabular}{@{} c l *{3}{S} *{5}{S} @{}}
\toprule
\multirow{2}{*}{\textbf{Extra}} & \multirow{2}{*}{\textbf{Feature}} &
\multicolumn{3}{c}{\textbf{Global metrics}} &
\multicolumn{5}{c}{\textbf{IoU per class}} \\
\cmidrule(lr){3-5}\cmidrule(lr){6-10}
\textbf{feat.} & \textbf{group} & {oAccu.} & {mAccu.} & {mIoU} &
{Ground} & {Stem} & {Canopy} & {Roots} & {Objects} \\
\midrule

\rowcolor{black!6}
0   & baseline-xyz                   & 0.848 & 0.741 & 0.634 & 0.851 & 0.401 & 0.663 & 0.736 & 0.516 \\

1   & xyz\_i0                        & 0.852 & 0.744 & 0.636 & 0.867 & 0.411 & 0.656 & 0.739 & 0.508 \\
3   & xyz\_IRZ                       & 0.850 & 0.760 & 0.640 & \first{0.873} & 0.384 & 0.655 & 0.733 & 0.553 \\
3   & xyz\_CAP                       & 0.859 & 0.806 & 0.689 & 0.867 & 0.424 & \third{0.676} & \first{0.748} & 0.731 \\
3   & xyz\_N3                        & 0.855 & \first{0.834} & \third{0.704} & 0.861 & \third{0.430} & 0.655 & 0.737 & \first{0.837} \\
3   & xyz\_PCA                       & 0.848 & 0.770 & 0.669 & 0.852 & 0.383 & 0.644 & 0.736 & 0.728 \\
6   & xyz\_IRZ\_N3                   & \second{0.861} & \second{0.823} & \second{0.705} & \second{0.872} & 0.424 & 0.671 & \first{0.748} & \second{0.811} \\
6   & xyz\_N3\_CAP                   & \first{0.866} & \third{0.832} & \first{0.712} & \third{0.870} & \first{0.456} & \second{0.685} & \second{0.752} & \third{0.796} \\
9   & xyz\_IRZ\_N3\_CAP              & \second{0.861} & 0.808 & 0.699 & 0.865 & \second{0.446} & \first{0.686} & 0.742 & 0.758 \\
12  & xyz\_IRZ\_N3\_CAP\_PCA         & 0.855 & 0.785 & 0.678 & 0.856 & 0.397 & 0.670 & 0.745 & 0.720 \\
\bottomrule
\end{tabular}
} 
\end{table*}

%% file: tex_of_fig_tab/fig_data_efficiency_accu.tex
\begin{figure}[!tb]
\centering
    \begin{subfigure}[b]{0.24\textwidth}
        \includegraphics[width=\textwidth]{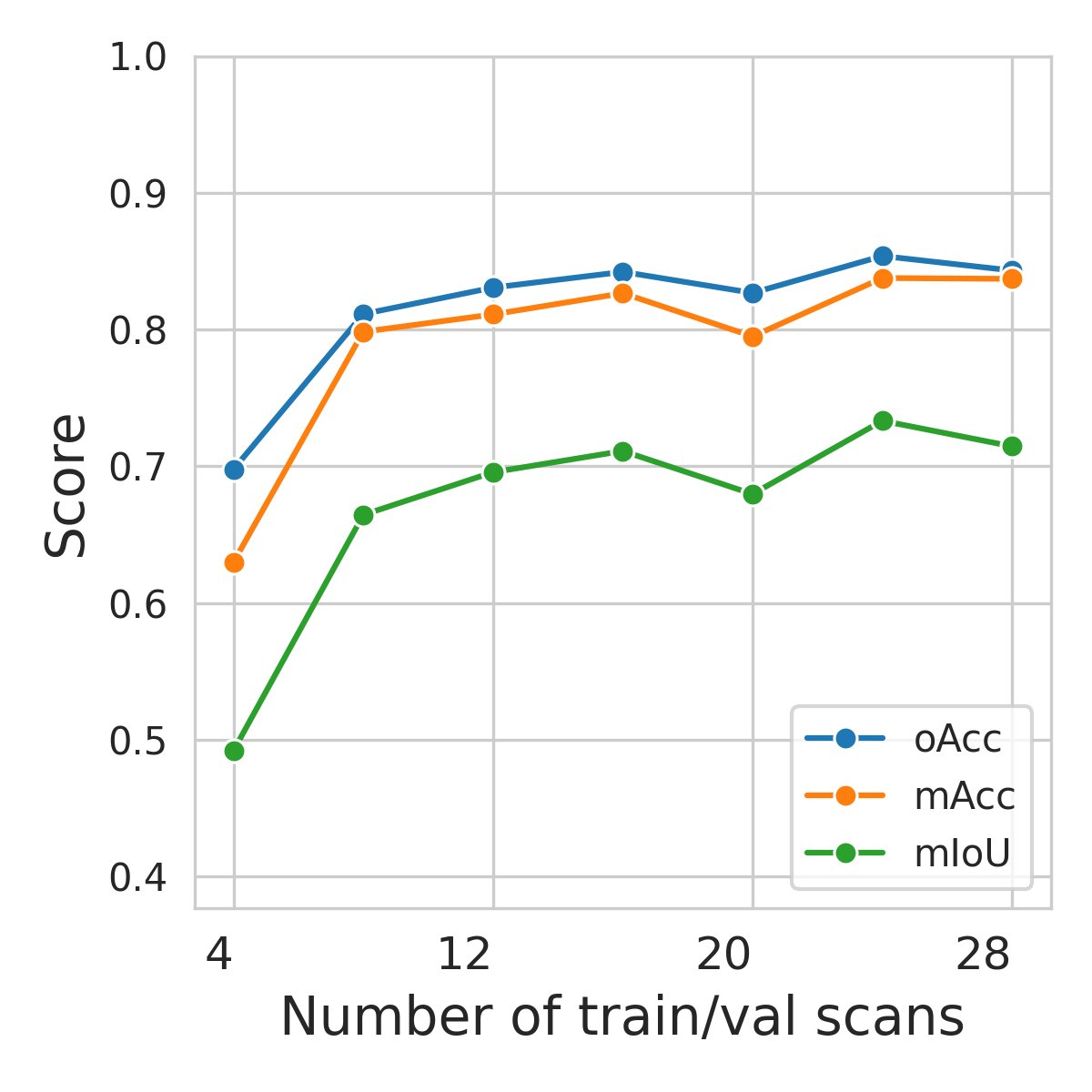}
        \caption{1-channel \\(Intensity)}
        \label{subfig:data_efficienty_accu-a}
    \end{subfigure}
    \hfill
    \begin{subfigure}[b]{0.24\textwidth}
        \includegraphics[width=\textwidth]{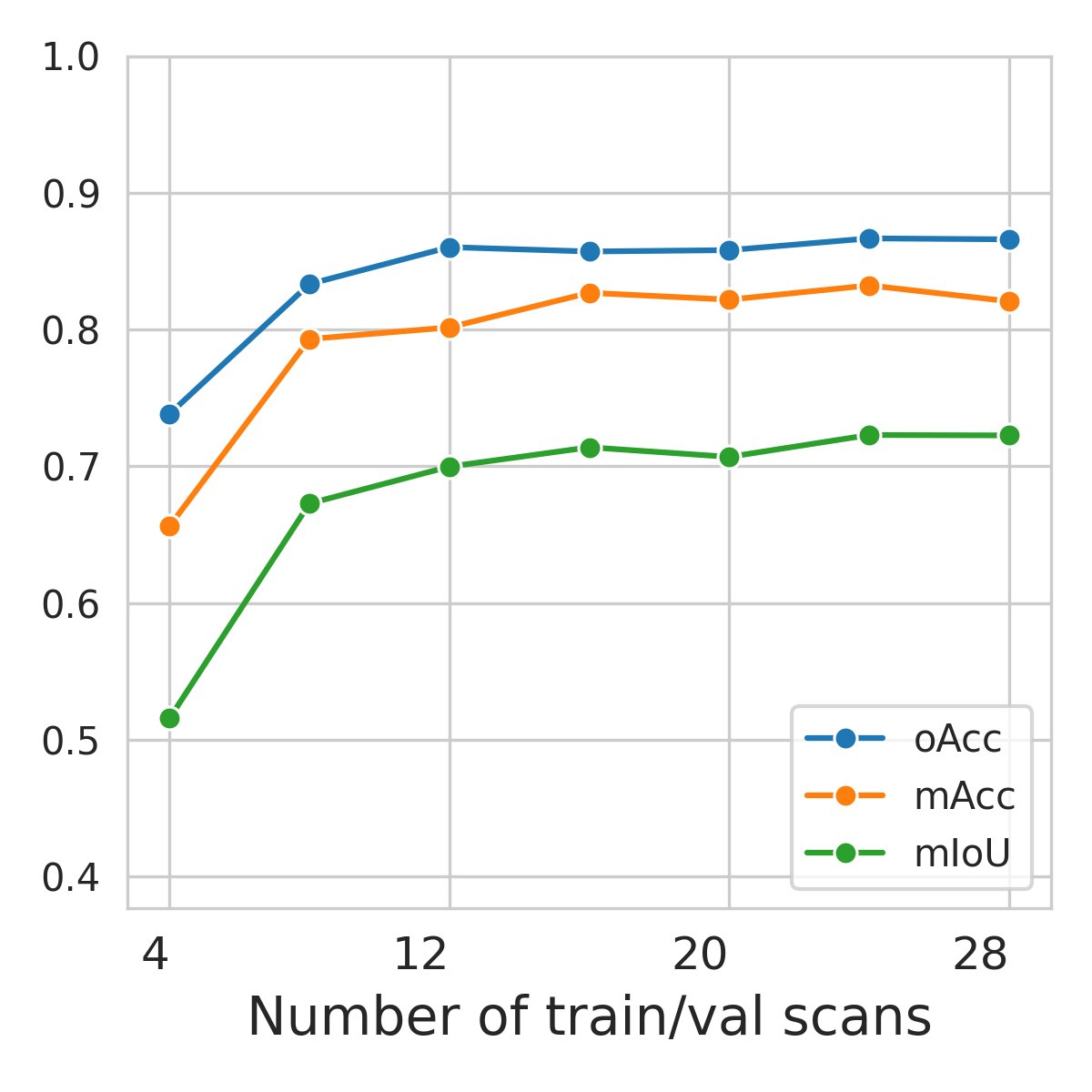}
        \caption{3-channel \\(I.R.Z.)}
        \label{subfig:data_efficienty_accu-b}
    \end{subfigure}
    \hfill
    \begin{subfigure}[b]{0.24\textwidth}
        \includegraphics[width=\textwidth]{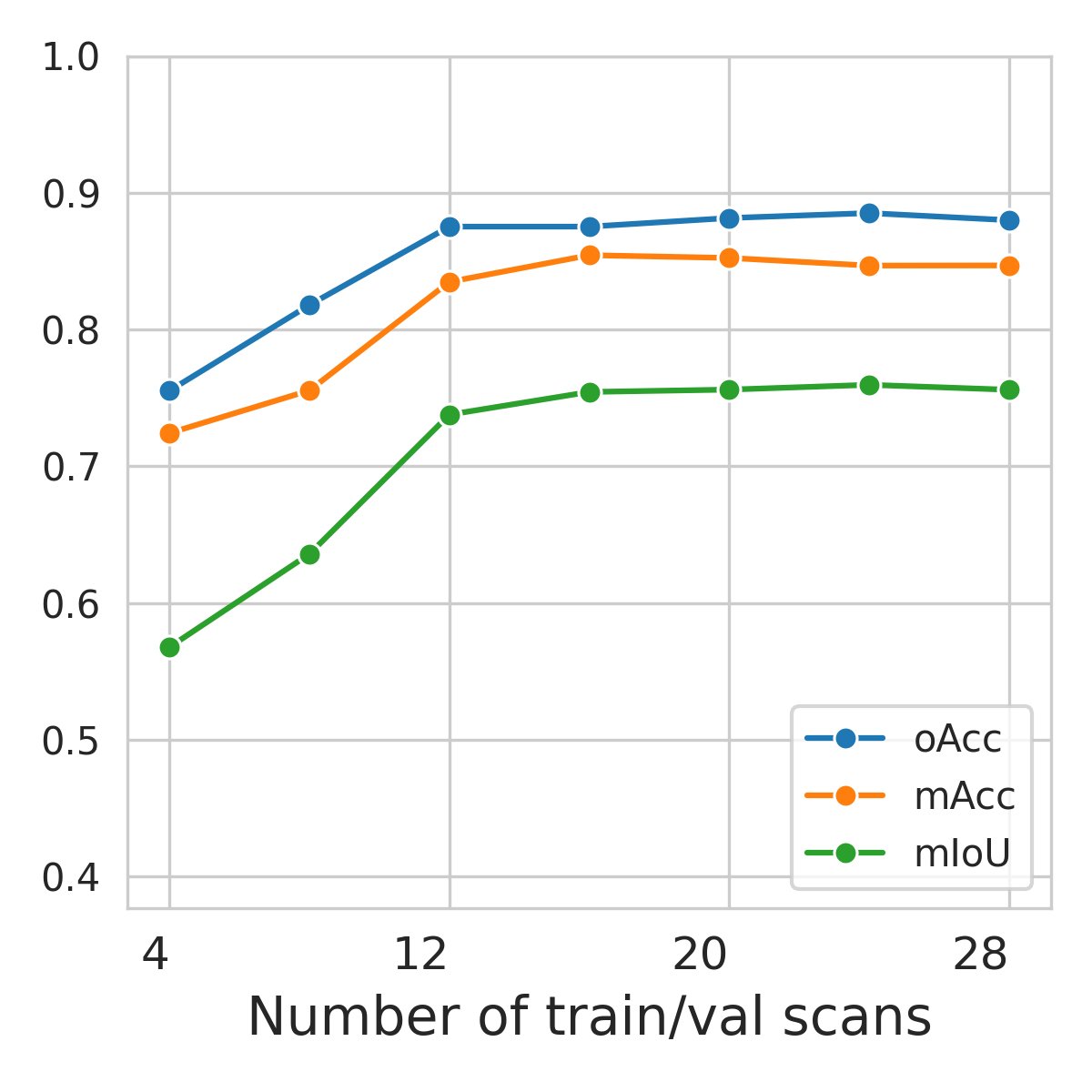}
        \caption{6-channel \\(IRZ\_CAP)}
        \label{subfig:data_efficienty_accu-c}
    \end{subfigure}
    \hfill
    \begin{subfigure}[b]{0.24\textwidth}
        \includegraphics[width=\textwidth]{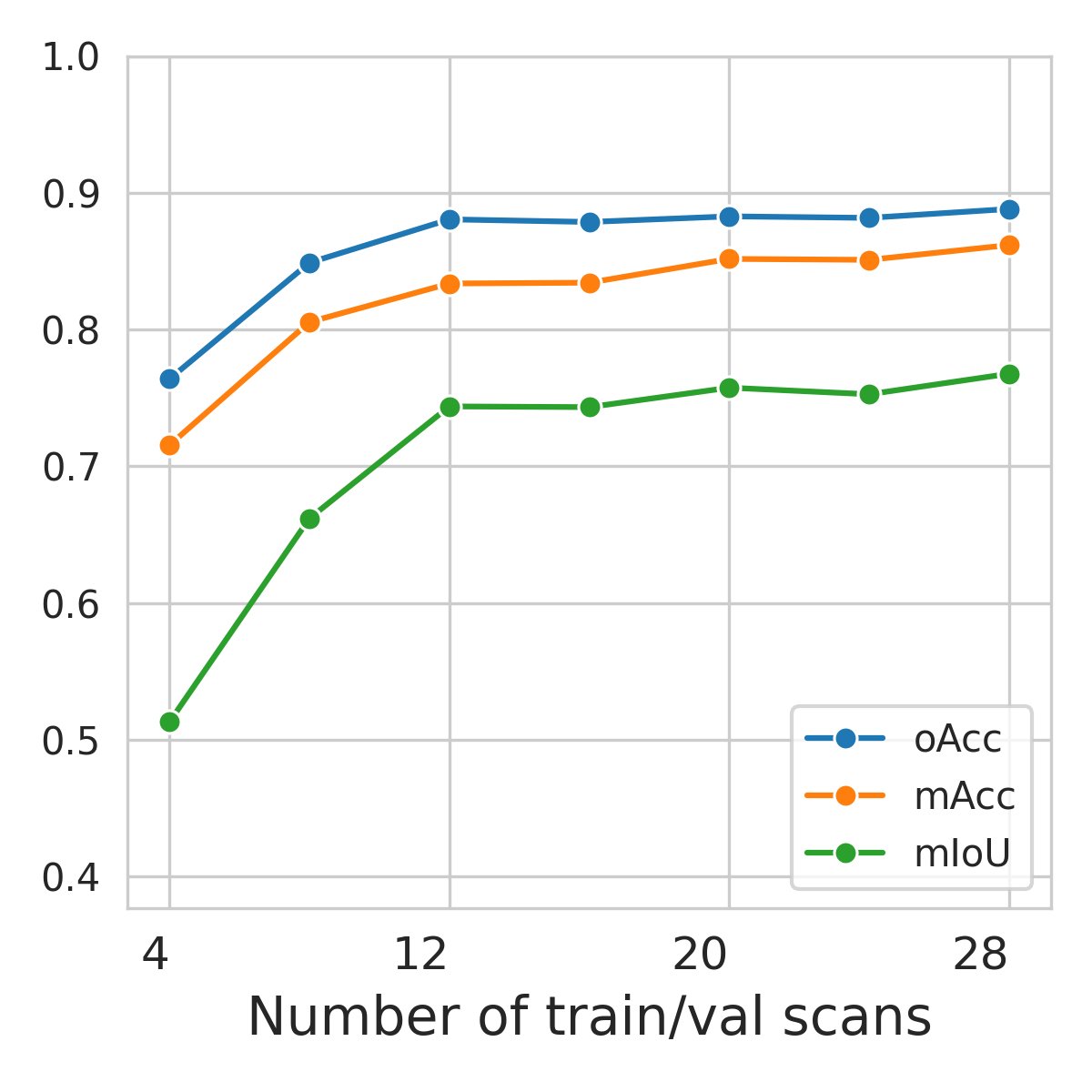}
        \caption{12-channel \\(IRZ\_N3\_CAP\_PCA)}
        \label{subfig:data_efficienty_accu-d}
    \end{subfigure}

\caption{Prediction performance of the ensemble model trained with increasing numbers of train/val scans (4–28) across four feature configurations.}
\label{fig:data_efficienty_accu}
\end{figure}

%% file: tex_of_fig_tab/fig_data_efficiency_uncert.tex
\begin{figure}[!htbp]
    \centering

    \begin{subfigure}[b]{0.24\textwidth}
        \includegraphics[width=\linewidth]{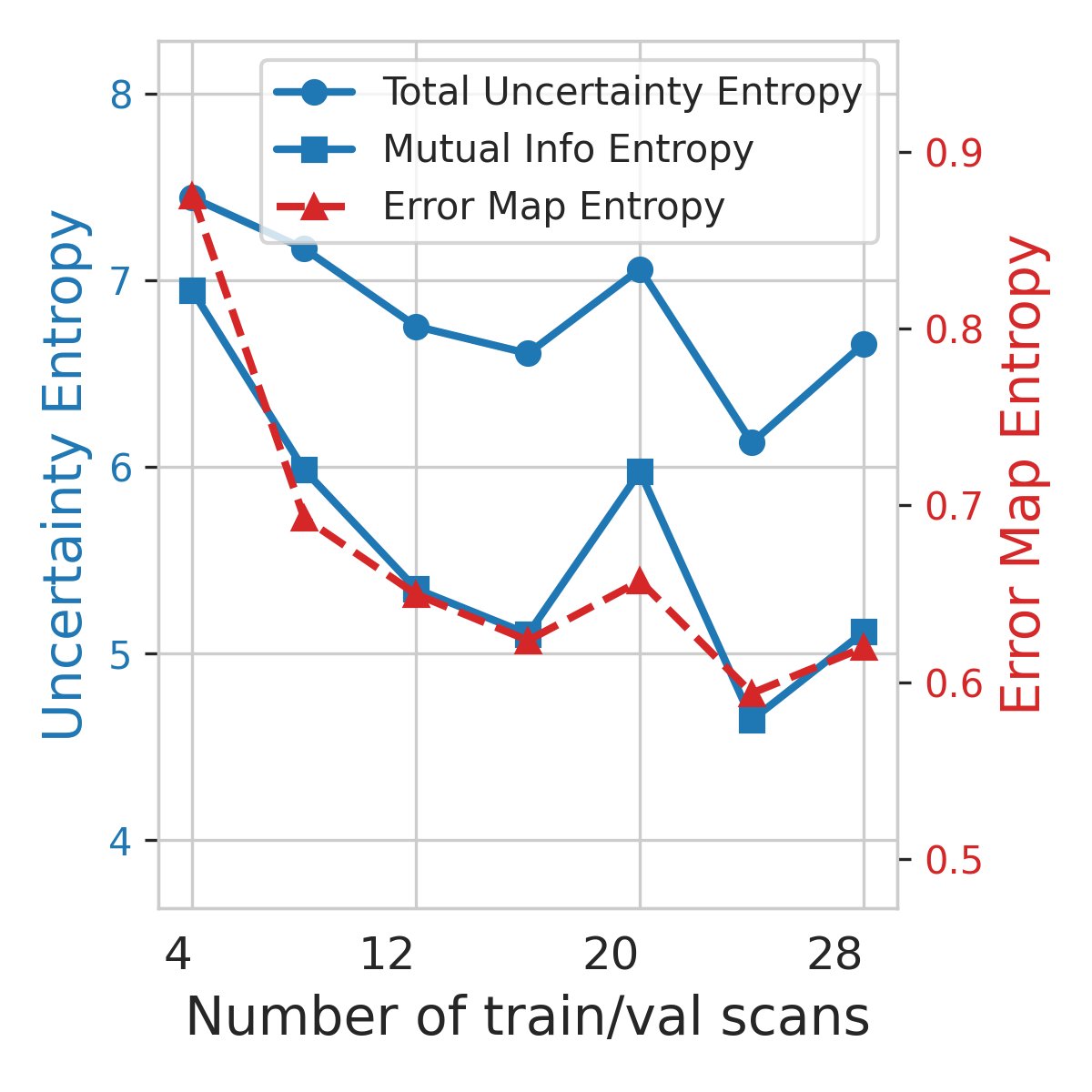}
        \caption{1-Ch}
    \end{subfigure}
    \hfill
    \begin{subfigure}[b]{0.24\textwidth}
        \includegraphics[width=\linewidth]{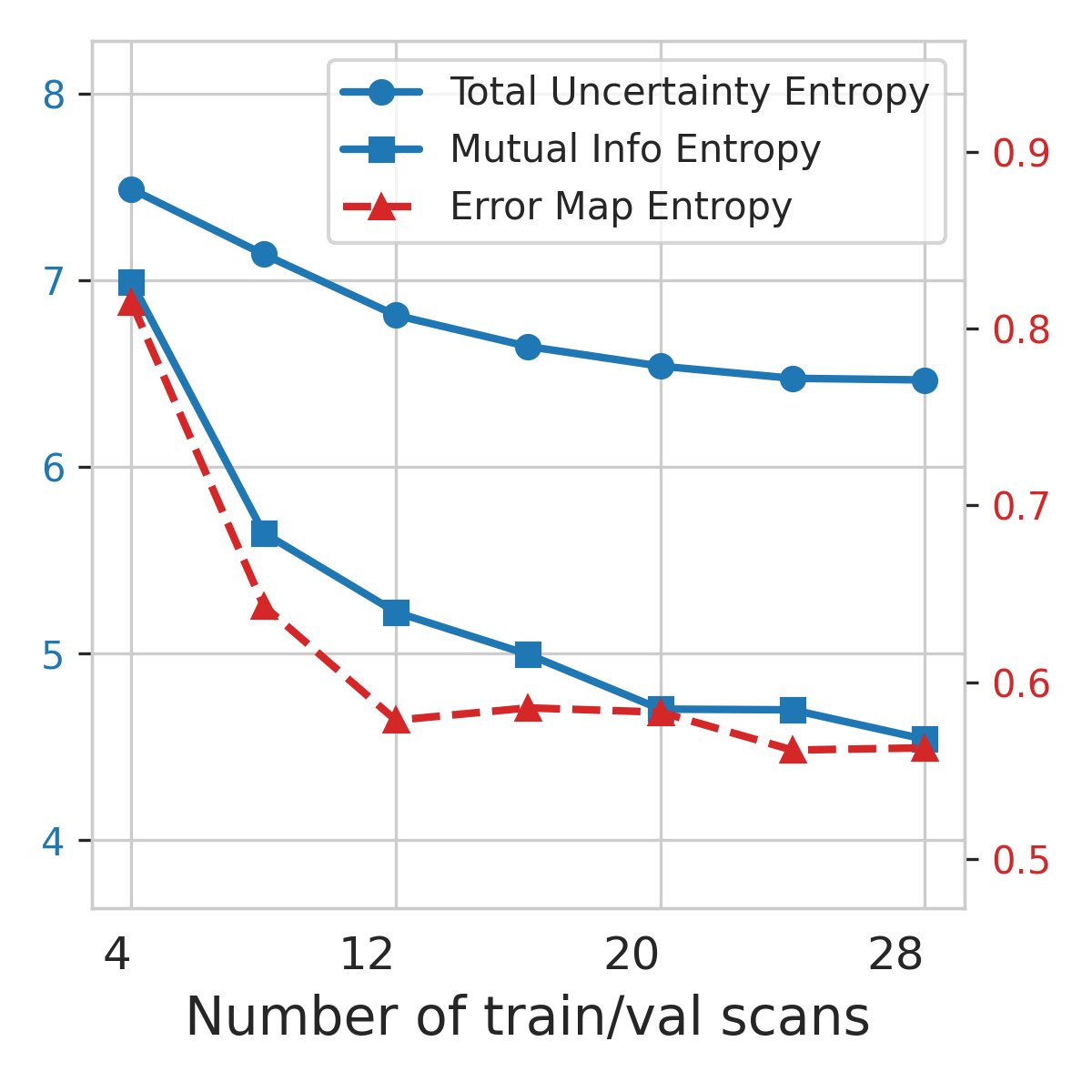}
        \caption{3-Ch}
    \end{subfigure}
    \hfill
    \begin{subfigure}[b]{0.24\textwidth}
        \includegraphics[width=\linewidth]{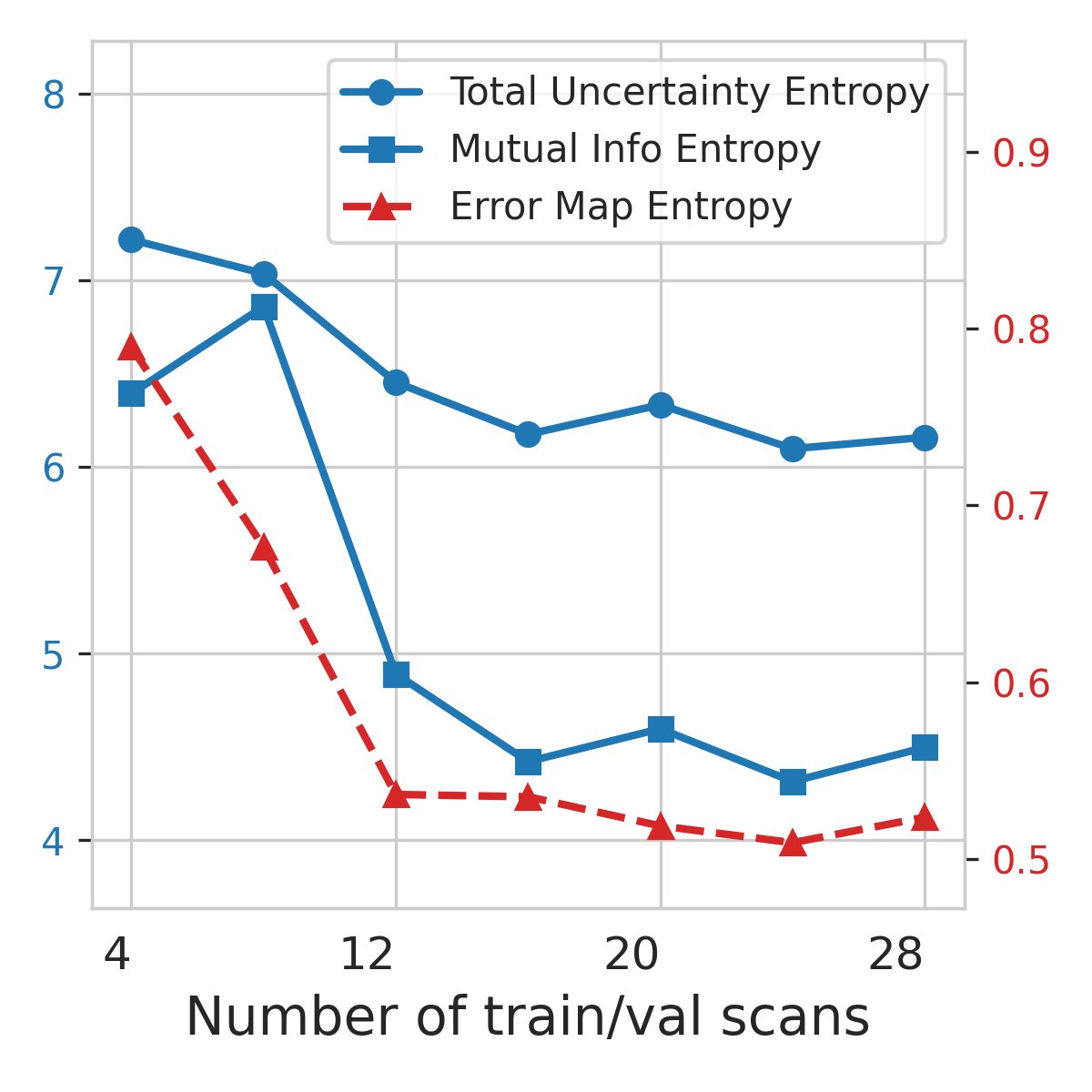}
        \caption{6-Ch}
    \end{subfigure}
    \hfill
    \begin{subfigure}[b]{0.24\textwidth}
        \includegraphics[width=\linewidth]{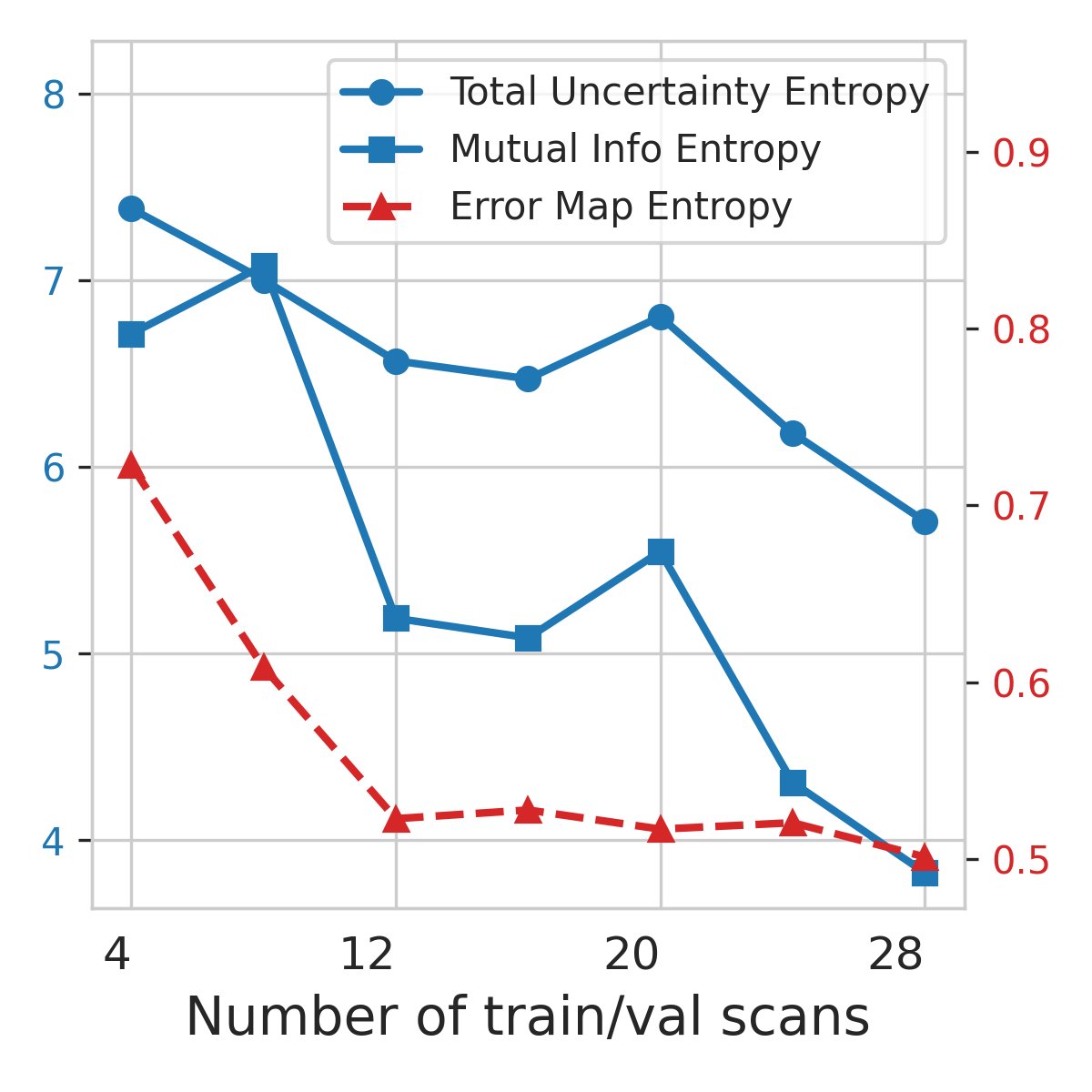}
        \caption{12-Ch}
    \end{subfigure}

    \vspace{2mm}  

    \begin{subfigure}[b]{0.24\textwidth}
        \includegraphics[width=\linewidth]{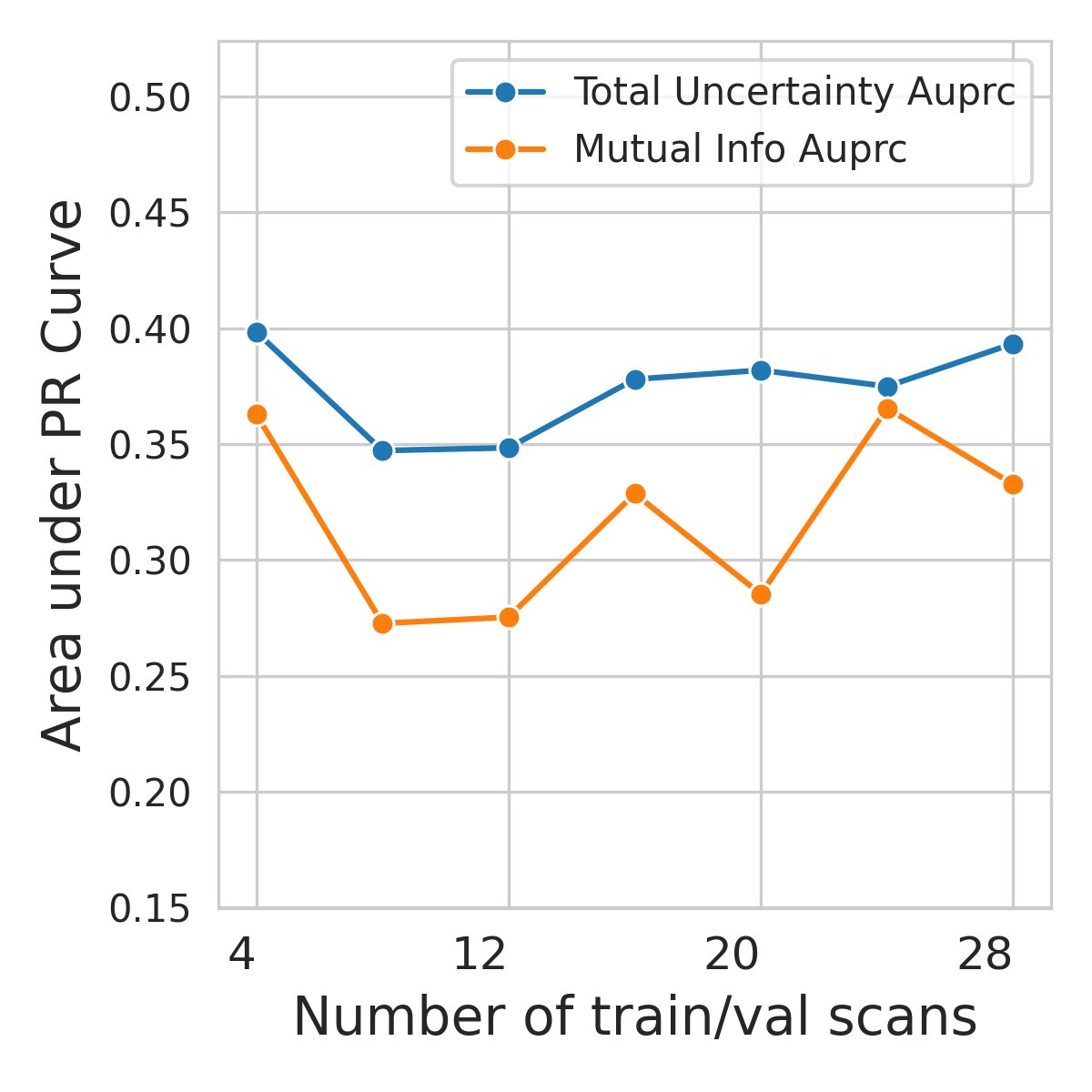}
        \caption{1-Ch}
    \end{subfigure}
    \hfill
    \begin{subfigure}[b]{0.24\textwidth}
        \includegraphics[width=\linewidth]{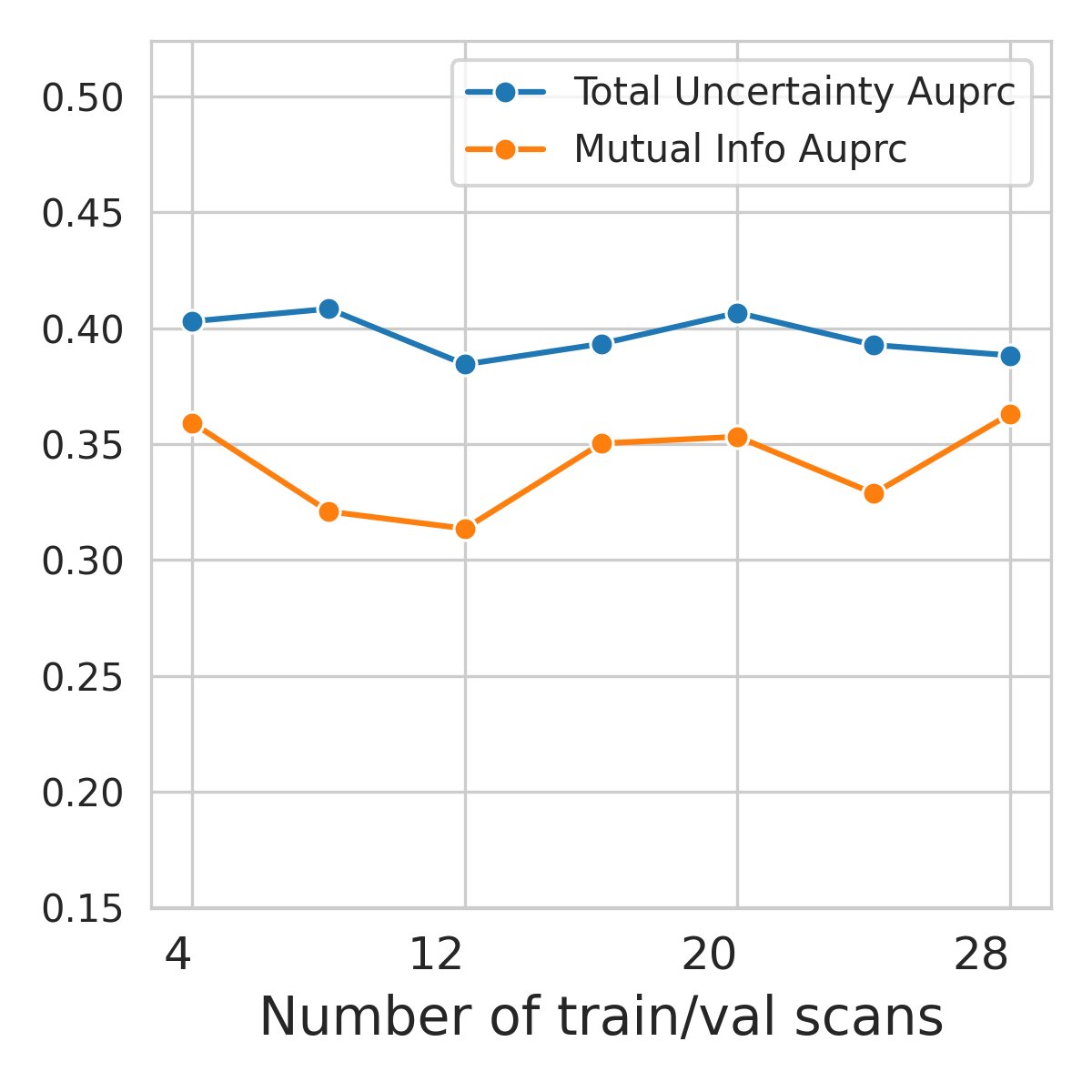}
        \caption{3-Ch}
    \end{subfigure}
    \hfill
    \begin{subfigure}[b]{0.24\textwidth}
        \includegraphics[width=\linewidth]{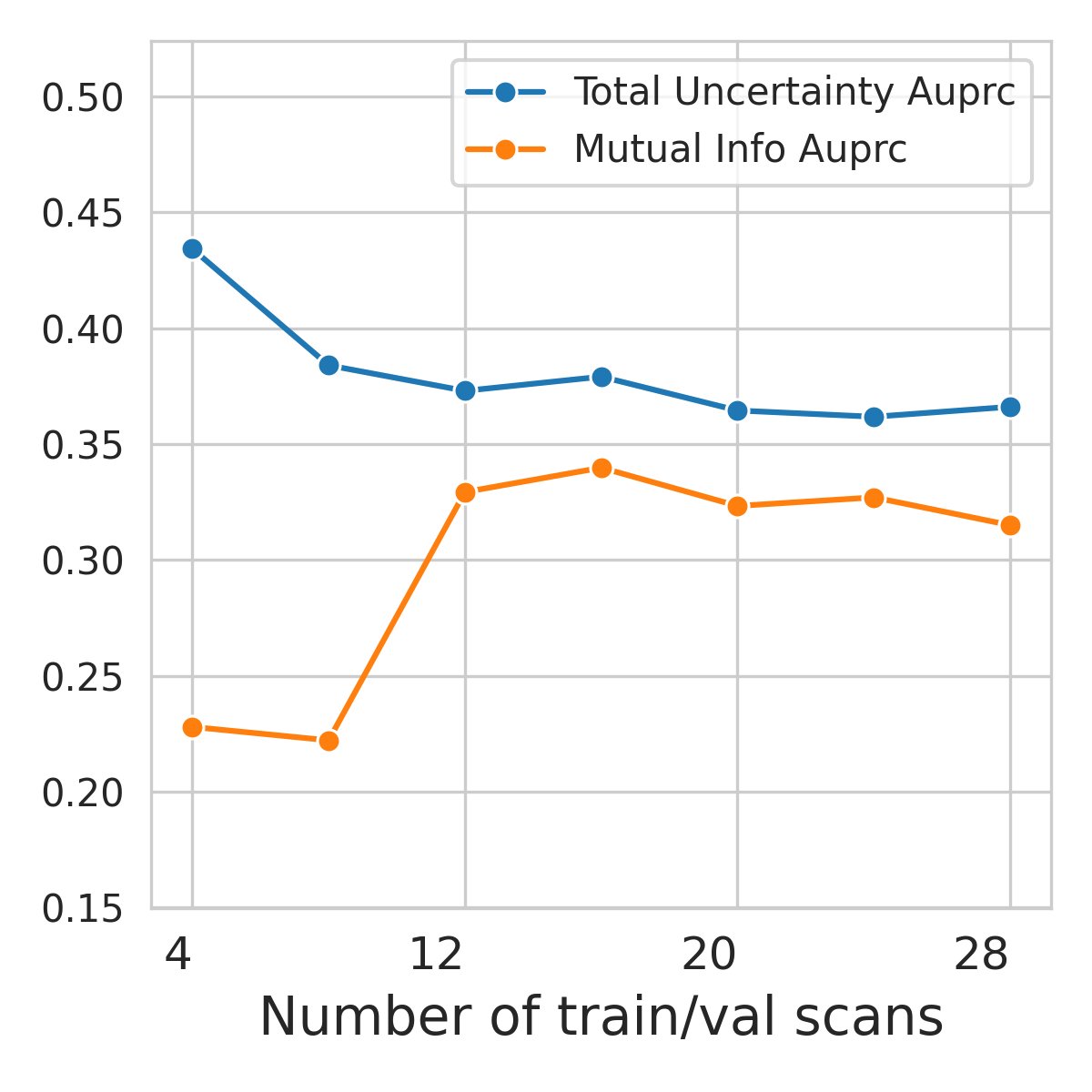}
        \caption{6-Ch}
    \end{subfigure}
    \hfill
    \begin{subfigure}[b]{0.24\textwidth}
        \includegraphics[width=\linewidth]{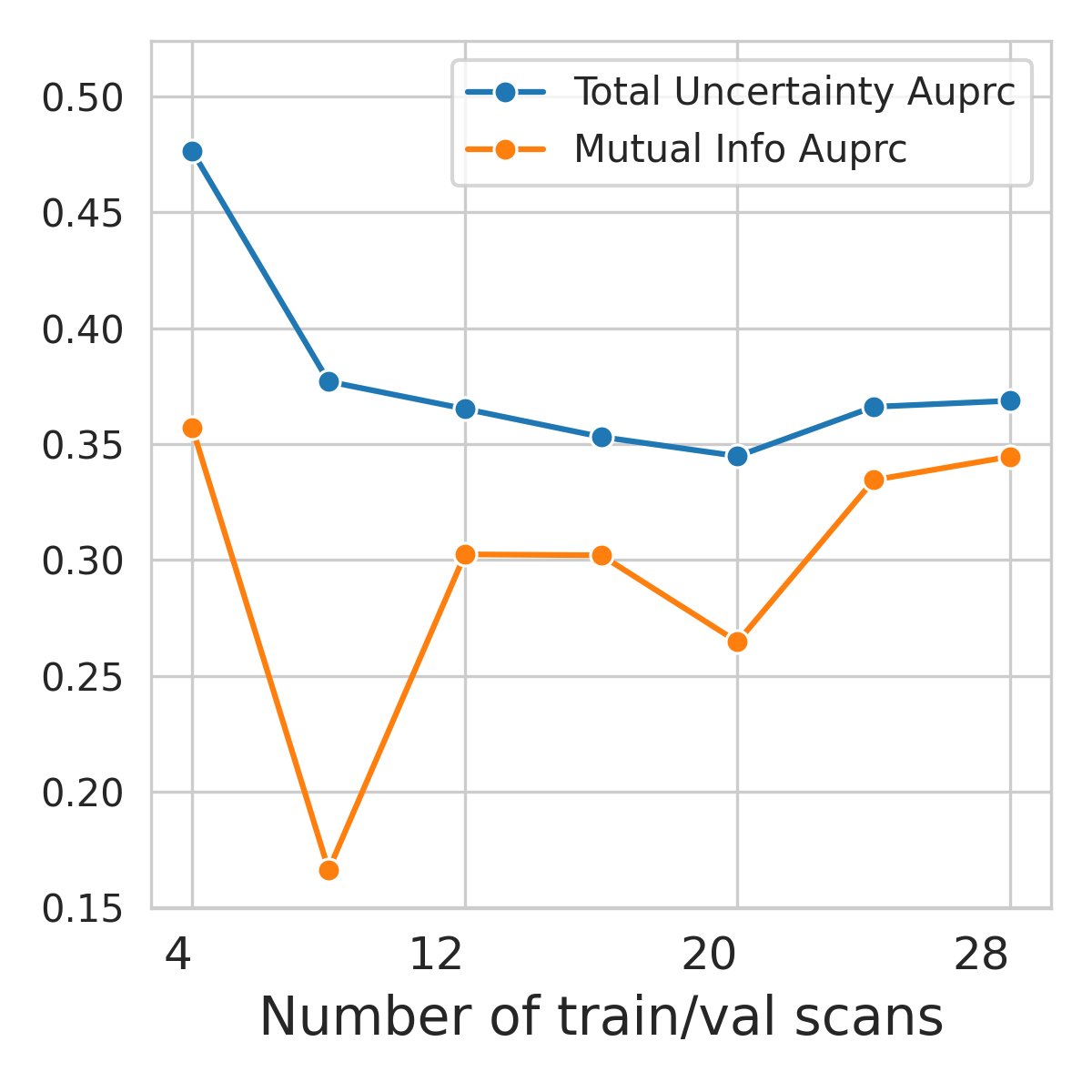}
        \caption{12-Ch}
    \end{subfigure}

    \caption{Uncertainty quantification trends for ensemble models trained with different feature configurations and varying numbers of train/val scans (4–28). Top row (a–d): Entropy of the total uncertainty map and mutual information map (blue), and prediction error map (red). Bottom row (e–h): Area under the precision-recall curve (AUPRC) measuring how well the mutual information map (orange) and total uncertainty map (blue) correlate with actual prediction errors.}
    \label{fig:data_efficienty_uncert}
\end{figure}

%% file: tex_of_fig_tab/fig_forestsemantic_2d.tex
\begin{figure}[!htb]
    \centering

    \begin{subfigure}{0.32\textwidth}
        \includegraphics[width=\linewidth]{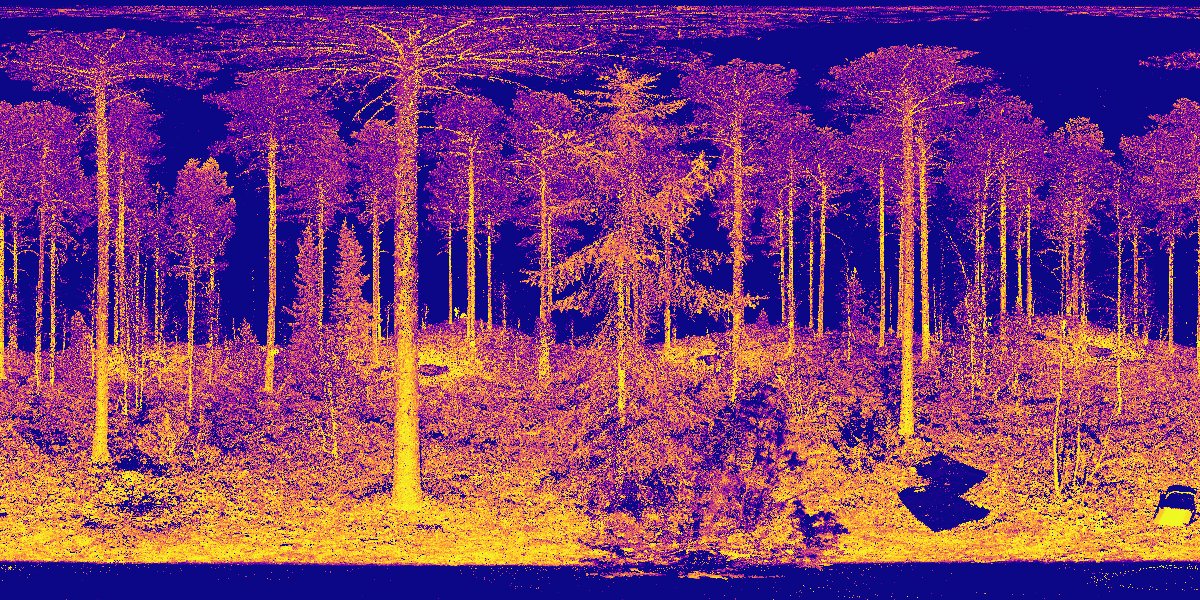}
        \caption*{(a) Intensity}
    \end{subfigure}
    \begin{subfigure}{0.32\textwidth}
        \includegraphics[width=\linewidth]{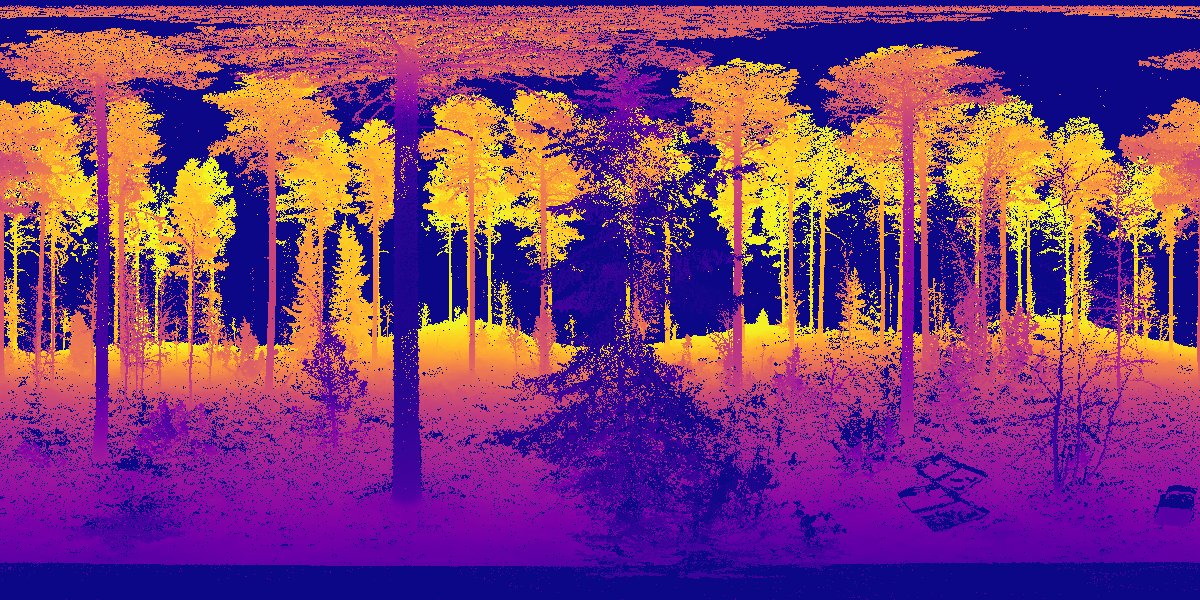}
        \caption*{(b) Range}
    \end{subfigure}
    \begin{subfigure}{0.32\textwidth}
        \includegraphics[width=\linewidth]{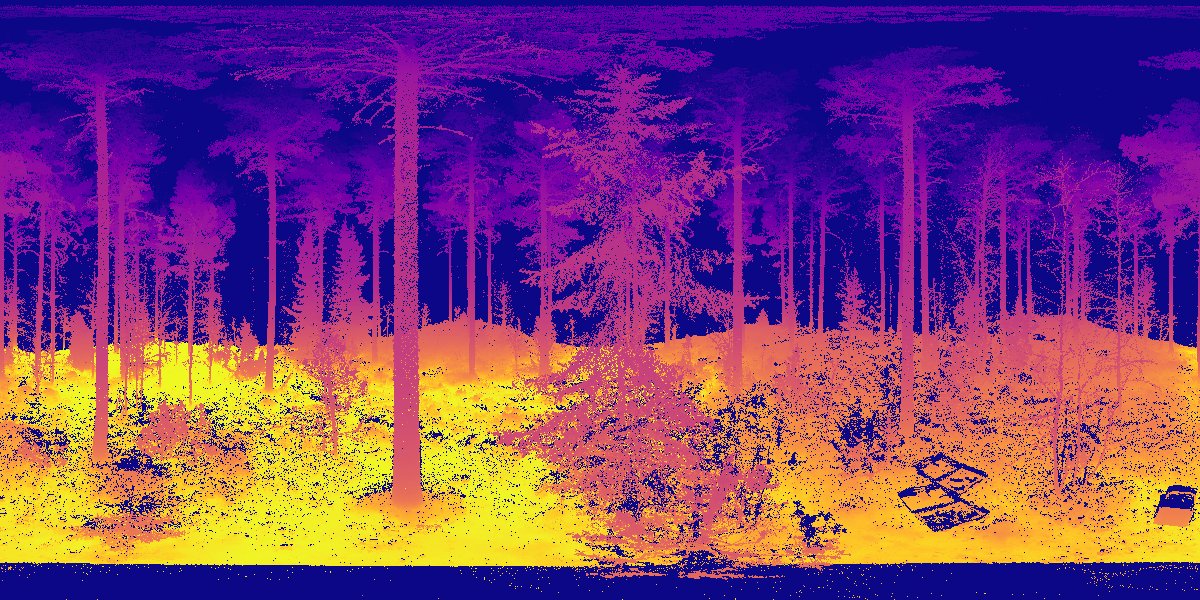}
        \caption*{(c) Z-Inv}
    \end{subfigure}

    \begin{subfigure}{0.32\textwidth}
        \includegraphics[width=\linewidth]{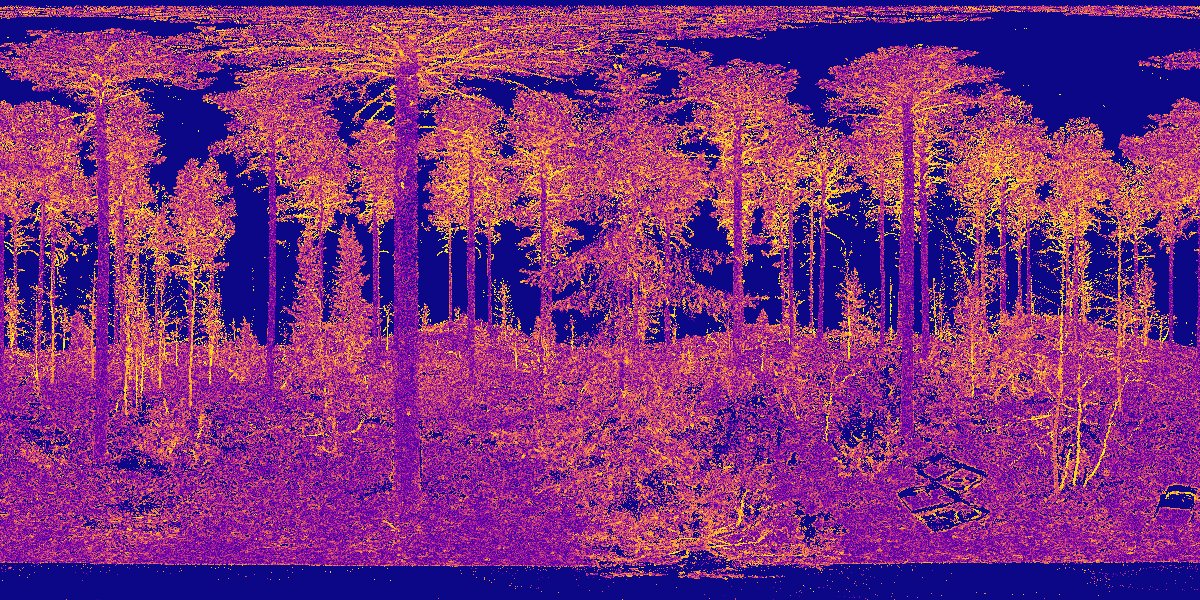}
        \caption*{(d) Anisotropy}
    \end{subfigure}
    \begin{subfigure}{0.32\textwidth}
        \includegraphics[width=\linewidth]{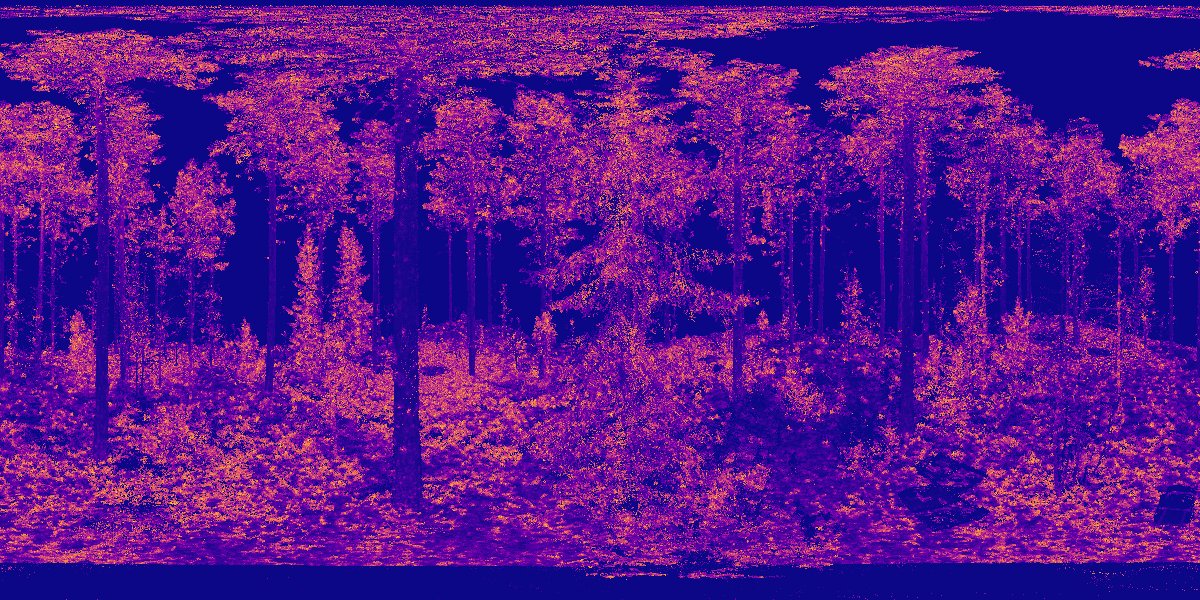}
        \caption*{(e) Curvature}
    \end{subfigure}
    \begin{subfigure}{0.32\textwidth}
        \includegraphics[width=\linewidth]{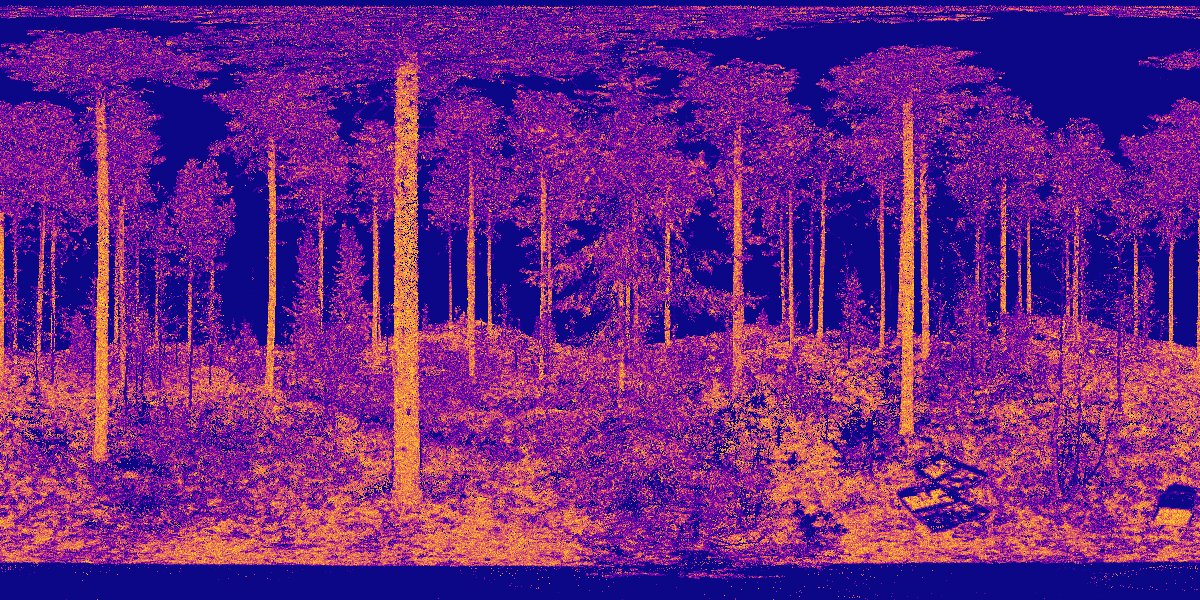}
        \caption*{(f) Planarity}
    \end{subfigure}

    \begin{subfigure}{0.32\textwidth}
        \includegraphics[width=\linewidth]{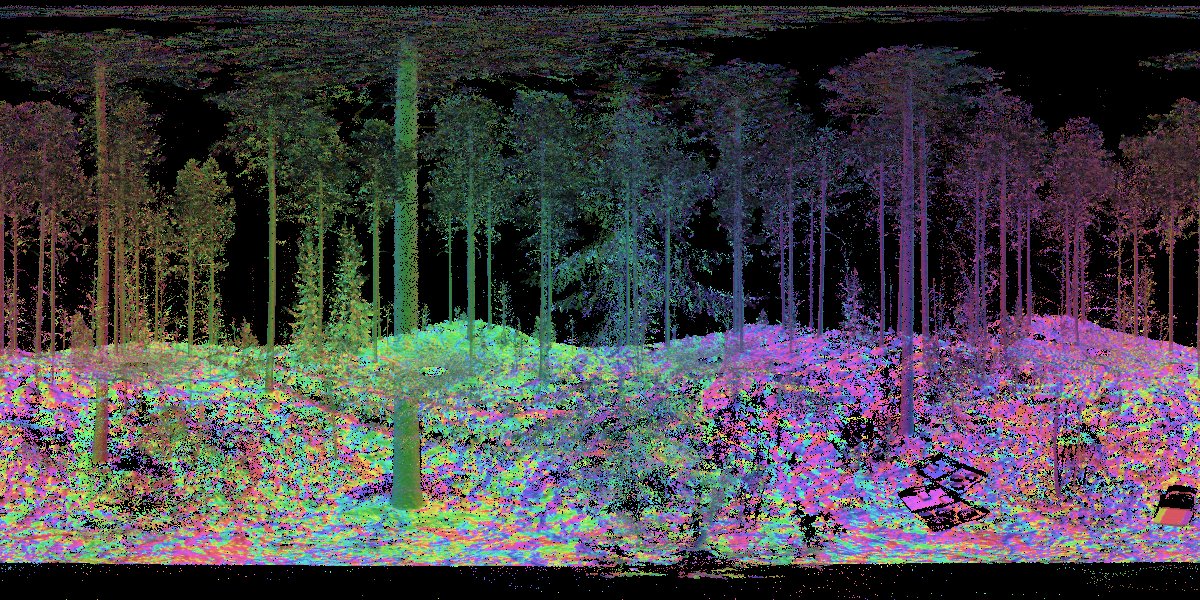}
        \caption*{(g) Normals pseudo color stack}
    \end{subfigure}
    \begin{subfigure}{0.32\textwidth}
        \includegraphics[width=\linewidth]{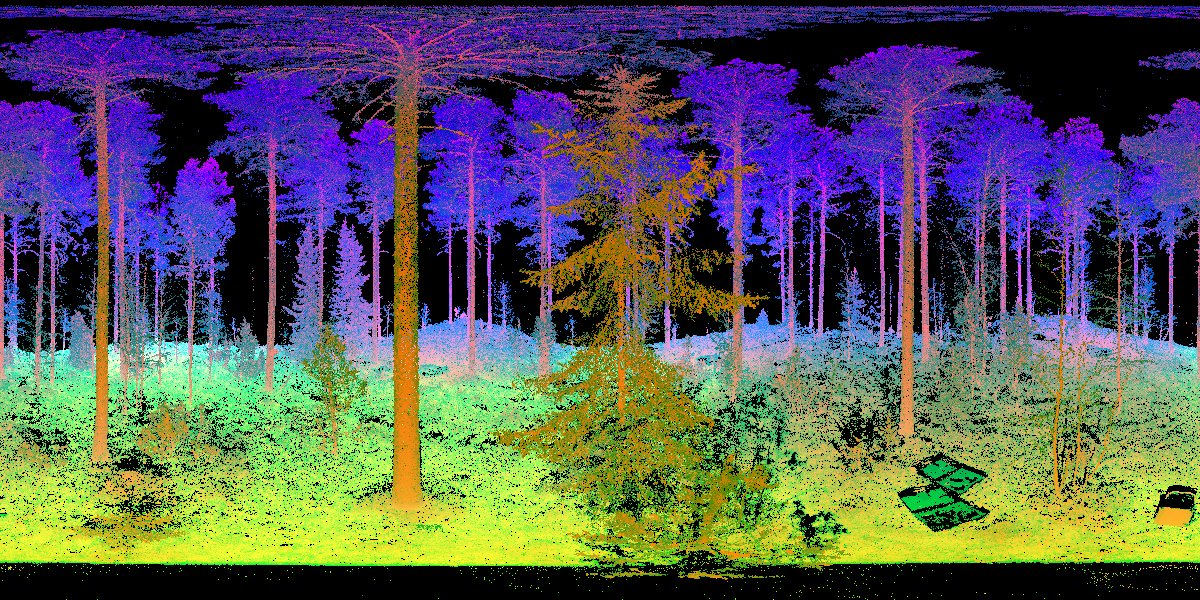}
        \caption*{(h) I.R.Z stack}
    \end{subfigure}
    \begin{subfigure}{0.32\textwidth}
        \includegraphics[width=\linewidth]{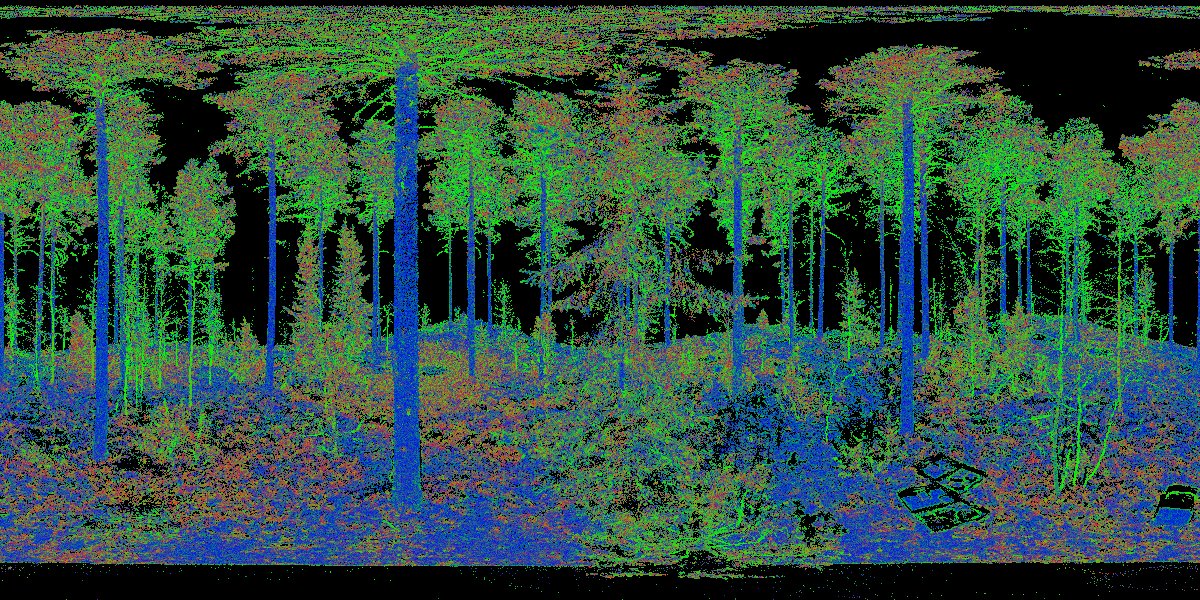}
        \caption*{(i) C.A.P stack}
    \end{subfigure}

    \begin{subfigure}{0.32\textwidth}
        \includegraphics[width=\linewidth]{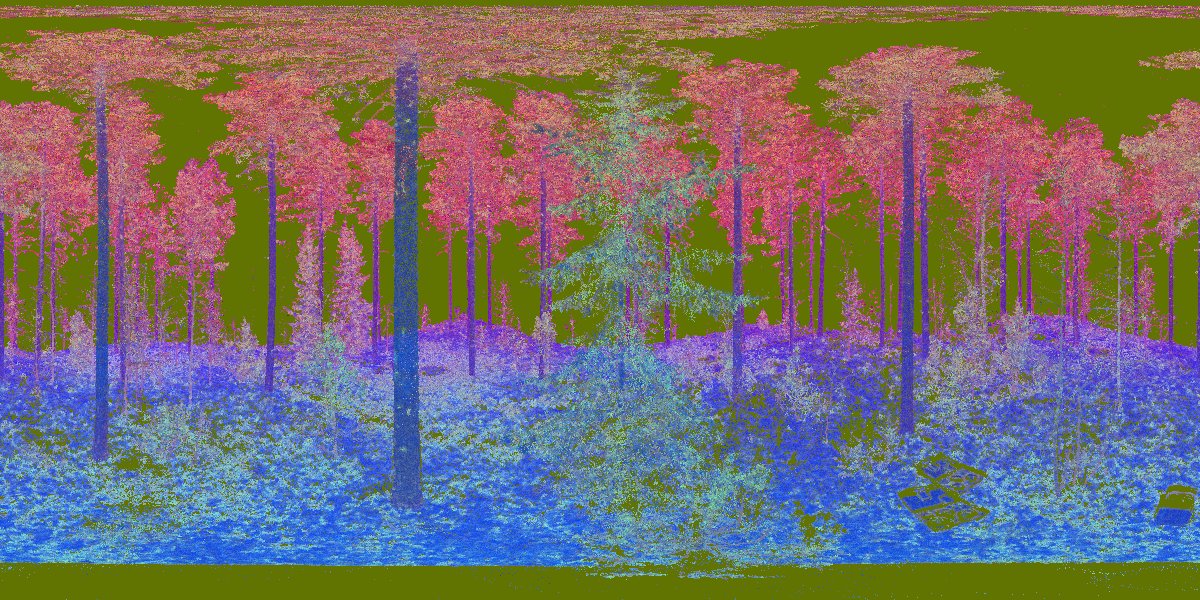}
        \caption*{(j) PCA stack}
    \end{subfigure}
    \begin{subfigure}{0.32\textwidth}
        \includegraphics[width=\linewidth]{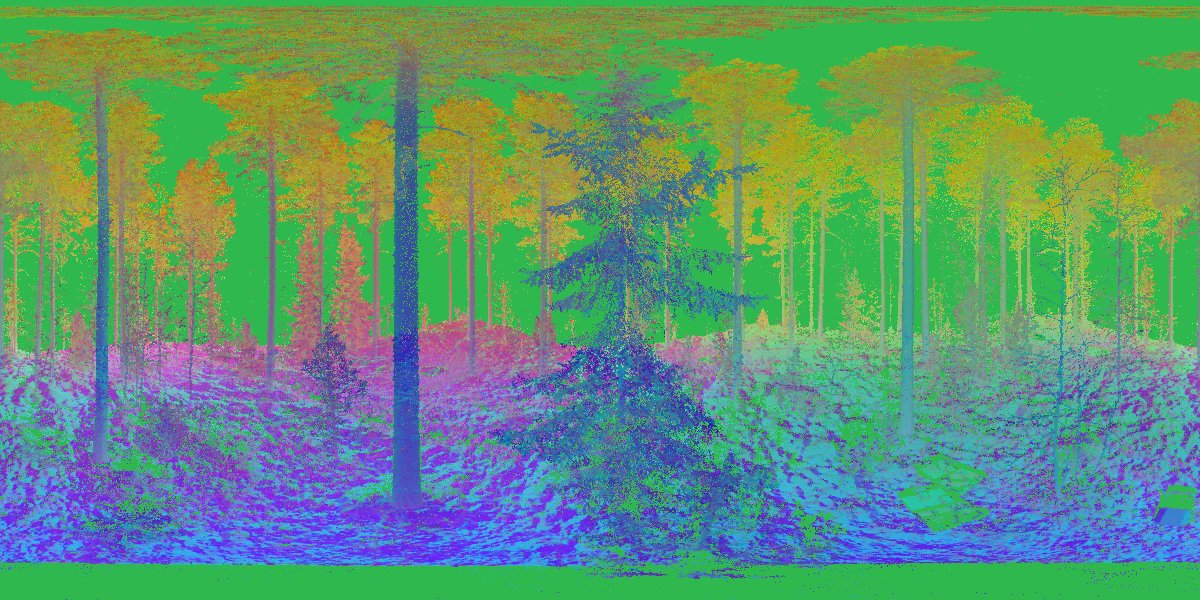}
        \caption*{(k) MNF stack}
    \end{subfigure}
    \begin{subfigure}{0.32\textwidth}
        \includegraphics[width=\linewidth]{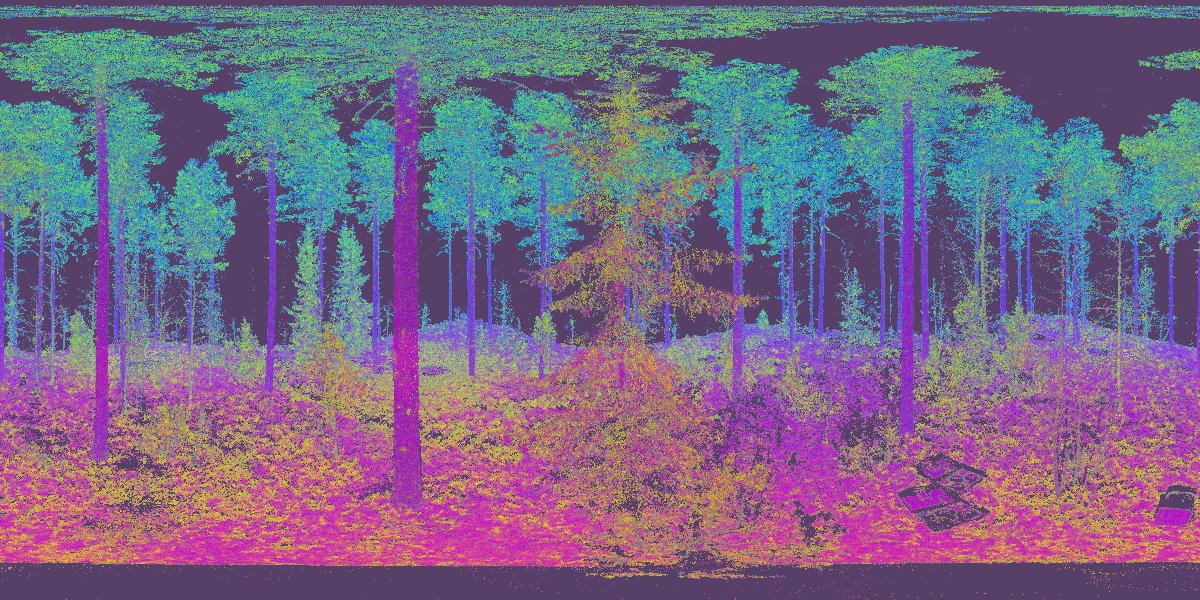}
        \caption*{(l) ICA stack}
    \end{subfigure}

    \vspace{0.5em}
    \begin{subfigure}[b]{0.5\textwidth}
        \centering
        \includegraphics[width=\linewidth]{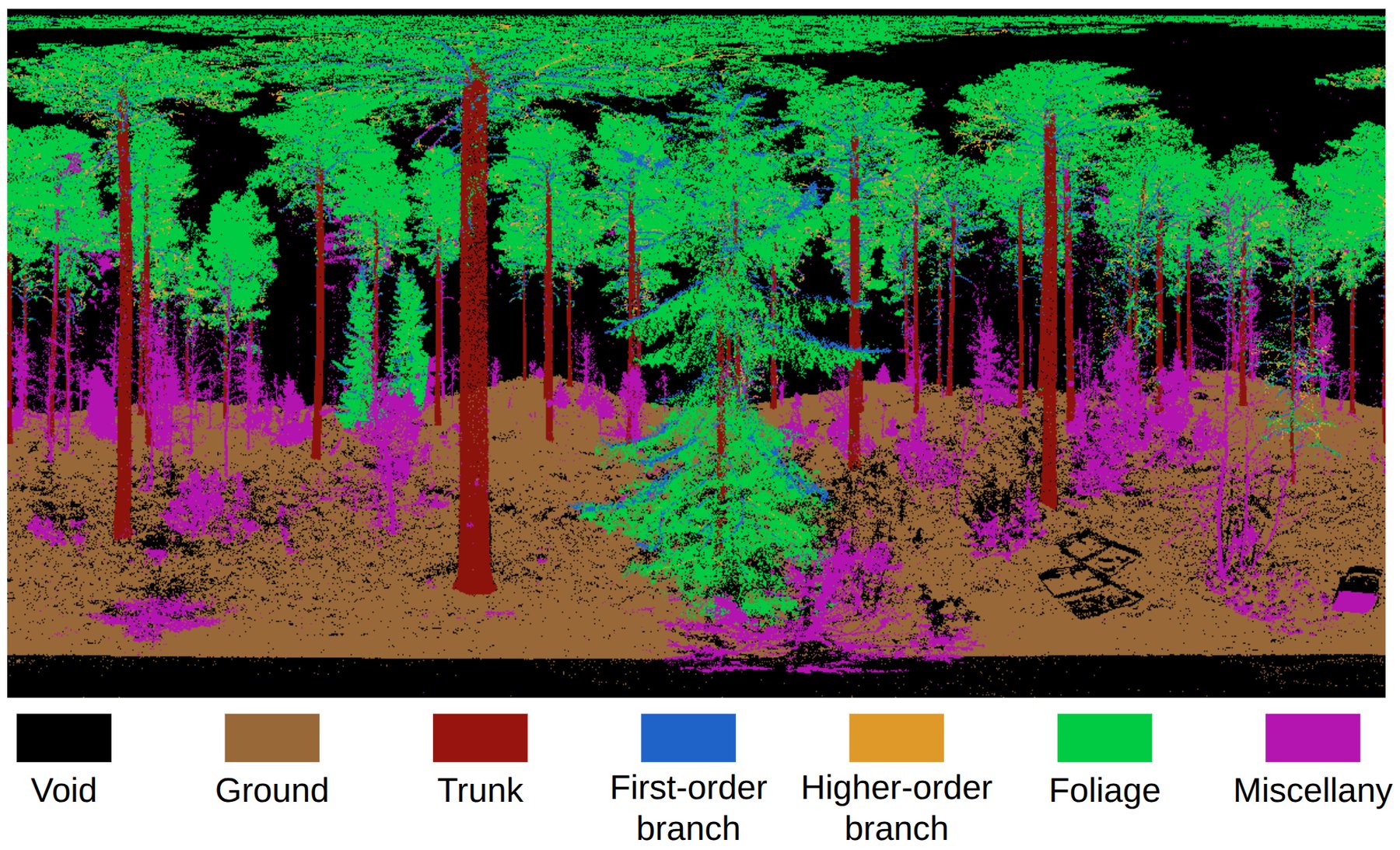}
        \caption*{(m) GT segmentation mask}
    \end{subfigure}
    \begin{subfigure}[b]{0.4\textwidth}
        \centering
        \includegraphics[width=\linewidth]{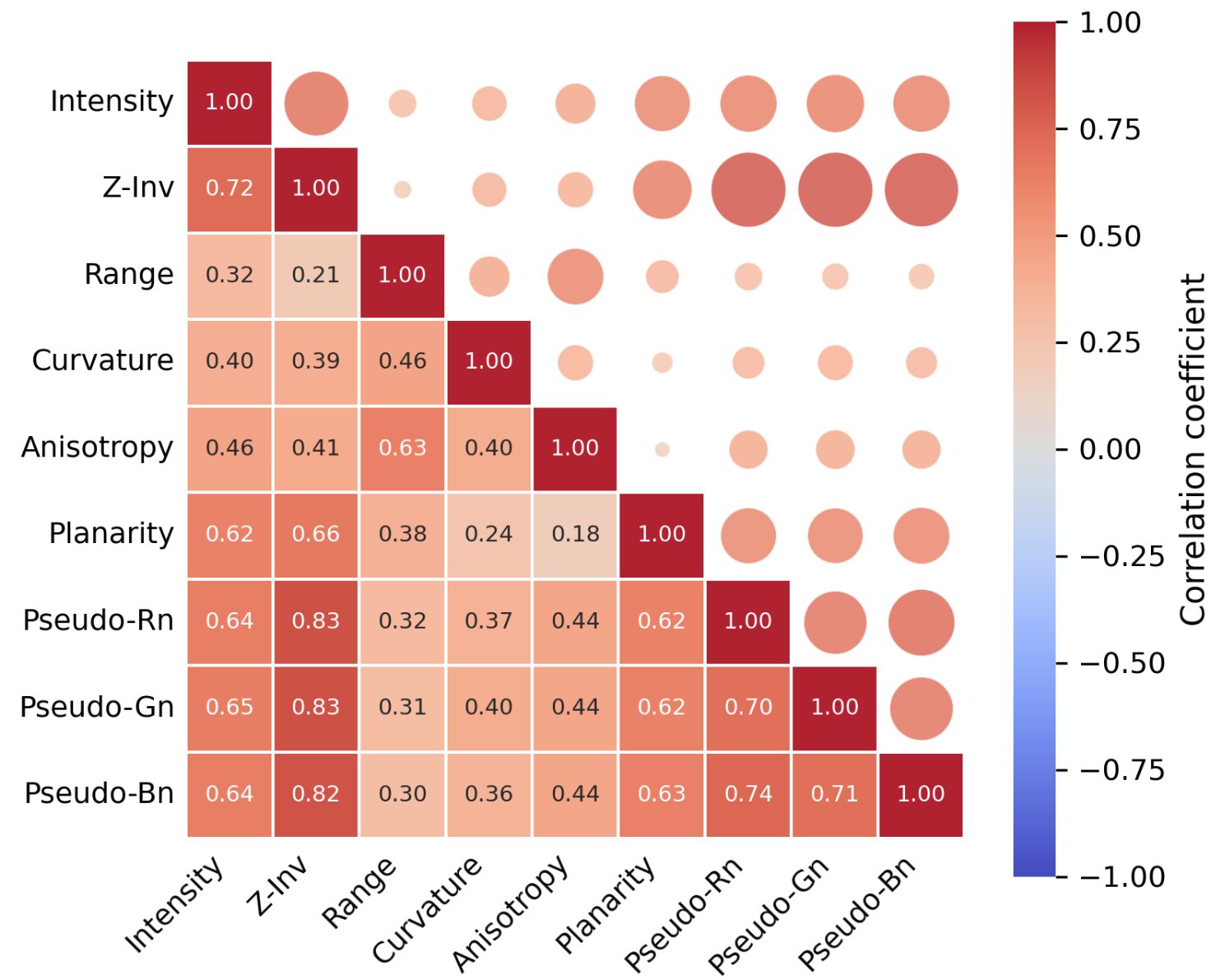}
        \caption*{(n) Correlation of 9 channels (a)-(g)}
    \end{subfigure}

    \caption{Spherical-projection views of a \emph{ForestSemantic} pseudo-scan, organized by feature type.
    (a–c) Basic features: preprocessed intensity, range, and Z-Inv;  
    (d–f) Geometric features: anisotropy, curvature, and planarity;  
    (g–i) Combined geometric and appearance features: normals, I.R.Z., and C.A.P;  
    (j–l) Statistical features: first three components of PCA, MNF, and ICA;  
    (m) Ground truth semantic segmentation mask projected into 2D from the 3D point cloud labels;
    (n) The corresponding correlation matrix of the 9 channels mentioned in (a)-(g).}
    \label{fig:forestsemantic_2d_visuals}
\end{figure}

%% file: tex_of_fig_tab/fig_forestsemantic_pcd.tex
\begin{figure}[!htbp]
    \centering

    \begin{subfigure}{0.32\textwidth}
        \includegraphics[width=\linewidth]{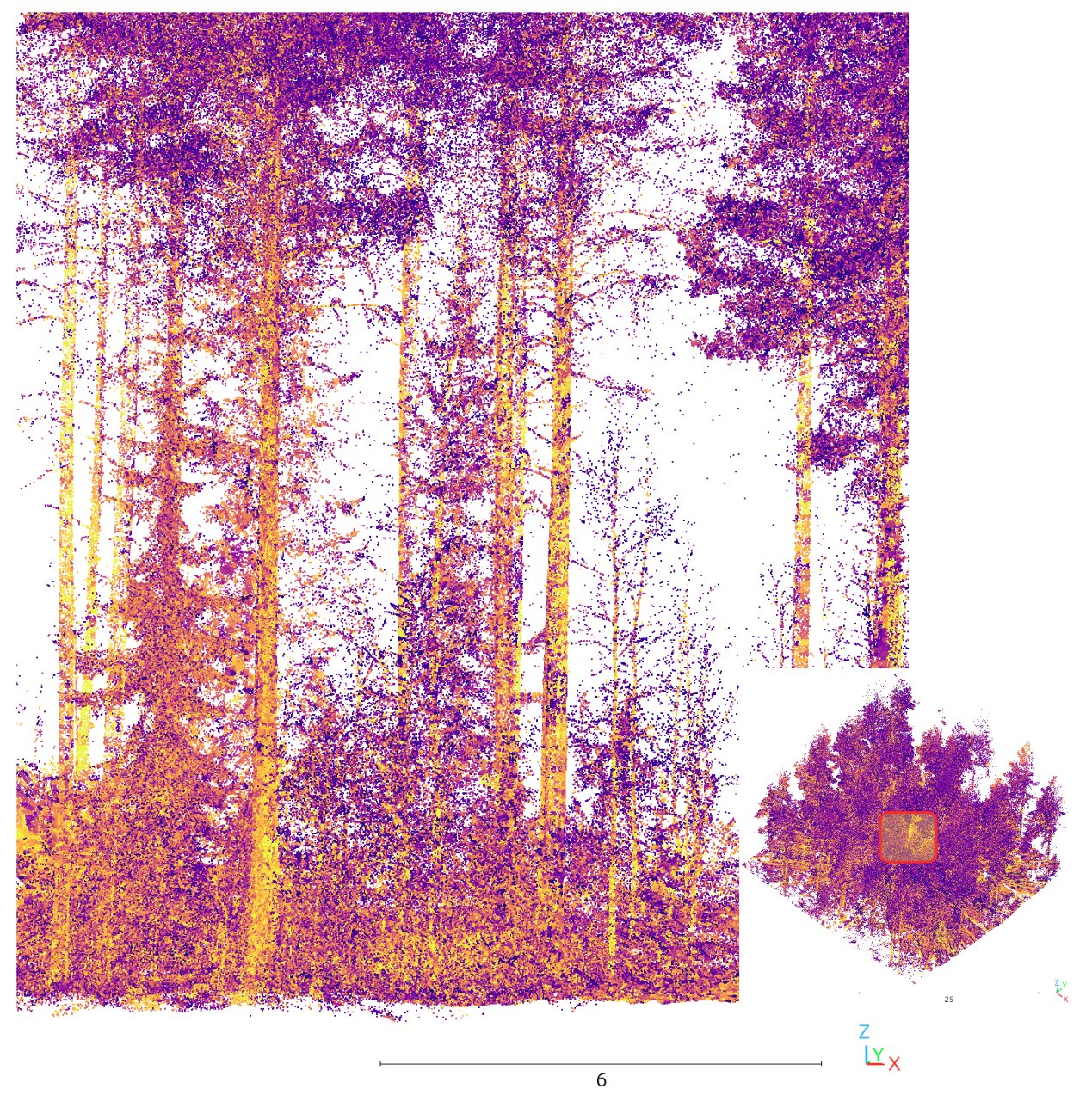}
        \caption*{(a) Intensity}
    \end{subfigure}
    \begin{subfigure}{0.32\textwidth}
        \includegraphics[width=\linewidth]{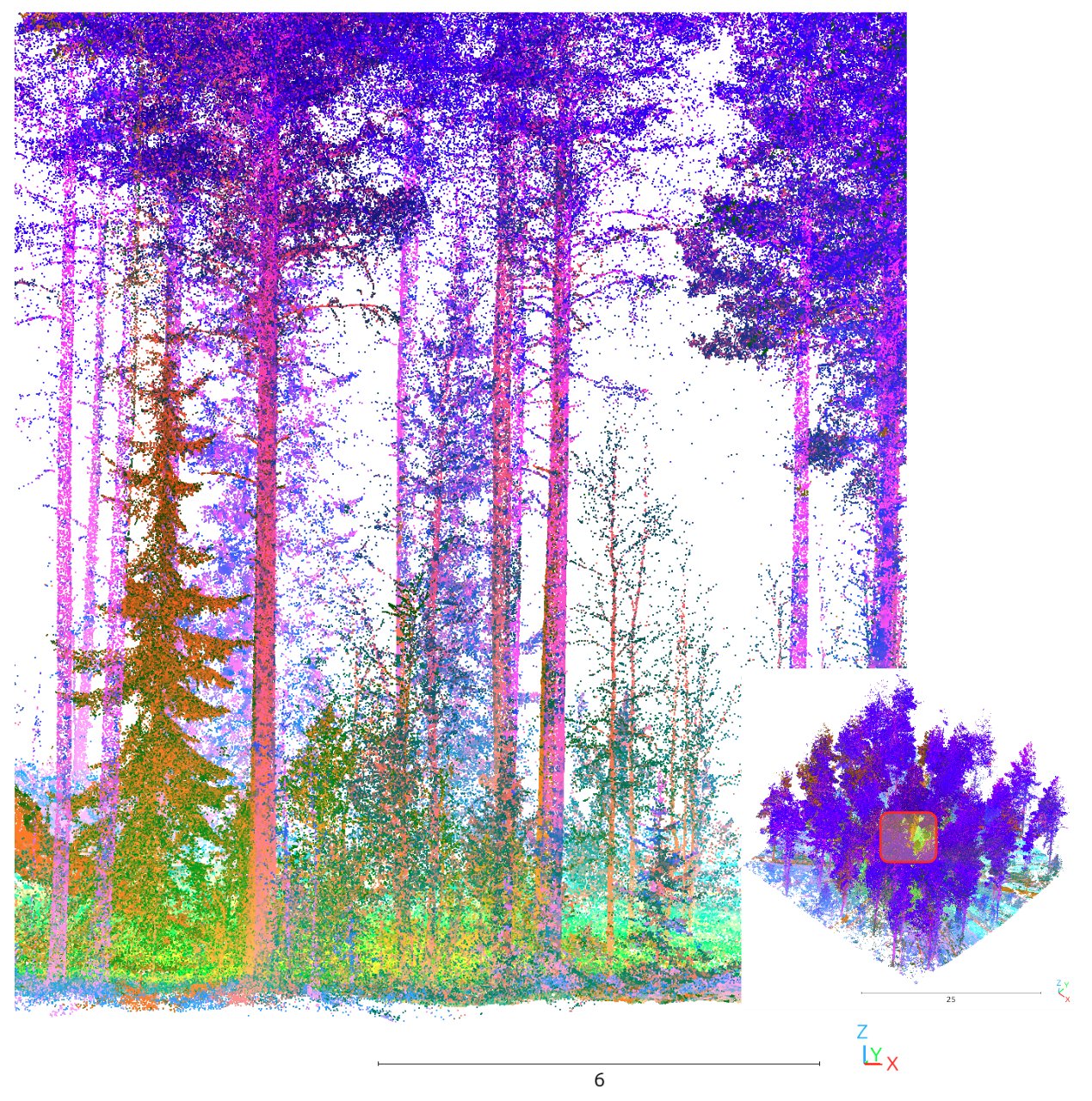}
        \caption*{(b) I.R.Z stack}
    \end{subfigure}
    \begin{subfigure}{0.32\textwidth}
        \includegraphics[width=\linewidth]{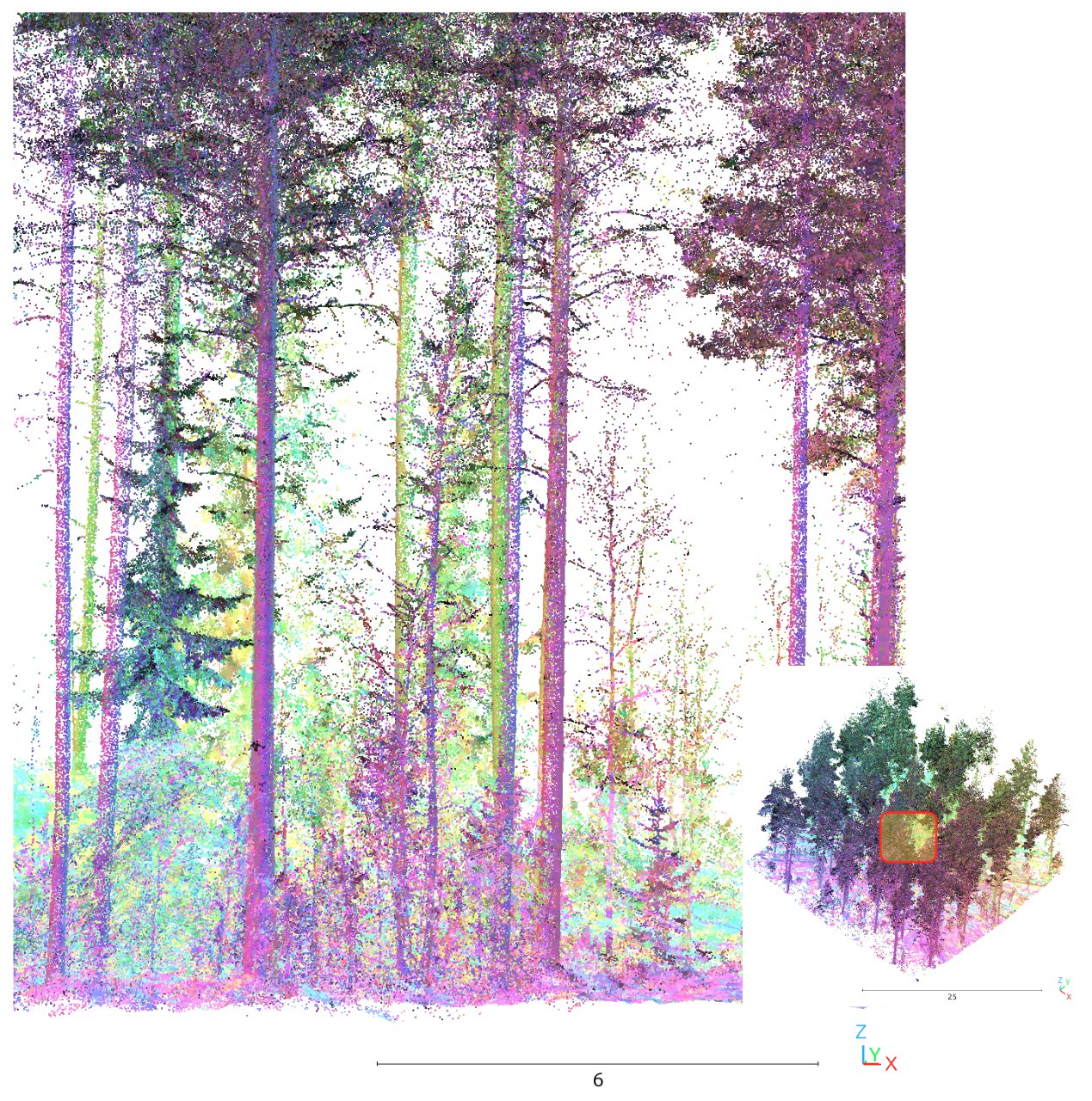}
        \caption*{(c) Normals stack}
    \end{subfigure}

    \begin{subfigure}{0.32\textwidth}
        \includegraphics[width=\linewidth]{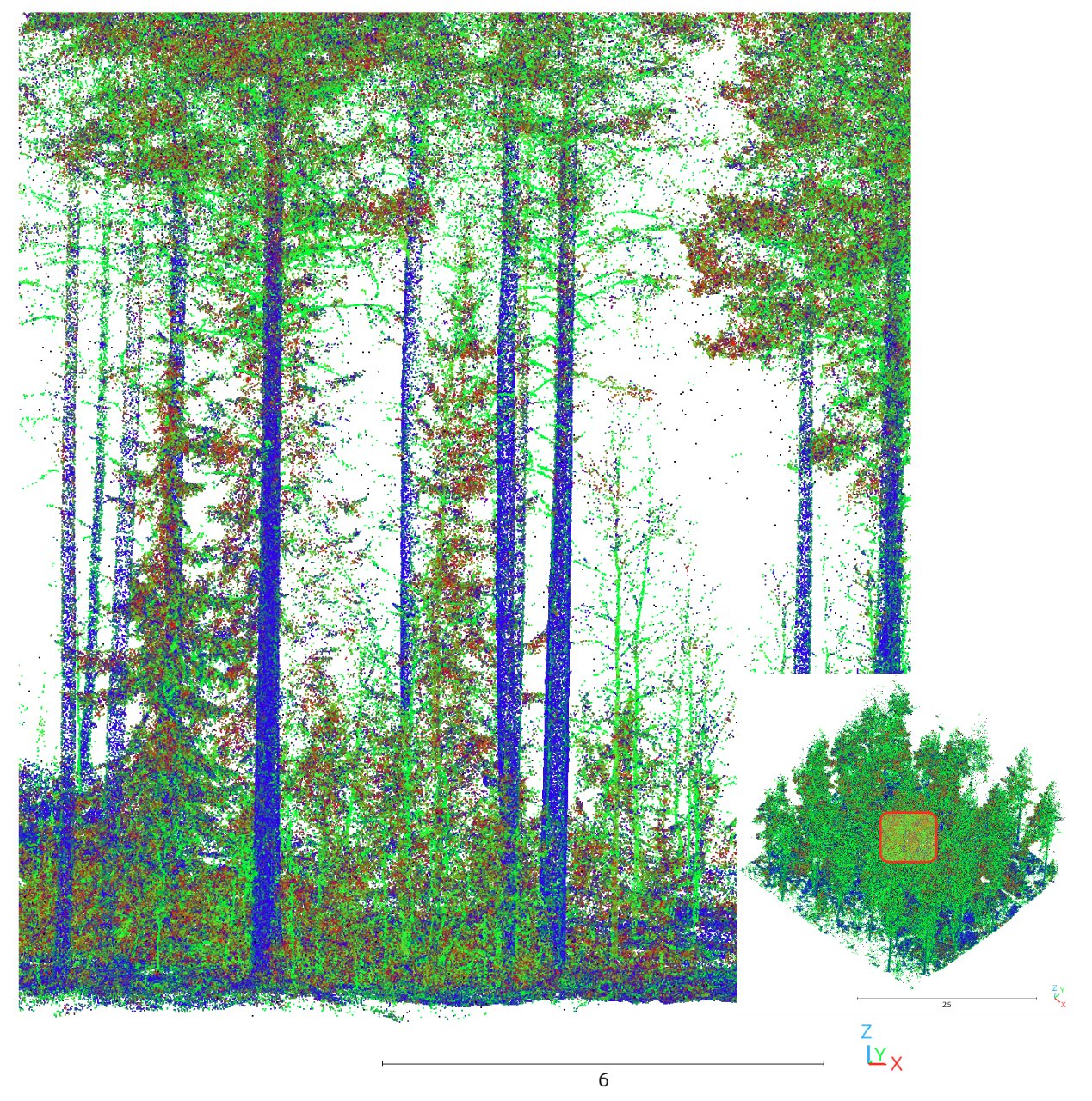}
        \caption*{(d) C.A.P stack}
    \end{subfigure}
    \begin{subfigure}{0.32\textwidth}
        \includegraphics[width=\linewidth]{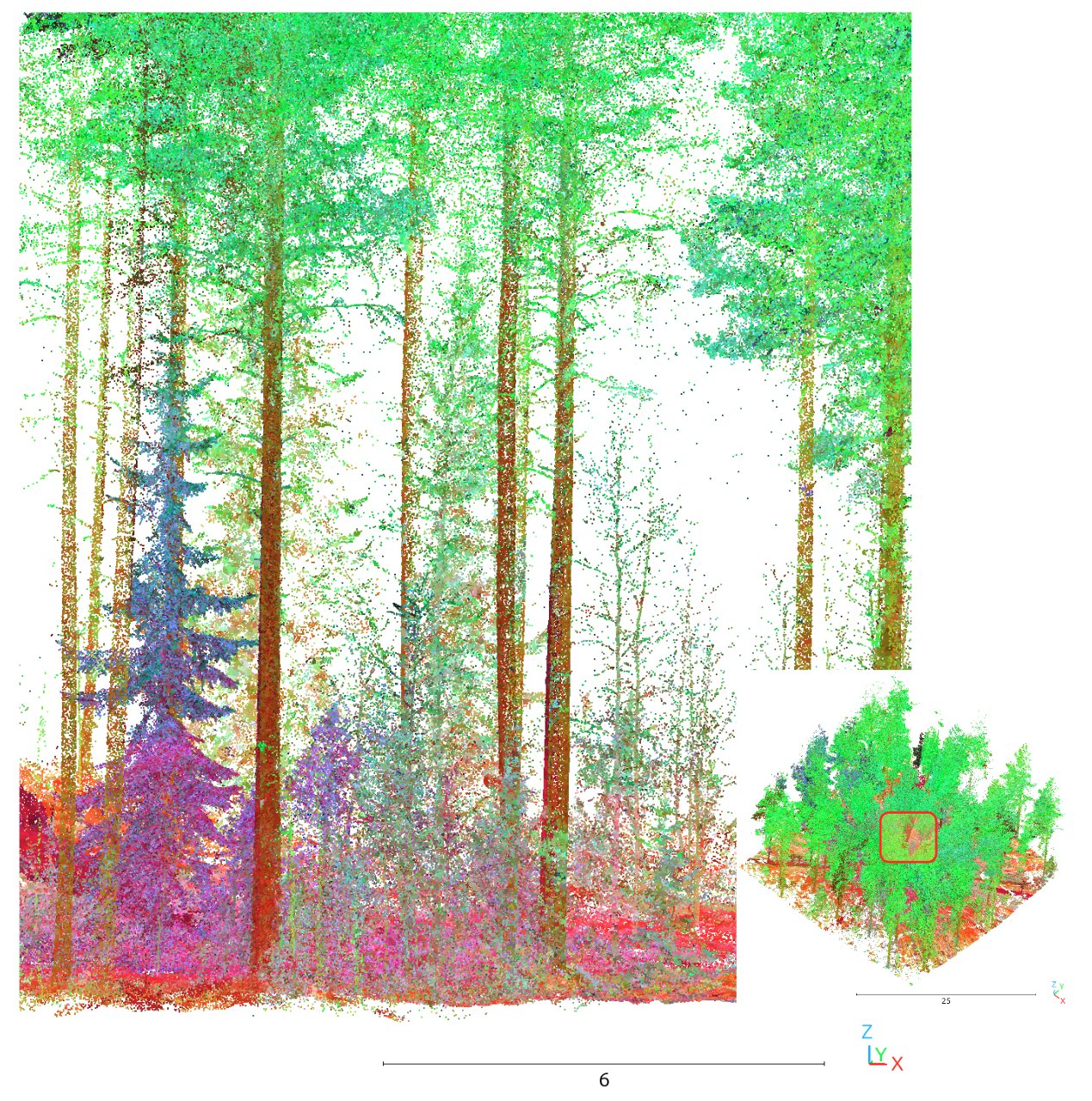}
        \caption*{(e) PCA stack}
    \end{subfigure}
    \begin{subfigure}{0.33\textwidth}
        \includegraphics[width=\linewidth]{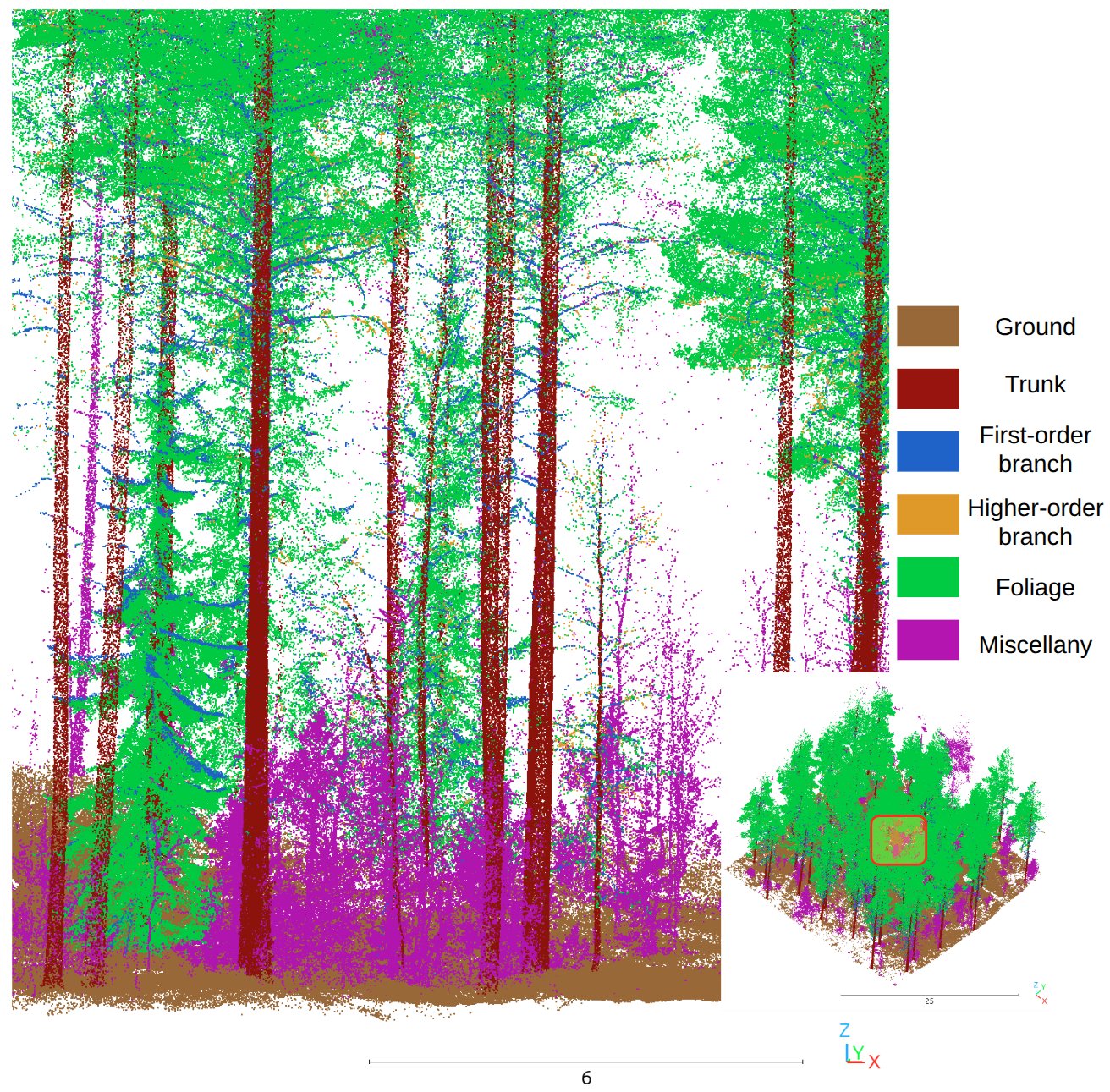}
        \caption*{(f) GT labels}
    \end{subfigure}

    \caption{
    Feature-driven colorization of the \emph{ForestSemantic} 3D point cloud. 
    Each inset shows a full overview of the entire plot to provide spatial context for the zoomed-in region. Animation available in~\ref{app:animations}-Table~\ref{tab:animations}}
    
    \label{fig:forestsemantic_3d_pcd}
\end{figure}

%% file: tex_of_fig_tab/fig_forestsemantic_balls.tex
\begin{figure}[!htbp]
    \centering
    \begin{minipage}[b]{0.32\textwidth}
        \begin{subfigure}[b]{\textwidth}
            \centering
            \includegraphics[width=\textwidth]{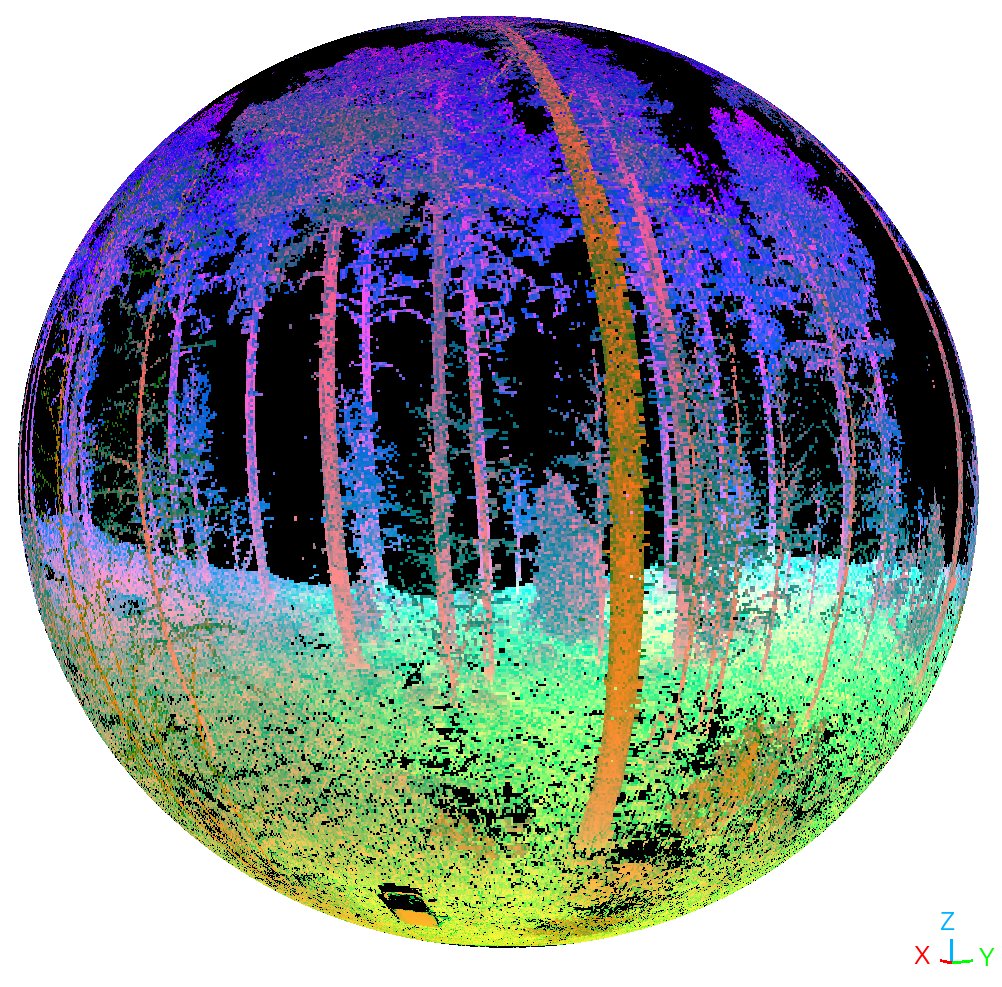}
            \caption*{(a) I.R.Z stack}
        \end{subfigure}
    \end{minipage}
    \begin{minipage}[b]{0.32\textwidth}
        \begin{subfigure}[b]{\textwidth}
            \centering
            \includegraphics[width=\textwidth]{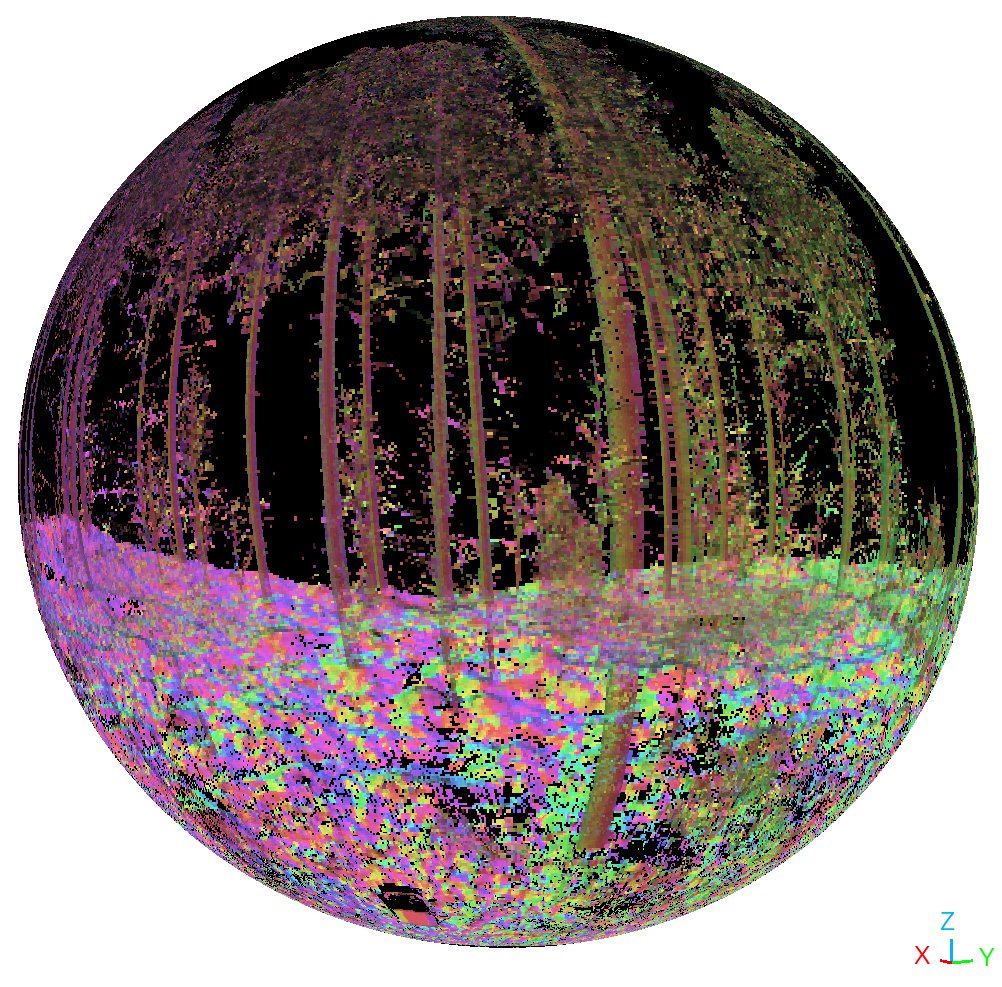}
            \caption*{(b) Normals pseudo color stack}
        \end{subfigure}
    \end{minipage}
    \begin{minipage}[b]{0.32\textwidth}
        \begin{subfigure}[b]{\textwidth}
            \centering
            \includegraphics[width=\textwidth]{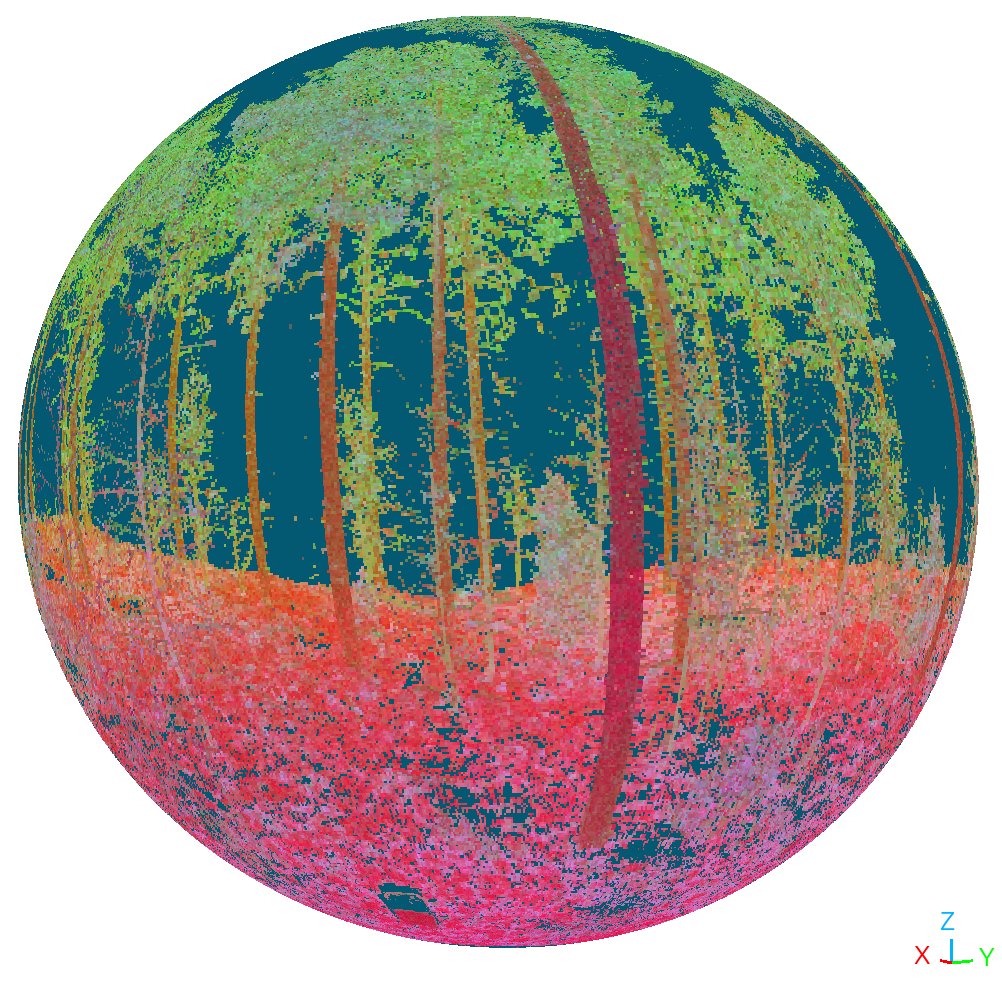}
            \caption*{(c) PCA stack}
        \end{subfigure}
    \end{minipage}
    
    \caption{Colorized virtual spheres of an example scan in the \textit{ForestSemantic} dataset with three different feature groups. Animation available in~\ref{app:animations}-Table~\ref{tab:animations}}
    \label{fig:forestsemantic_balls}
\end{figure}

%% file: tex_of_fig_tab/tab_forestsemantic_feature_compare.tex
\begin{table*}[htbp]
    \centering
    \footnotesize
    \caption{Comparison of feature–input combinations for semantic segmentation on \emph{ForestSemantic}.  
             Within each metric column, the top three scores are highlighted with progressively darker blue shading  
             (dark = best, medium = second, light = third); the baseline \textit{irz\_raw} row is shown in light gray for reference.}
    \vspace{0.4em}
    \resizebox{\textwidth}{!}{
    \begin{tabular}{ccccc|ccccccc|c}
        \toprule
        \multirow{2}{*}{\textbf{Feat.\#}} &
        \multirow{2}{*}{\textbf{Feature}} &
        \multicolumn{3}{c|}{\textbf{Global Metrics}} &
        \multicolumn{7}{c|}{\textbf{IoU per class}} &
        \multirow{2}{*}{\shortstack[c]{\textbf{mIoU}\\\textbf{Void excl.}}} \\
        \cmidrule(lr){3-5}\cmidrule(lr){6-12}
        & &
        \textbf{oAcc.} & \textbf{mAcc.} & \textbf{mIoU} &
        \textbf{Void} & \textbf{Ground} & \textbf{Trunk} &
        \shortstack{\textbf{1$^\mathrm{st}$-order}\\\textbf{branch}} &
        \shortstack{\textbf{Higher-order}\\\textbf{branch}} &
        \textbf{Foliage} & \textbf{Misc.} & \\
        \midrule
\rowcolor{black!6}
3 & irz\_raw &
0.870 & 0.537 & 0.453 &
0.952 & 0.748 & 0.483 &
0.090 & 0.010 & 0.633 & 0.254 & 0.393 \\

3 & I.R.Z &
 0.876 & 0.540 & 0.455 & 0.958 & 0.771 & 0.490 & %
 0.082 & 0.002 & 0.642 & 0.238 & 0.397 \\

3 & C.A.P &
 0.881 & 0.570 & 0.479 & 0.971 & 0.758 & 0.548 & %
 \third{0.131} & \second{0.013} & 0.655 & \third{0.280} & 0.421 \\

3 & N3 &
 0.878 & 0.547 & 0.461 & 0.957 & 0.767 & 0.561 & %
 0.039 & 0.009 & 0.640 & 0.251 & 0.403 \\

3 & PCA &
 0.885 & 0.538 & 0.460 & 0.971 & 0.766 & 0.541 & %
 0.053 & 0.002 & 0.670 & 0.213 & 0.406 \\

3 & MNF &
 0.789 & 0.466 & 0.363 & 0.826 & 0.610 & 0.420 & %
 0.011 & 0.000 & 0.506 & 0.170 & 0.309 \\

3 & ICA &
 0.845 & 0.509 & 0.411 & 0.899 & 0.701 & 0.489 & %
 0.012 & 0.000 & 0.598 & 0.166 & 0.362 \\

6 & IRZ\_CAP &
 \third{0.895} & \second{0.583} & \third{0.496} & %
 \second{0.986} & 0.809 & 0.579 & %
 0.130 & 0.010 & \third{0.679} & 0.280 & \third{0.441} \\

6 & IRZ\_N3 &
 0.890 & 0.567 & 0.484 & %
 0.969 & \third{0.810} & 0.582 & %
 0.098 & 0.001 & 0.669 & 0.260 & 0.432 \\

6 & N3\_CAP &
 0.890 & 0.581 & 0.495 & %
 0.977 & 0.786 & \third{0.600} & %
 0.120 & \first{0.015} & 0.674 & \first{0.293} & 0.439 \\

9 & IRZ\_N3\_CAP &
 \second{0.896} & \third{0.582} & \second{0.501} & %
 \third{0.982} & \second{0.815} & \second{0.606} & %
 \second{0.132} & \third{0.013} & \second{0.683} & 0.275 & \second{0.450} \\

12 & IRZ\_N3\_CAP\_PCA &
 \first{0.902} & \first{0.595} & \first{0.511} & %
 \first{0.990} & \first{0.833} & \first{0.640} & %
 \first{0.133} & 0.010 & \first{0.693} & \second{0.280} & \first{0.462} \\
        \bottomrule
    \end{tabular}}
    \par\vspace{0.5em}
    \begin{flushleft}
    \footnotesize\emph{Feature codes:}
    \textit{IRZ\_raw} = intensity, range, $Z$ (raw);
    \textit{I.R.Z} = pre-processed intensity, range, $Z_{\text{inv}}$;
    \textit{C.A.P} = curvature, anisotropy, planarity;
    \textit{N3} = pseudo-RGB from normals;
    \textit{PCA} = first three principal components of \textit{PCA};
    \textit{MNF} = first three components of \textit{MNF};
    \textit{ICA} = first three components of \textit{ICA};
    \textit{IRZ\_CAP} = I.R.Z + C.A.P;
    \textit{IRZ\_N3} = I.R.Z + N3;
    \textit{N3\_CAP} = N3 + C.A.P;
    \textit{IRZ\_N3\_CAP} = I.R.Z + N3 + C.A.P;
    \textit{IRZ\_N3\_CAP\_PCA} = IRZ\_N3\_CAP + PCA of that set.
    \end{flushleft}
    \label{tab:forestsemantic-feature-comparison}
\end{table*}

%% file: tex_of_fig_tab/fig_semantic3d_2d.tex
\begin{figure}[!htb]
    \centering

    \begin{subfigure}{0.82\textwidth}
        \includegraphics[width=\linewidth]{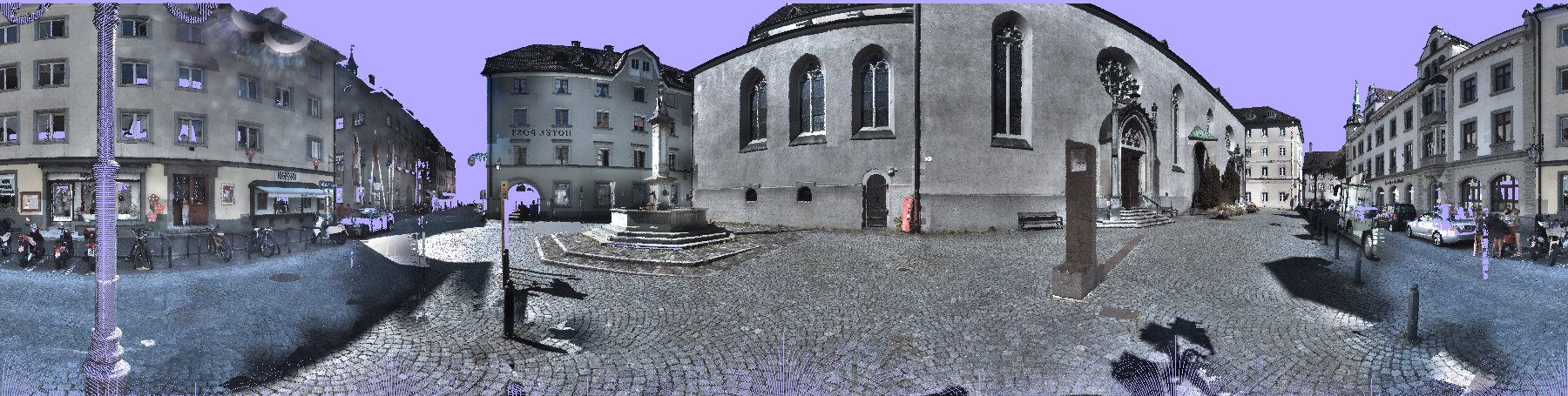}
        \caption{Original RGB}
    \end{subfigure}
    \vspace{0.5em}
    \begin{subfigure}{0.32\textwidth}
        \includegraphics[width=\linewidth]{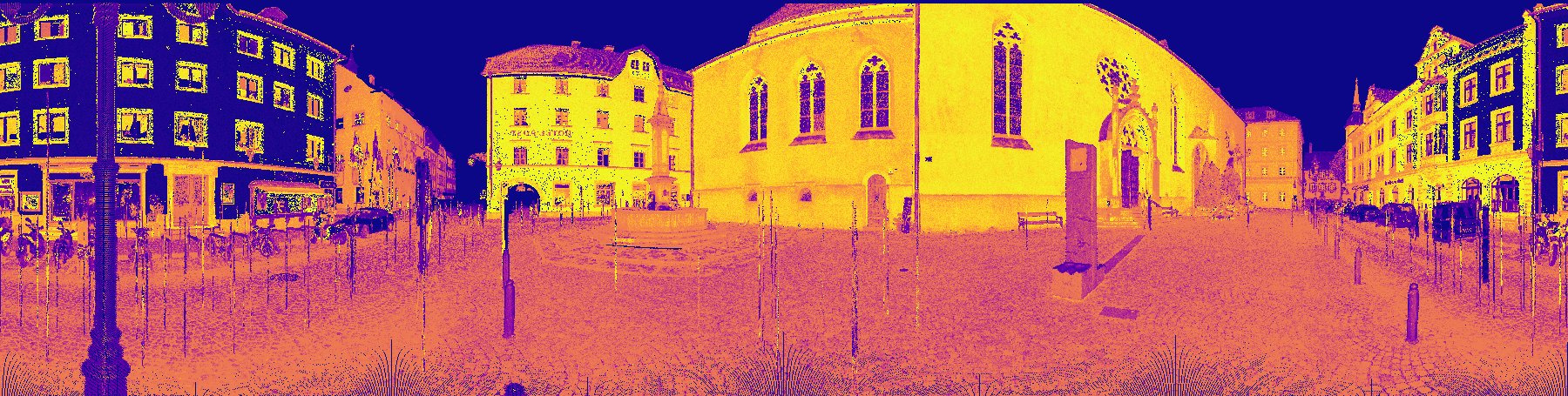}
        \caption{Intensity}
    \end{subfigure}
    \begin{subfigure}{0.32\textwidth}
        \includegraphics[width=\linewidth]{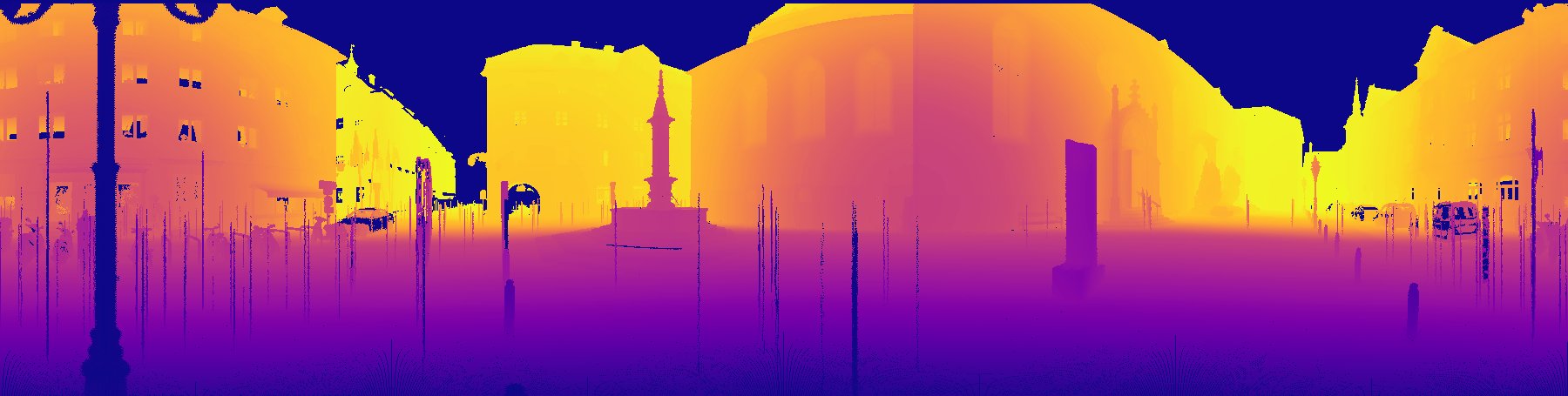}
        \caption{Range}
    \end{subfigure}
    \begin{subfigure}{0.32\textwidth}
        \includegraphics[width=\linewidth]{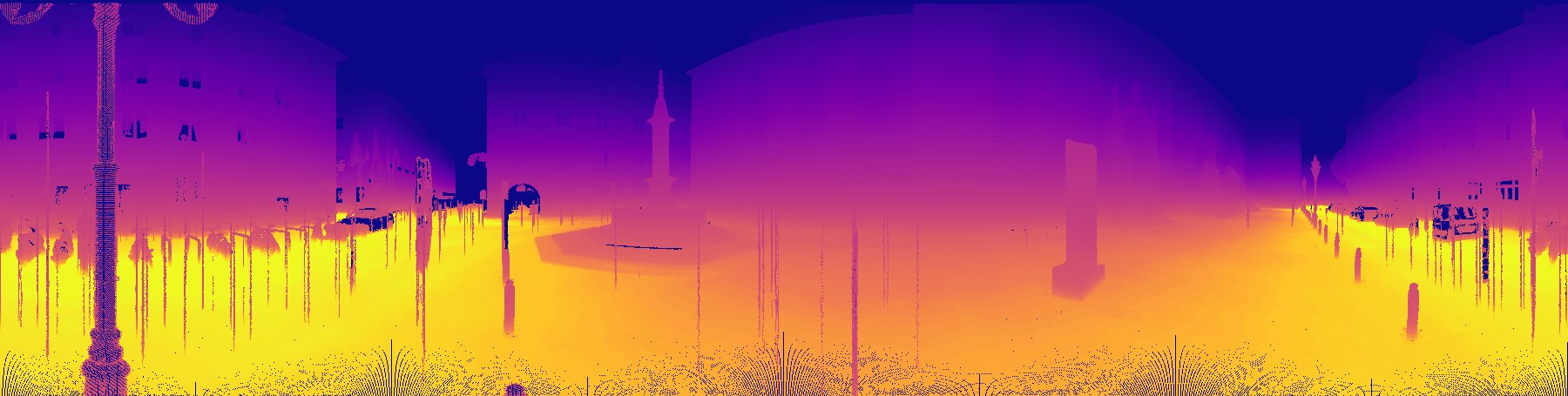}
        \caption{Z-Inv}
    \end{subfigure}

    \begin{subfigure}{0.32\textwidth}
        \includegraphics[width=\linewidth]{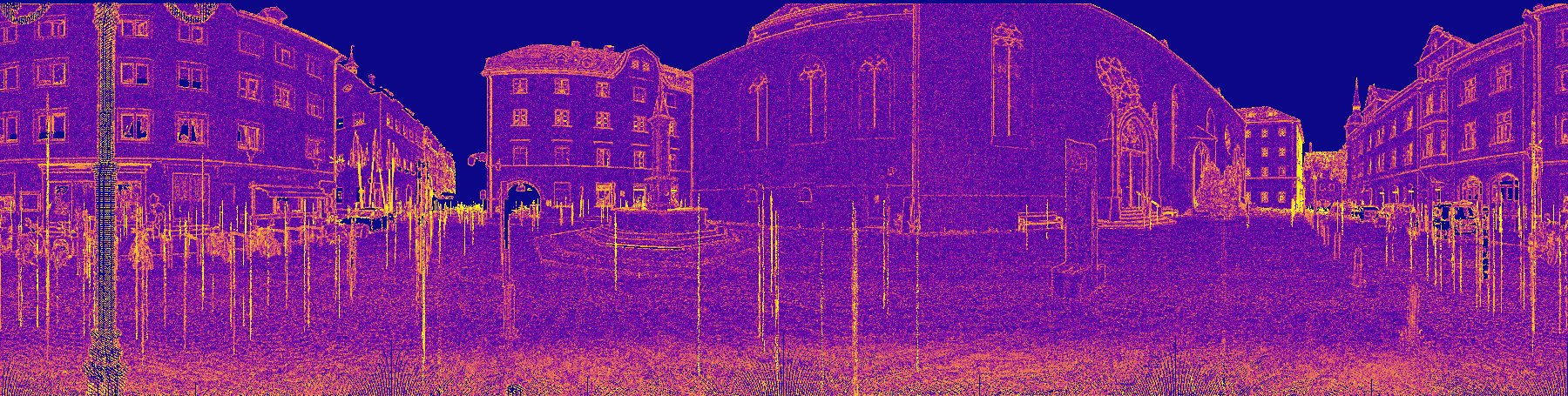}
        \caption{Anisotropy}
    \end{subfigure}
    \begin{subfigure}{0.32\textwidth}
        \includegraphics[width=\linewidth]{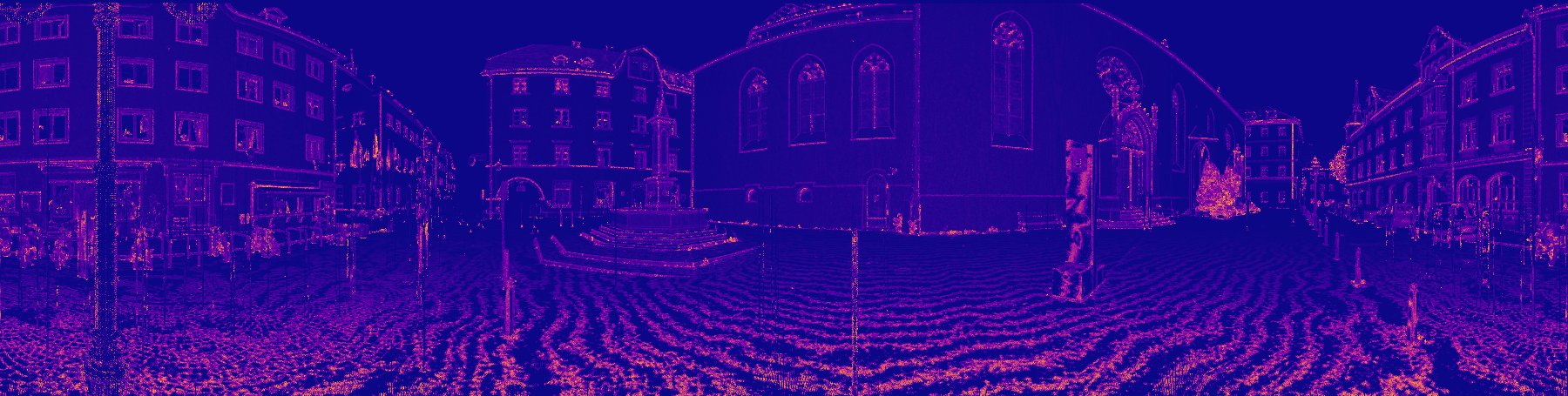}
        \caption{Curvature}
    \end{subfigure}
    \begin{subfigure}{0.32\textwidth}
        \includegraphics[width=\linewidth]{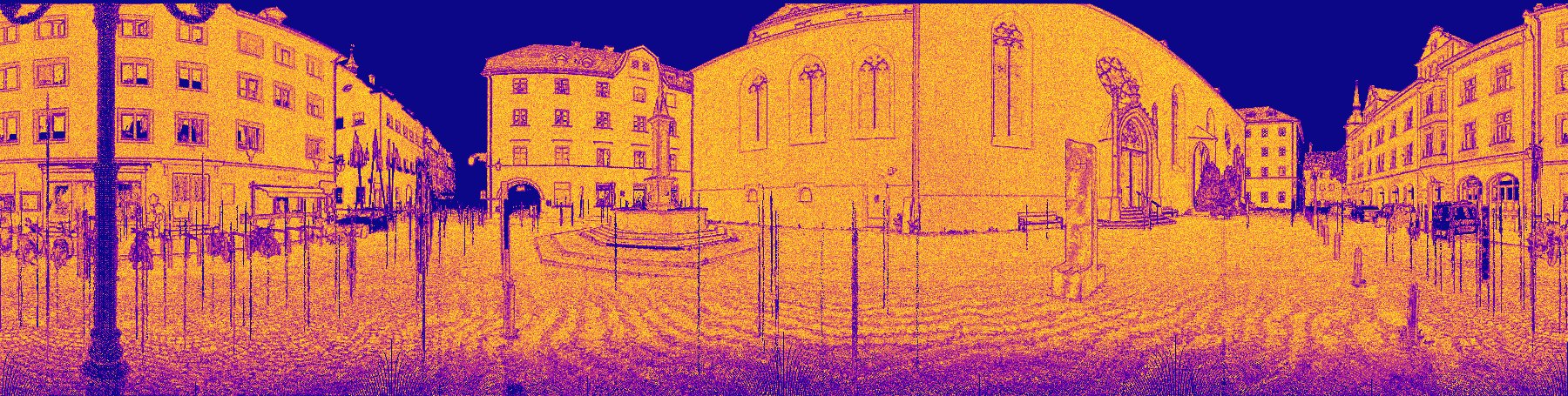}
        \caption{Planarity}
    \end{subfigure}

    \begin{subfigure}{0.32\textwidth}
        \includegraphics[width=\linewidth]{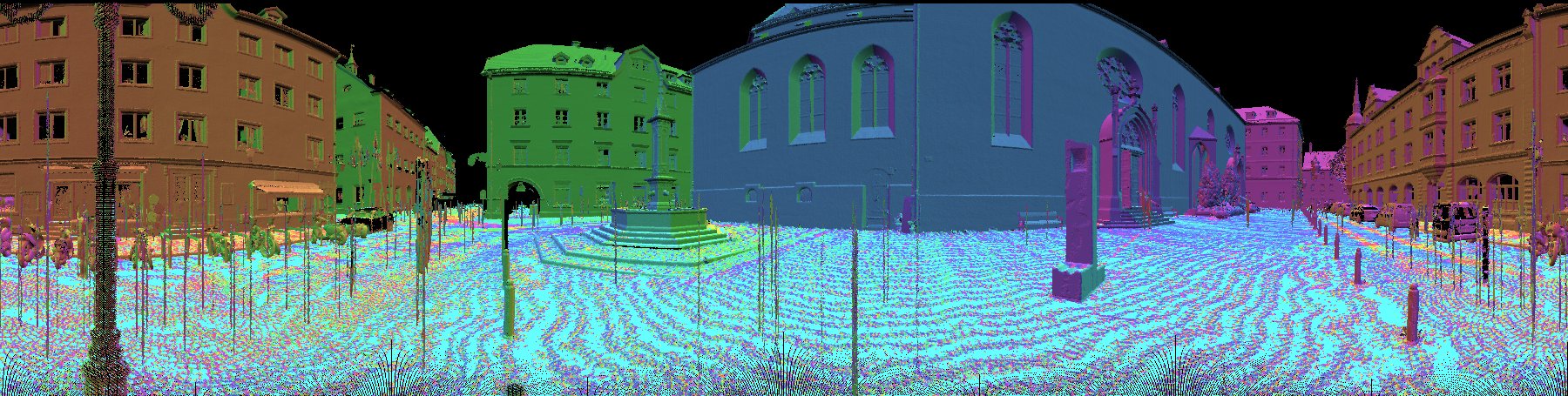}
        \caption{Normals pseudo color stack}
    \end{subfigure}
    \begin{subfigure}{0.32\textwidth}
        \includegraphics[width=\linewidth]{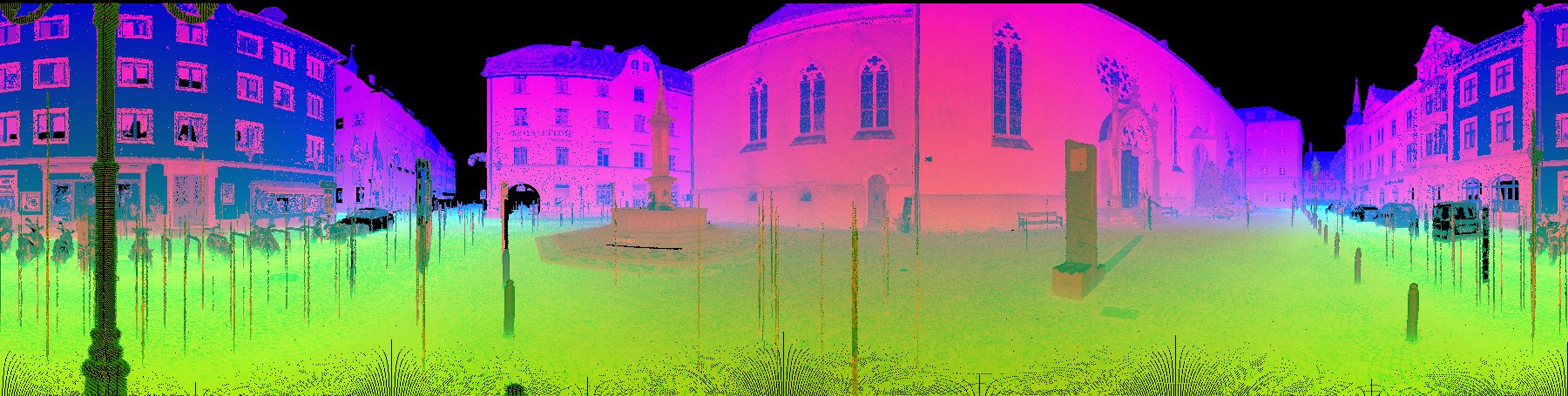}
        \caption{I.R.Z stack}
    \end{subfigure}
    \begin{subfigure}{0.32\textwidth}
        \includegraphics[width=\linewidth]{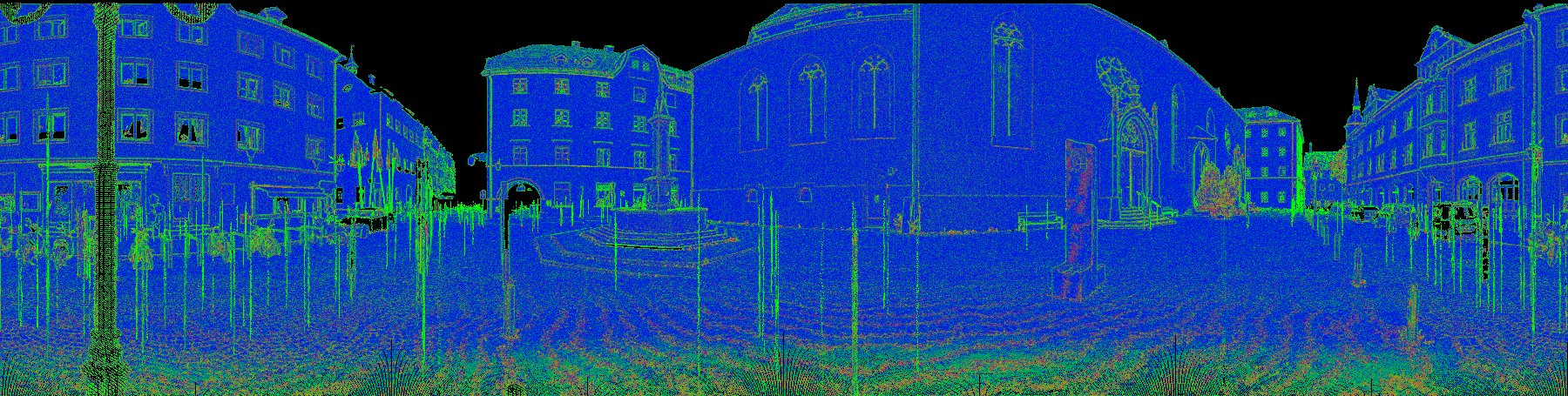}
        \caption{C.A.P stack}
    \end{subfigure}

    \begin{subfigure}{0.32\textwidth}
        \includegraphics[width=\linewidth]{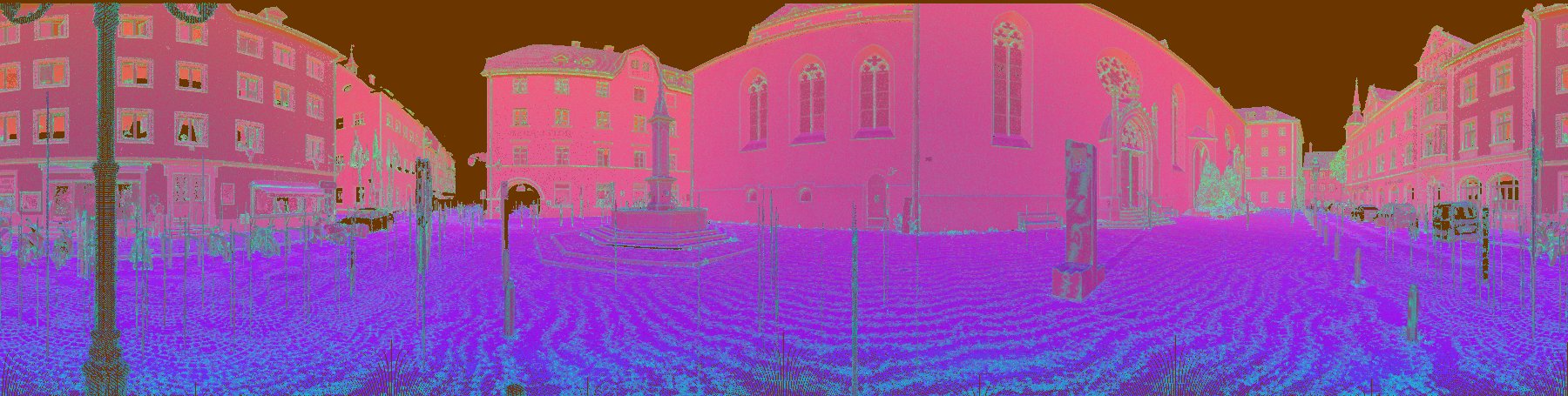}
        \caption{PCA stack}
    \end{subfigure}
    \begin{subfigure}{0.32\textwidth}
        \includegraphics[width=\linewidth]{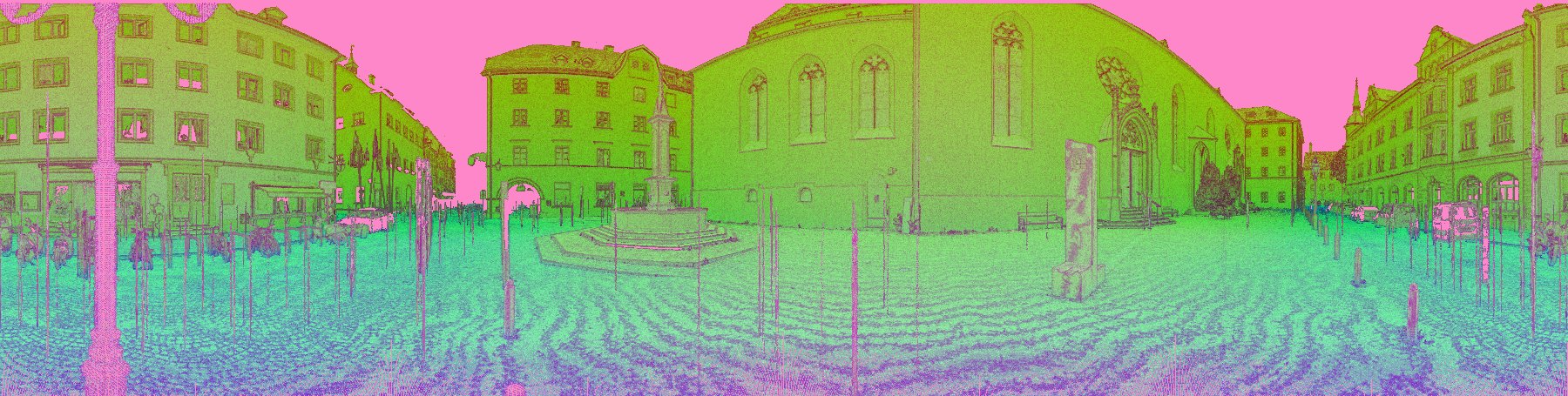}
        \caption{MNF stack}
    \end{subfigure}
    \begin{subfigure}{0.32\textwidth}
        \includegraphics[width=\linewidth]{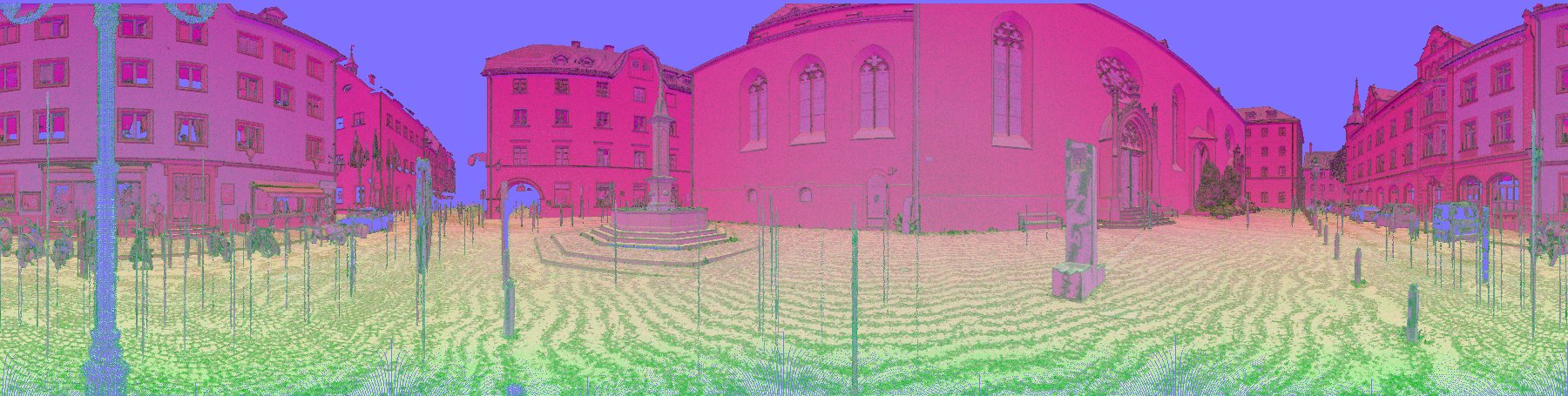}
        \caption{ICA stack}
    \end{subfigure}

    \vspace{0.5em}
    \begin{subfigure}[t]{0.66\textwidth}
        \centering
        \includegraphics[width=\linewidth]{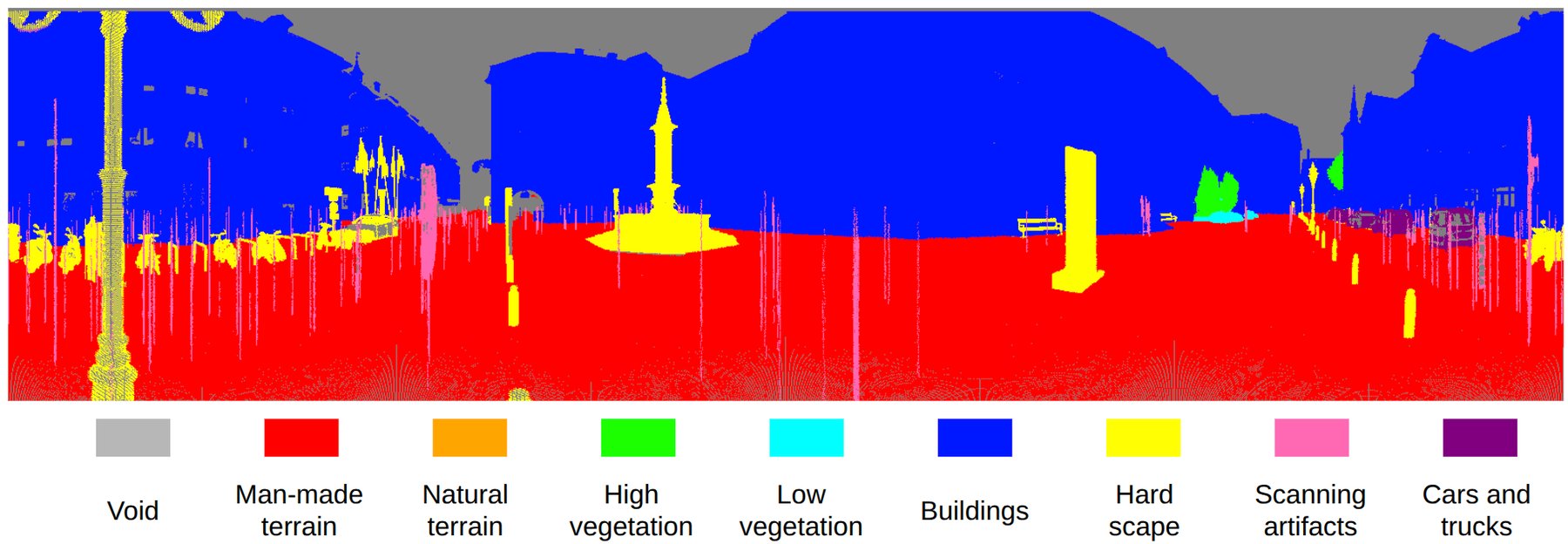}
        \caption{GT segmentation mask}
    \end{subfigure}
    \begin{subfigure}[t]{0.32\textwidth}
        \centering
        \includegraphics[width=\linewidth]{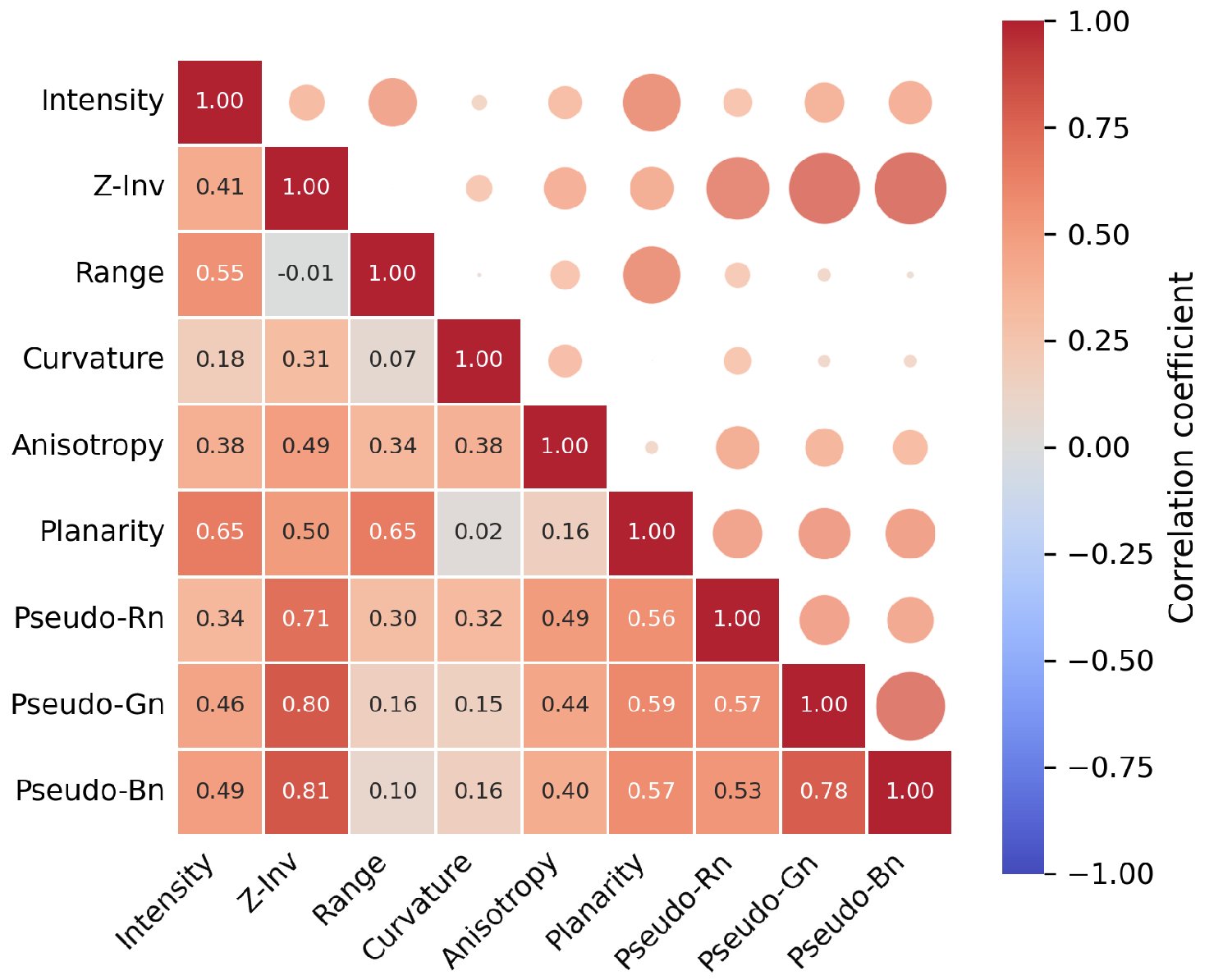}
        \caption{Correlation of 9 channels (a)-(g)}
    \end{subfigure}

    \caption{Spherical-projection maps of a \emph{Semantic3D} scan, organized by feature type.}
    \label{fig:Semantic3D_spherical_maps}
\end{figure}

%% file: tex_of_fig_tab/fig_semantic3d_pcd.tex
\begin{figure}[htbp]
    \centering
    \includegraphics[width=1\linewidth]{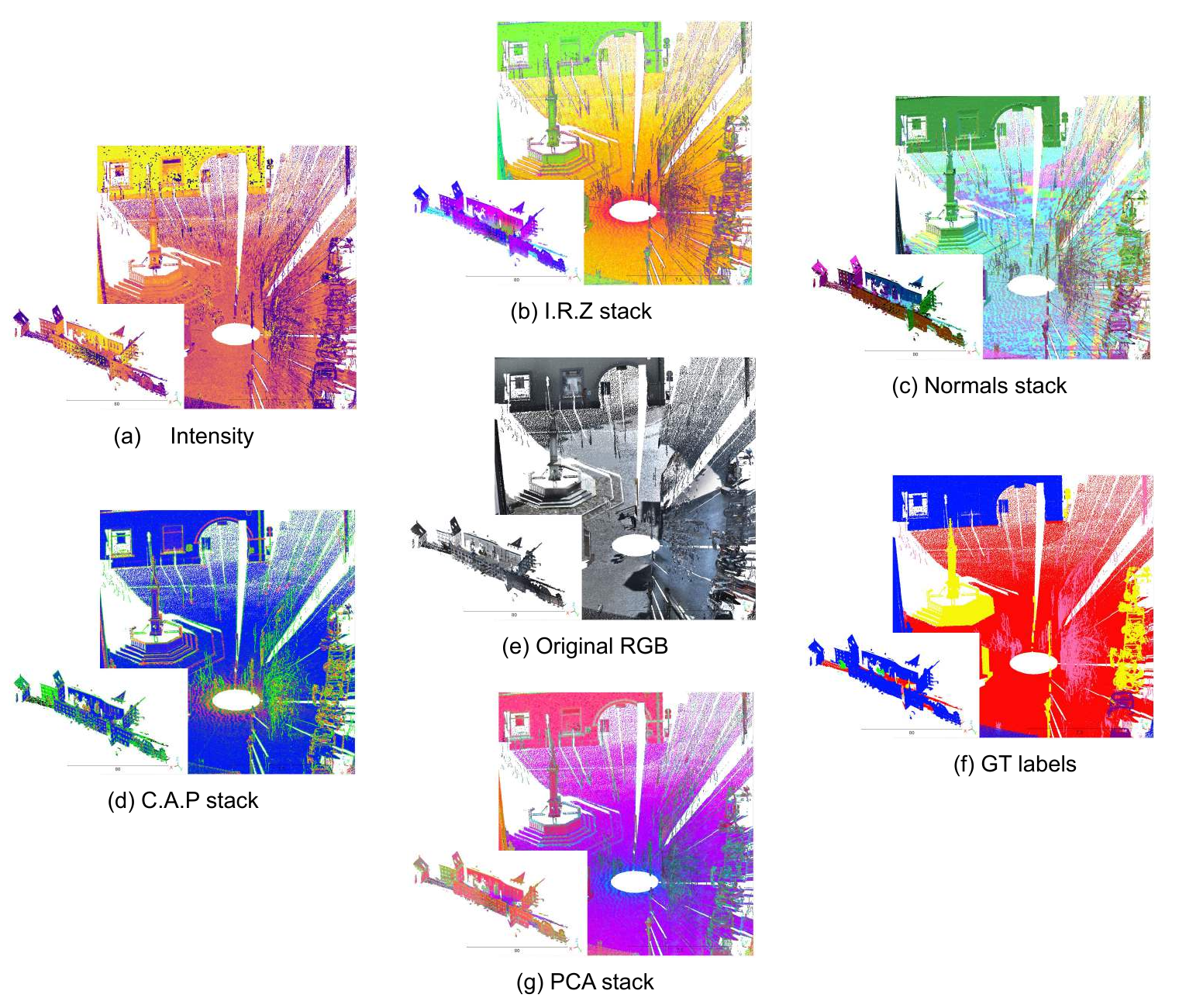}
    \caption{Feature-driven colorization of a \emph{Semantic3D} point cloud.
    Each inset shows a full overview of the entire plot to provide spatial context for the zoomed-in region.}
    \label{fig:semantic3d-pcd}
\end{figure}

%% file: tex_of_fig_tab/fig_semantic3d_balls.tex
\begin{figure}[htbp]
    \centering
    \begin{minipage}[b]{0.24\textwidth}
        \begin{subfigure}[b]{\textwidth}
            \centering
            \includegraphics[width=\textwidth]{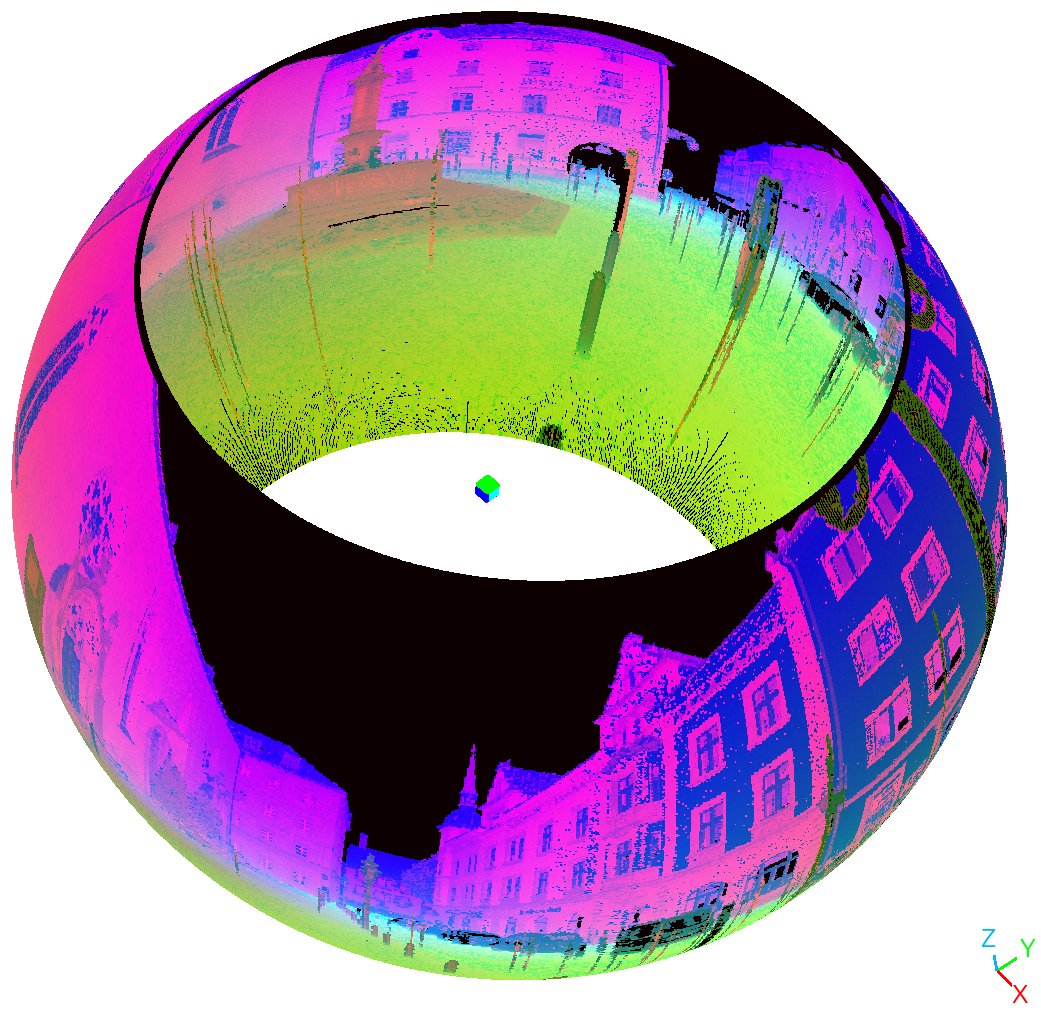}
            \caption*{(a) I.R.Z stack}
        \end{subfigure}
    \end{minipage}
    \begin{minipage}[b]{0.24\textwidth}
        \begin{subfigure}[b]{\textwidth}
            \centering
            \includegraphics[width=\textwidth]{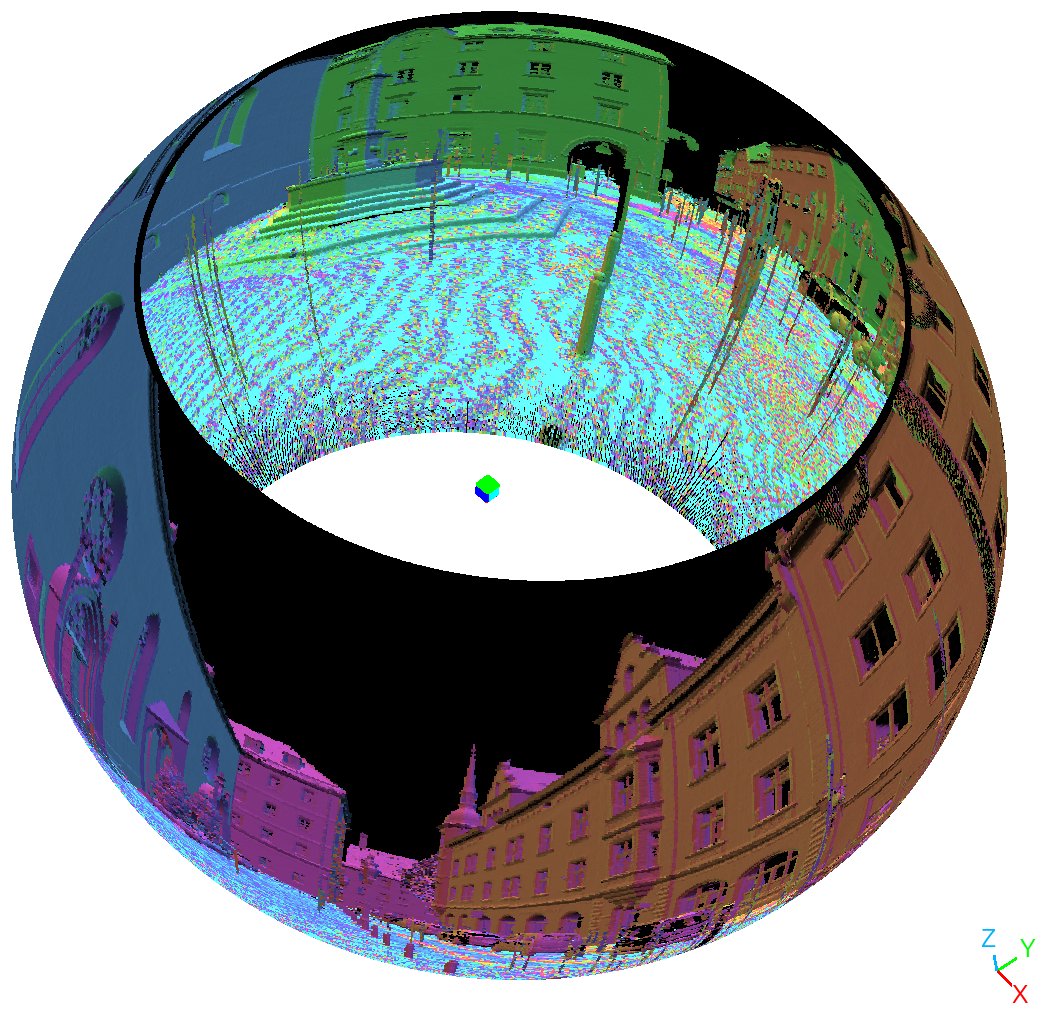}
            \caption*{(b) Normals stack}
        \end{subfigure}
    \end{minipage}
    \begin{minipage}[b]{0.24\textwidth}
        \begin{subfigure}[b]{\textwidth}
            \centering
            \includegraphics[width=\textwidth]{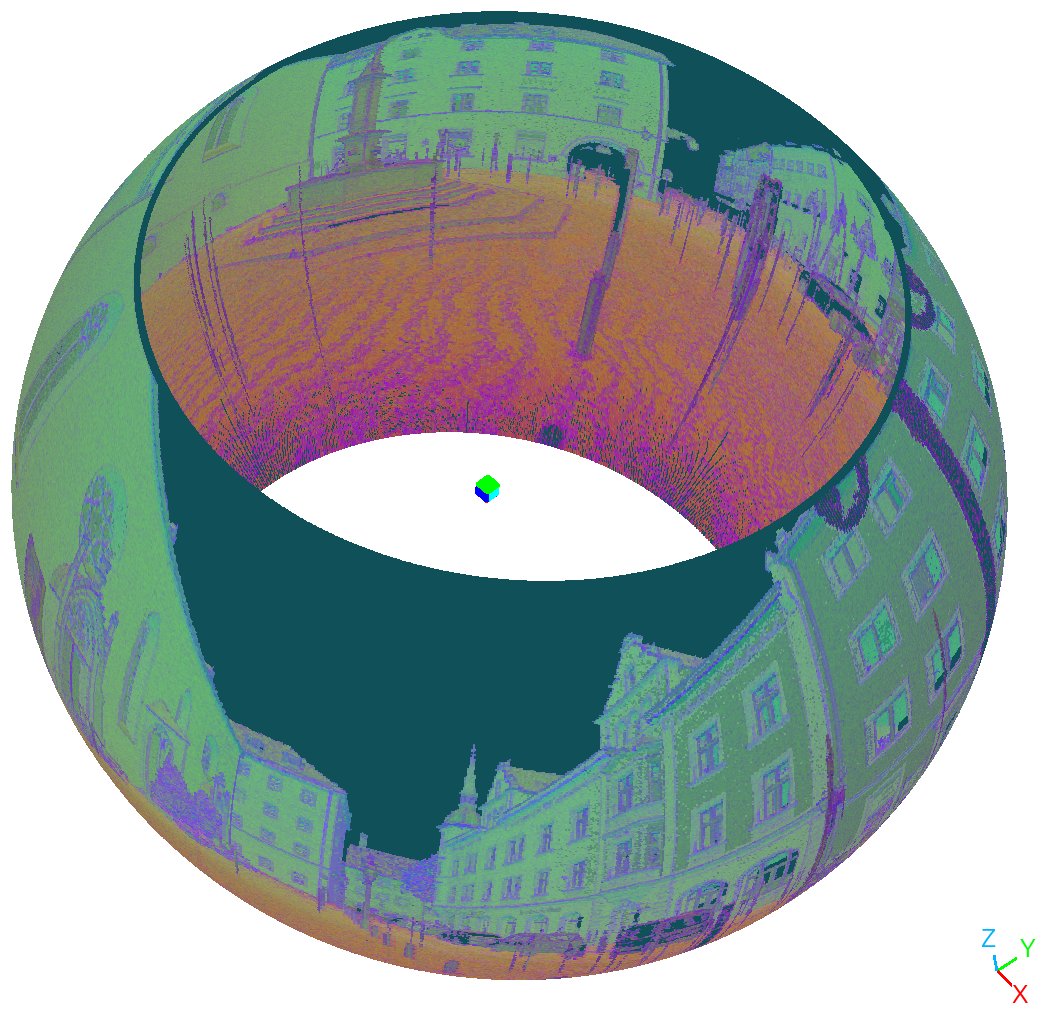}
            \caption*{(c) PCA stack}
        \end{subfigure}
    \end{minipage}
    \begin{minipage}[b]{0.24\textwidth}
        \begin{subfigure}[b]{\textwidth}
            \centering
            \includegraphics[width=\textwidth]{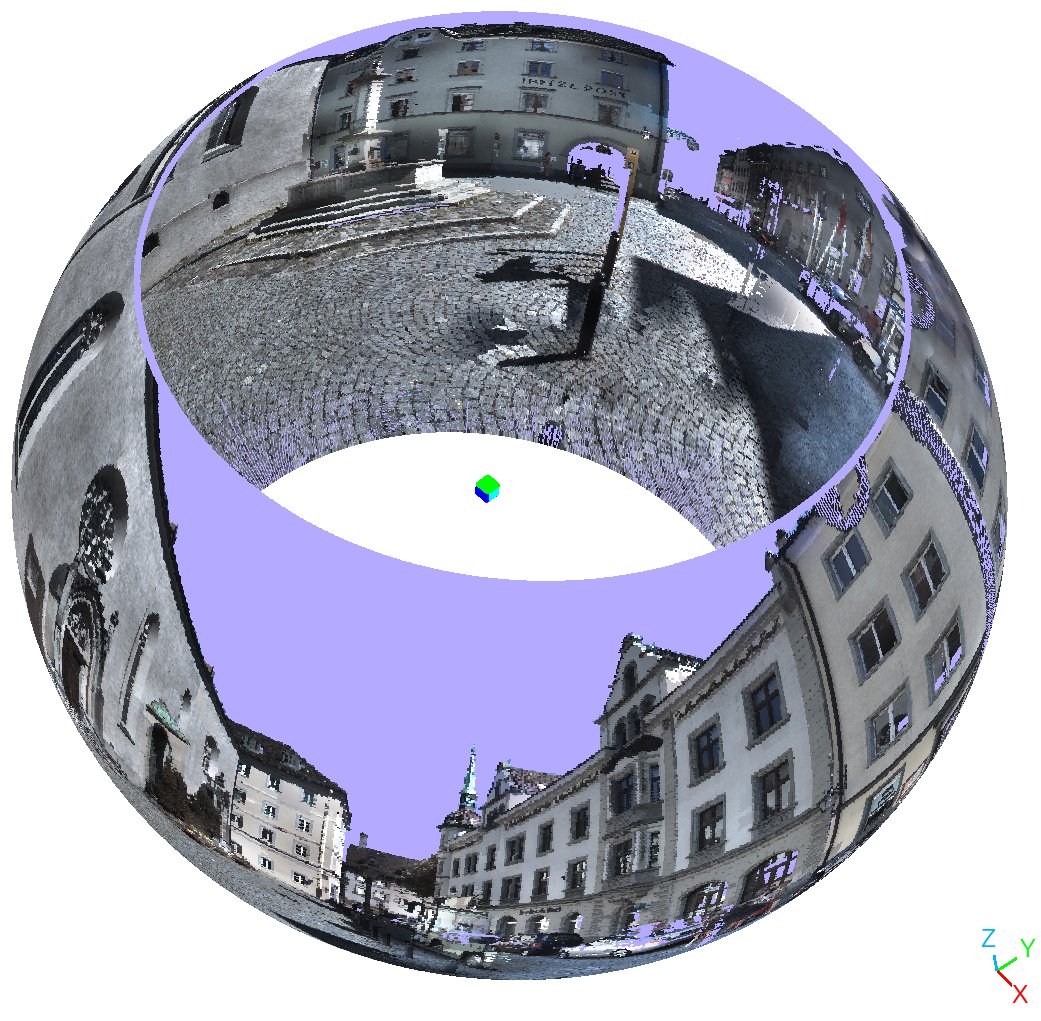}
            \caption*{(d) Original RGB}
        \end{subfigure}
    \end{minipage}
    
    \caption{Virtual spheres of an example \textit{Semantic3D} scan colorized with three different feature groups. Animation available in~\ref{app:animations}-Table~\ref{tab:animations}}
    \label{fig:semantic3d_balls}
\end{figure}

%% file: tex_of_fig_tab/tab_semantic3d_feature_compare.tex
\begin{table*}[htbp]
\centering
\footnotesize        
\caption{Comparison of feature-input combinations for semantic segmentation on \emph{Semantic3D}.  
Within each metric column, the top three scores are highlighted with progressively darker blue shading  
(dark = best, medium = second, light = third).}
\vspace{0.4em}
\resizebox{\textwidth}{!}{
\begin{tabular}{ccccc|ccccccccc|c}
\toprule
\multirow{2}{*}{\textbf{Feat.\#}} &
\multirow{2}{*}{\textbf{Feature}} &
\multicolumn{3}{c|}{\textbf{Global Metrics}} &
\multicolumn{9}{c|}{\textbf{IoU per class}} &
\multirow{2}{*}{\shortstack[c]{\textbf{mIoU}\\\textbf{Void excl.}}} \\
\cmidrule(lr){3-5}\cmidrule(lr){6-14}
& &
\textbf{oAcc.} & \textbf{mAcc.} & \textbf{mIoU} &
\textbf{Void} &
\shortstack{\textbf{Man-}\\\textbf{made terr.}} &
\shortstack{\textbf{Natural}\\\textbf{terr.}} &
\shortstack{\textbf{High}\\\textbf{veg.}} &
\shortstack{\textbf{Low}\\\textbf{veg.}} &
\textbf{Buildings} &
\shortstack{\textbf{Remain.}\\\textbf{hardscape}} &
\shortstack{\textbf{Scanning}\\\textbf{artifacts}} &
\shortstack{\textbf{Cars \&}\\\textbf{trucks}} & \\
\midrule

\rowcolor{black!6}
3 & RGB &
0.782 & 0.477 & 0.388 &
0.763 & 0.581 & 0.463 & 0.293 & \third{0.215} &
0.755 & 0.141 & 0.036 & 0.245 & 0.341 \\

3 & irz\_raw &
0.746 & 0.383 & 0.300 &
0.780 & 0.497 & 0.214 & 0.255 & 0.046 &
0.694 & 0.098 & 0.056 & 0.058 & 0.240 \\

3 & I.R.Z &
0.783 & 0.509 & 0.411 &
0.778 & 0.569 & 0.428 & 0.360 & 0.072 &
0.763 & 0.133 & 0.313 & 0.285 & 0.365 \\

3 & C.A.P &
0.807 & 0.513 & 0.422 &
0.817 & 0.604 & 0.425 & 0.398 & 0.154 &
0.803 & 0.178 & 0.216 & 0.204 & 0.436 \\

3 & N3 &
0.799 & 0.545 & 0.451 &
0.764 & 0.597 & 0.509 & \third{0.440} & 0.168 &
0.799 & 0.139 & 0.300 & 0.342 & 0.412 \\

3 & PCA &
0.777 & 0.461 & 0.381 &
0.757 & 0.569 & 0.373 & 0.370 & 0.089 &
0.780 & 0.112 & 0.178 & 0.199 & 0.334 \\

3 & MNF &
0.730 & 0.457 & 0.351 &
0.719 & 0.471 & 0.356 & 0.316 & 0.191 &
0.709 & 0.106 & 0.168 & 0.119 & 0.305 \\

3 & ICA &
0.741 & 0.481 & 0.375 &
0.694 & 0.560 & 0.326 & 0.343 & 0.167 &
0.715 & 0.116 & 0.242 & 0.214 & 0.335 \\

6 & IRZ\_CAP &
0.822 & 0.579 & 0.480 &
0.828 & 0.640 & 0.476 & 0.419 & 0.202 &
0.832 & 0.176 & \third{0.360} & 0.386 & 0.436 \\

6 & IRZ\_N3 &
\second{0.848} & \second{0.589} & \second{0.504} &
\second{0.854} & \first{0.698} & \first{0.584} & \second{0.465} & 0.160 &
0.836 & 0.157 & \second{0.375} & \second{0.411} & \second{0.461} \\

6 & N3\_CAP &
0.832 & 0.576 & 0.490 &
0.813 & 0.629 & \second{0.539} & 0.584 & \second{0.217} &
\third{0.841} & \third{0.199} & 0.233 & 0.357 & 0.450 \\

9 & IRZ\_N3\_CAP &
\first{0.854} & \first{0.615} & \first{0.516} &
\first{0.878} & \second{0.690} & \third{0.532} & 0.421 & \first{0.225} &
\first{0.883} & \second{0.201} & \first{0.407} & \first{0.412} & \first{0.471} \\

12 & IRZ\_N3\_CAP\_PCA &
\third{0.836} & \third{0.585} & \third{0.496} &
\third{0.830} & \third{0.647} & 0.527 & \first{0.467} & \third{0.178} &
\second{0.857} & \first{0.206} & 0.357 & \third{0.390} & \third{0.454} \\

\bottomrule
\end{tabular}}
\par\vspace{0.5em}
\begin{flushleft}
\footnotesize\emph{Feature codes:}
\textit{IRZ\_raw} = intensity, range, $Z$ (raw);
\textit{I.R.Z} = pre-processed intensity, range, $Z_{\text{inv}}$;
\textit{C.A.P} = curvature, anisotropy, planarity;
\textit{N3} = pseudo-RGB from normals;
\textit{PCA} = first three principal components of \textit{PCA};
\textit{MNF} = first three components of \textit{MNF};
\textit{ICA} = first three components of \textit{ICA};
\textit{IRZ\_CAP} = I.R.Z + C.A.P;
\textit{IRZ\_N3} = I.R.Z + N3;
\textit{N3\_CAP} = N3 + C.A.P;
\textit{IRZ\_N3\_CAP} = I.R.Z + N3 + C.A.P;
\textit{IRZ\_N3\_CAP\_PCA} = IRZ\_N3\_CAP + PCA of that set.
\end{flushleft}
\label{tab:semantic3d-feature-comparison}
\end{table*}

%% file: Supple-Mater0.tex
\section{Annotation Guideline for Mangrove3D Dataset}
\label{app:mangrove3d-dataset}

The annotation guideline for making semantic segmentation masks of spherical maps of the \emph{Mangrove3D} dataset can be accessed via this \href{https://drive.google.com/file/d/1S8VgYlTX0ORs9HeCIrMU0g3UhkRZFUlP/view?usp=drive_link}{link}.

%% file: Supple-Mater1.tex
\section{Supplementary Material for Stage 1}\label{app:stage1}
\paragraph{Preprocessing details}
Table~S1 lists preprocessing steps for all feature maps, including intensity, range, Z-inv, anisotropy, curvature, planarity, and statistical stacks.

\input{tex_of_fig_tab/tab_preprocessing_feat_map}

\paragraph{Additional feature maps}
Fig.~\ref{fig:feature_maps_single_mangrove3d} illustrates single-channel maps from scan \#\texttt{site1\_01} of the Mangrove3D dataset (intensity, range, Z-inv, anisotropy, curvature, planarity).  

\input{tex_of_fig_tab/fig_feature_maps_single_mangrove3d}

\paragraph{Eigenvalue-based Geometric Descriptors}

For each point $i$, the local covariance matrix was computed as
\begin{equation}
\mathbf{C}_i = \tfrac{1}{N-1}\sum_{j=1}^{N} 
(\mathbf{x}_j - \bar{\mathbf{x}})(\mathbf{x}_j - \bar{\mathbf{x}})^\top ,
\end{equation}
where $\mathbf{x}_j \in \mathbb{R}^3$ denotes the coordinates of the $j$-th neighbor, 
$\bar{\mathbf{x}}$ is the neighborhood centroid, and $N$ is the number of points in the neighborhood. 
For the \emph{Mangrove3D} dataset, we adopted a fixed neighborhood configuration with $neighbor\_radius = 0.06~m$ and $max\_neighbors = 50$, selected empirically to balance robustness to point density variation and sensitivity to local geometric structure.  

Eigen-decomposition of $\mathbf{C}_i$ yields ordered eigenvalues 
$\lambda_1 \leq \lambda_2 \leq \lambda_3$, which describe the variance along orthogonal axes. 
From these, we define three commonly used geometric descriptors:
\begin{equation}
\kappa = \frac{\lambda_1}{\lambda_1 + \lambda_2 + \lambda_3}, \quad
A = \frac{\lambda_3 - \lambda_2}{\lambda_3}, \quad
P = \frac{\lambda_2 - \lambda_1}{\lambda_3}.
\end{equation}

Here, curvature ($\kappa$) reflects how ``linear'' or ``planar'' a neighborhood is, 
approaching zero for flat surfaces and increasing when points are distributed 
along a sharp edge or corner.  
Anisotropy ($A$) quantifies the dominance of one principal axis over the others, 
with higher values indicating elongated, line-like structures (e.g., stems or branches).  
Planarity ($P$) measures the degree to which points lie within a 2D plane, 
with higher values indicating flat surfaces (e.g., ground or large leaves).  
These features provide compact measures of local surface geometry and have been 
widely used for point-cloud classification and segmentation~\cite{harshit2022geometric,drones5040104,ijgi10030187}.

\paragraph{Feature correlations}
Fig.~\ref{fig:mangrove3d-corr-mtx} presents the Pearson correlation matrix among feature maps, validating inter-feature relationships.  

\input{tex_of_fig_tab/fig_mangrove3d-corr-mtx}

\paragraph{Stacked channels}
Fig.~\ref{fig:feature_maps_stack_mangrove3d} shows pseudo-color images from stacked feature channels (I.R.Z, C.A.P, and normal-based encodings).  
\input{tex_of_fig_tab/fig_feature_maps_stack_mangrove3d}

\paragraph{Additional Dimensionality-Reduction Methods}
In addition to PCA, we explored two other dimensionality-reduction techniques commonly applied in remote sensing. 
\begin{itemize}
    \item \textbf{MNF}~\cite{green1988transformation}: a two-step transform that first whitens data using an estimated noise covariance matrix, then applies PCA on the noise-whitened data. The resulting components are ranked by signal-to-noise ratio. 
    \item \textbf{ICA}~\cite{hyvarinen2000independent}: a statistical technique that seeks latent components which are mutually independent and non-Gaussian, thus capturing higher-order relationships beyond variance and correlation.
\end{itemize}\label{app:mnf-ica}

%% file: tex_of_fig_tab/tab_preprocessing_feat_map.tex

\begin{table}[htbp]
  \centering
  \scriptsize
  \caption{Preprocessing steps for the feature maps shown in
  Figures~\ref{fig:feature_maps_single_mangrove3d} and~\ref{fig:feature_maps_stack_mangrove3d}.}
  \label{tab:preprocessing_feat_map}
  \setlength{\tabcolsep}{3.5pt}
  \renewcommand{\arraystretch}{1.05}
  \begin{tabularx}{0.98\linewidth}{
      L{2.0cm}     
      L{1.55cm}    
      L{2.0cm}     
      L{2.9cm}     
      Y            
  }
    \toprule
    \textbf{Feature Group} & \textbf{Fig.} & \textbf{Map} & \textbf{Meaning} & \textbf{Preprocessing} \\
    \midrule

    \multirow{3}{*}{\makecell[tl]{Basic}}
      & Fig.~\ref{fig:feature_maps_single_mangrove3d}(a) & Intensity
      & Radiometric intensity
      & \multirow{2}{*}{Global histogram stretch; normalize to [0.01, 1.0]} \\
      & Fig.~\ref{fig:feature_maps_single_mangrove3d}(b) & Range
      & Distance from scanner & \\
      & Fig.~\ref{fig:feature_maps_single_mangrove3d}(c) & Z-Inv
      & Inverted height above ground
      & $H=Z-Z_{\min}$; negate $H$; normalize to [0.01, 1.0] \\
    \midrule

    \multirow{3}{*}{\makecell[tl]{Geometric}}
      & Fig.~\ref{fig:feature_maps_single_mangrove3d}(d) & Anisotropy
      & Directional structure
      & \multirow{3}{*}{None} \\
      & Fig.~\ref{fig:feature_maps_single_mangrove3d}(e) & Curvature
      & Local surface dev. & \\
      & Fig.~\ref{fig:feature_maps_single_mangrove3d}(f) & Planarity
      & Planar alignment & \\
    \midrule

    \multirow{3}{*}{\makecell[tl]{Direct Stack}}
      & Fig.~\ref{fig:feature_maps_stack_mangrove3d}(a) & Normals
      & X/Y/Z of surface normals
      & Convert to azimuth/elevation; map to HSV (azimuth $\to$ hue, elevation $\to$ value, saturation = 0.6); convert to RGB \\
      & Fig.~\ref{fig:feature_maps_stack_mangrove3d}(b) & I.R.Z stack
      & Intensity, Range, Z-Inv $\to$ R/G/B
      & Channel order: I$\to$R, Rg$\to$G, Z$\to$B \\
      & Fig.~\ref{fig:feature_maps_stack_mangrove3d}(c) & C.A.P stack
      & Curvature, Anisotropy, Planarity $\to$ R/G/B
      & Channel order: C$\to$R, A$\to$G, P$\to$B \\
    \midrule

    \multirow{3}{*}{\makecell[tl]{Statistical Stack}}
      & Fig.~\ref{fig:feature_maps_stack_mangrove3d}(d) & PCA
      & First 3 PCs
      & \multirow{3}{*}{Normalize each to [0,1]; stack to RGB} \\
      & Fig.~\ref{fig:feature_maps_stack_mangrove3d}(e) & MNF
      & First 3 MNFs & \\
      & Fig.~\ref{fig:feature_maps_stack_mangrove3d}(f) & ICA
      & First 3 ICs & \\
    \bottomrule
  \end{tabularx}
\end{table}

%% file: tex_of_fig_tab/fig_feature_maps_single_mangrove3d.tex
\begin{figure}[htbp]
    \centering
    \begin{minipage}[b]{0.48\textwidth}
        \begin{subfigure}[b]{\textwidth}
            \centering
            \includegraphics[width=\textwidth]{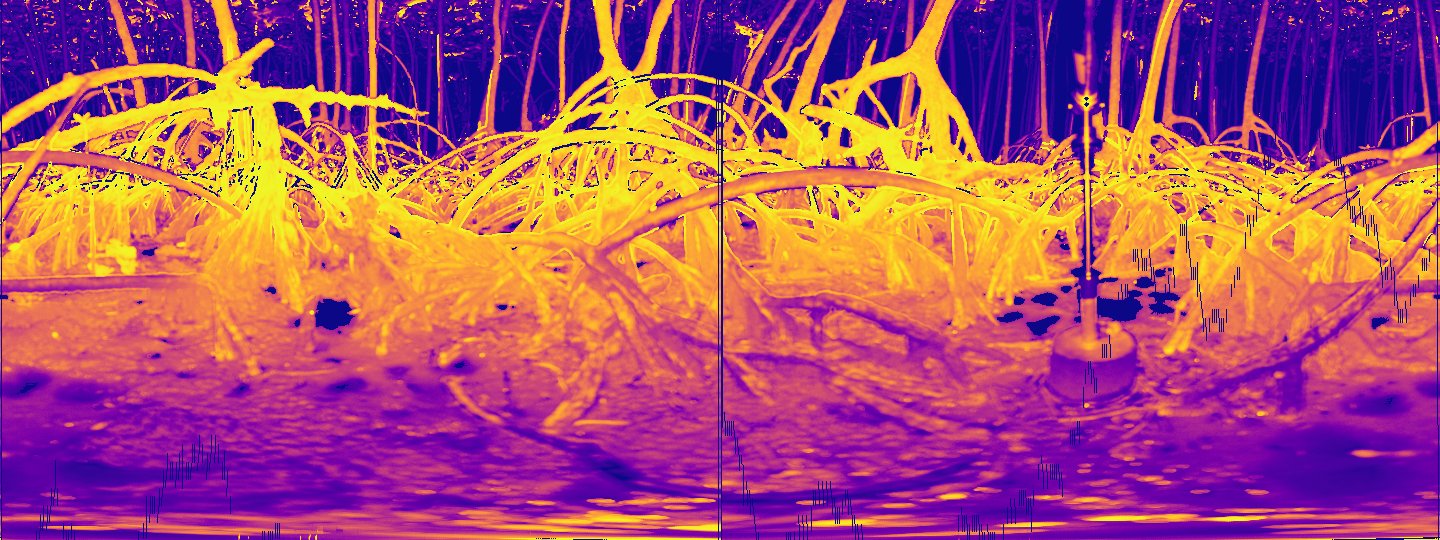}
            \caption*{(a) Intensity map}
        \end{subfigure}
        \vspace{0.5em}
        \begin{subfigure}[b]{\textwidth}
            \centering
            \includegraphics[width=\textwidth]{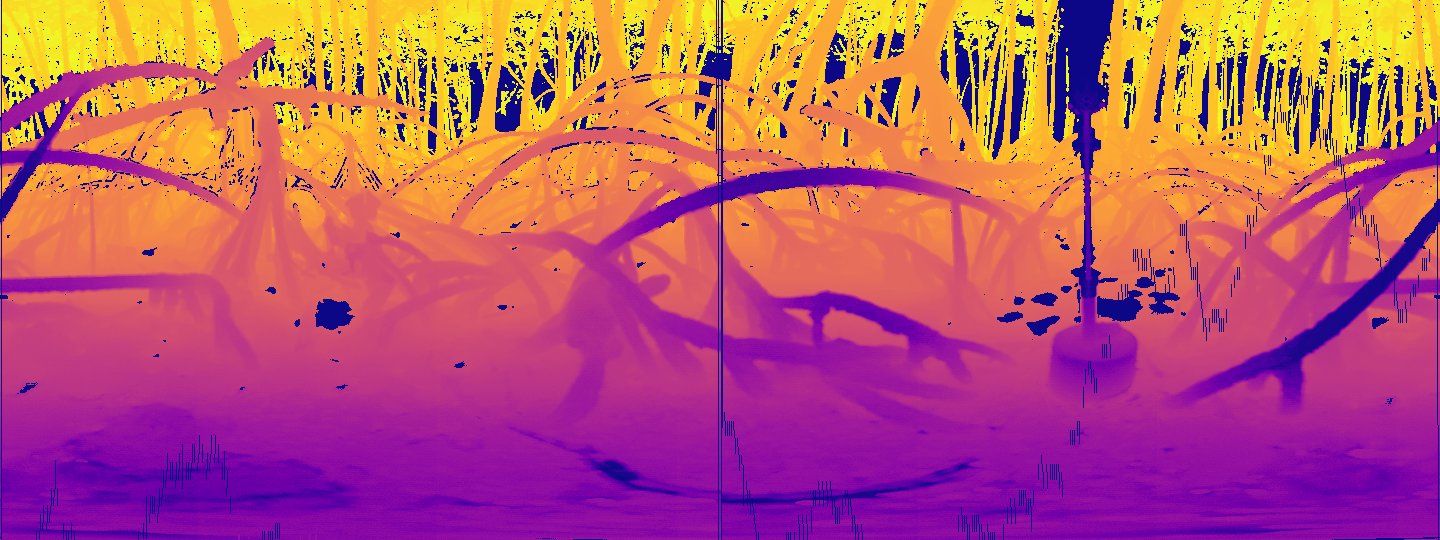}
            \caption*{(b) Range map}
        \end{subfigure}
        \vspace{0.5em}
        \begin{subfigure}[b]{\textwidth}
            \centering
            \includegraphics[width=\textwidth]{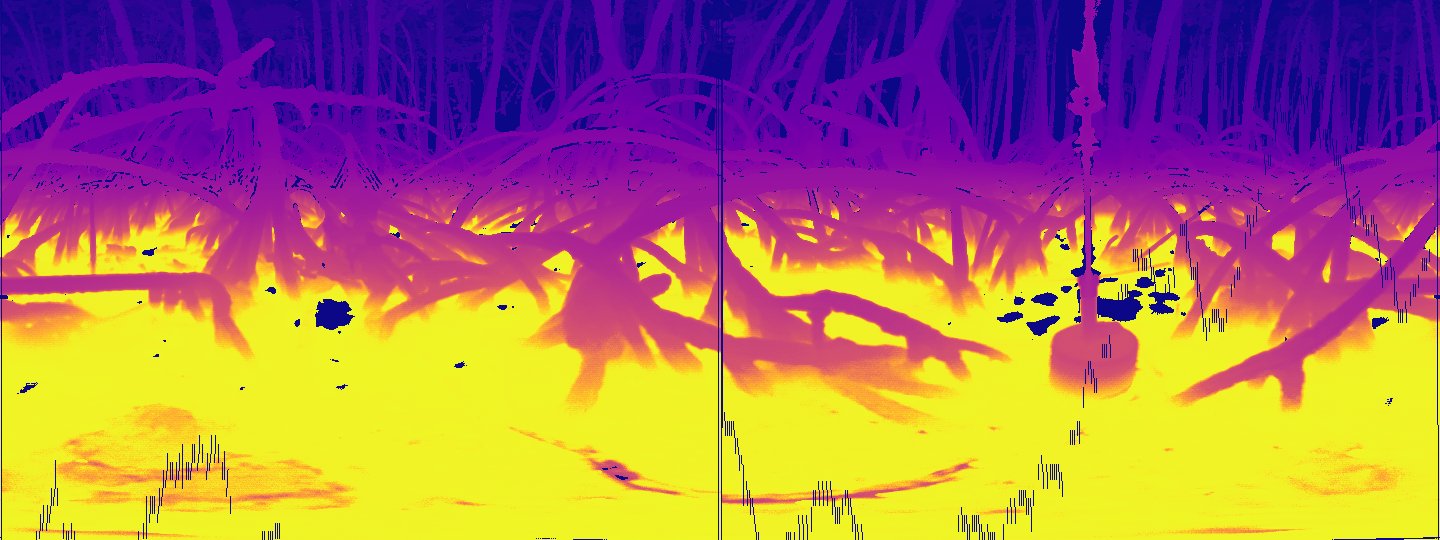}
            \caption*{(c) Z-Inverse map}
        \end{subfigure}
    \end{minipage}
    \hfill
    \begin{minipage}[b]{0.48\textwidth}
        \begin{subfigure}[b]{\textwidth}
            \centering
            \includegraphics[width=\textwidth]{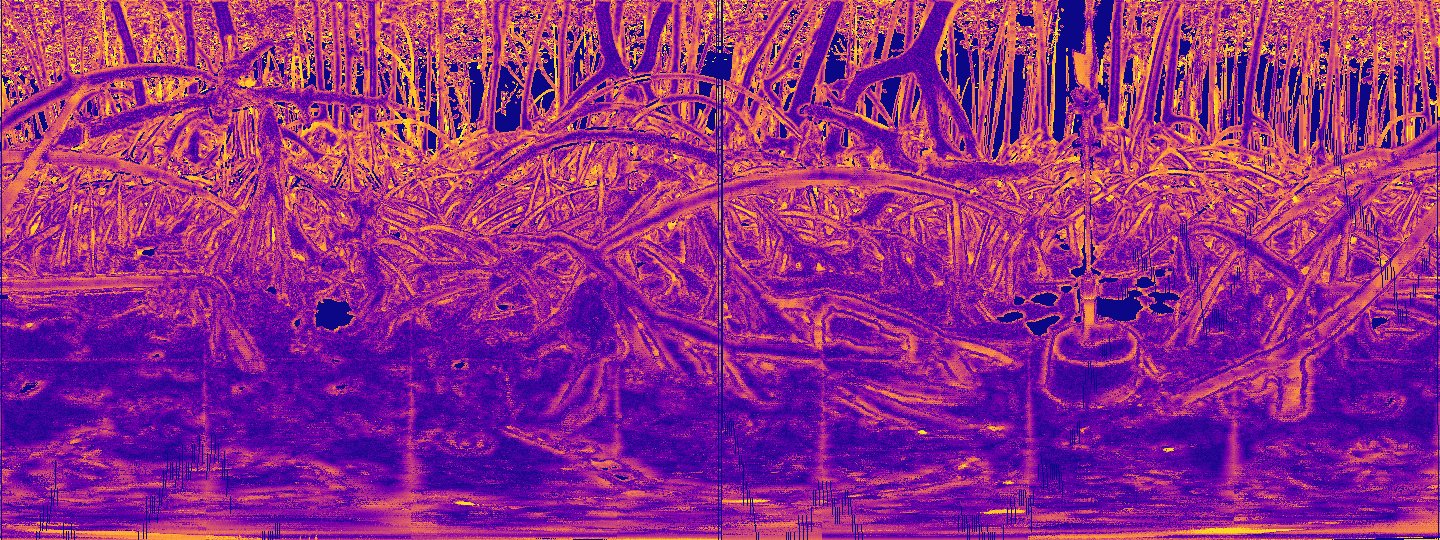}
            \caption*{(d) Anisotropy map}
        \end{subfigure}
        \vspace{0.5em}
        \begin{subfigure}[b]{\textwidth}
            \centering
            \includegraphics[width=\textwidth]{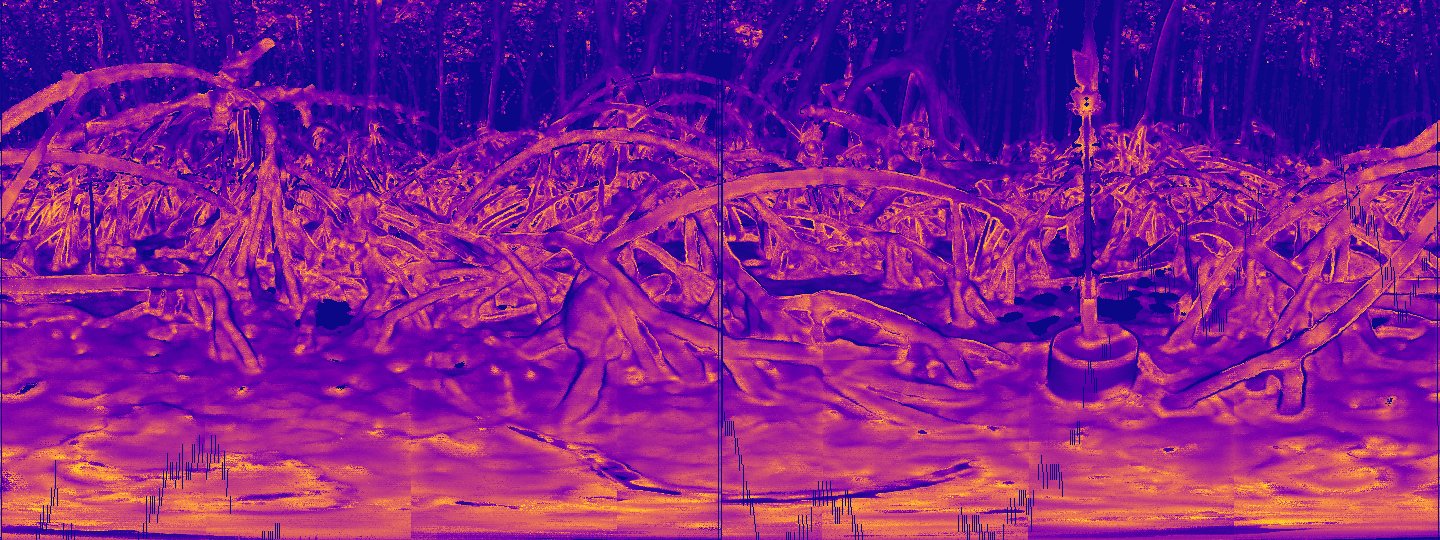}
            \caption*{(e) Curvature map}
        \end{subfigure}
        \vspace{0.5em}
        \begin{subfigure}[b]{\textwidth}
            \centering
            \includegraphics[width=\textwidth]{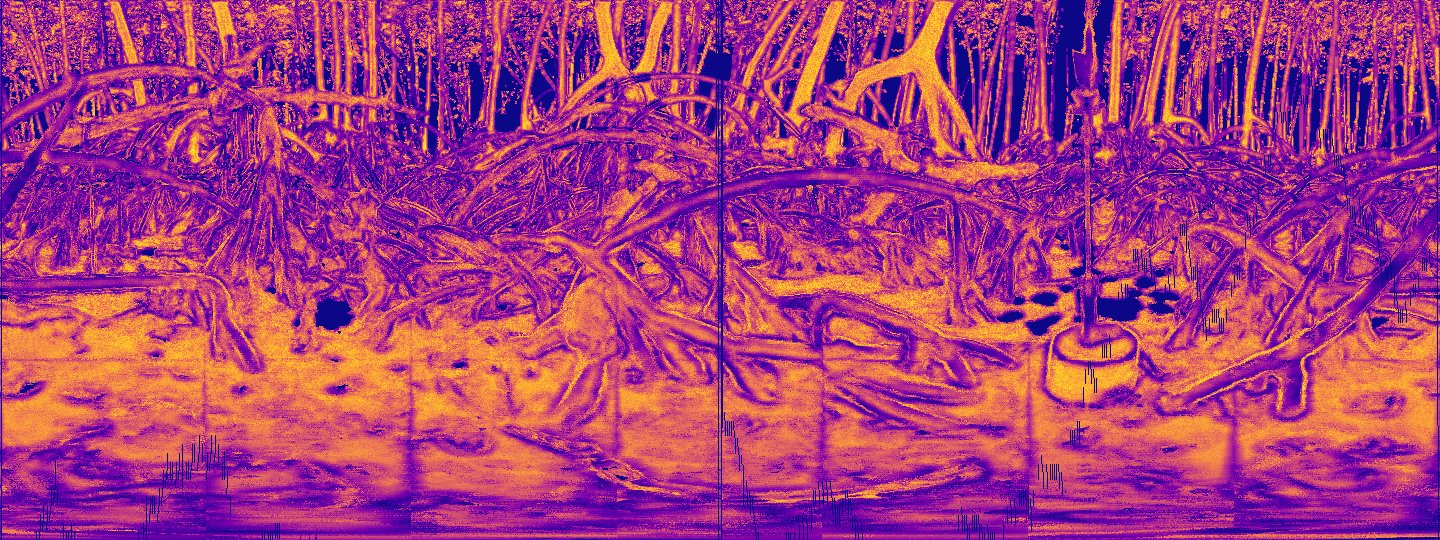}
            \caption*{(f) Planarity map}
        \end{subfigure}
    \end{minipage}

    \caption{Single channel feature maps generated from the 2D spherical projection of the terrestrial LiDAR point cloud. Each map represents a different basic or geometric attribute. For visualization consistency, all maps were normalized to the [0,1] range and visualized using the plasma colormap, where deep blue indicates low values and bright yellow denotes high values. }
    \label{fig:feature_maps_single_mangrove3d}
\end{figure}

%% file: tex_of_fig_tab/fig_mangrove3d-corr-mtx.tex
\begin{figure}[htbp]
    \centering
    \includegraphics[width=0.6\textwidth]{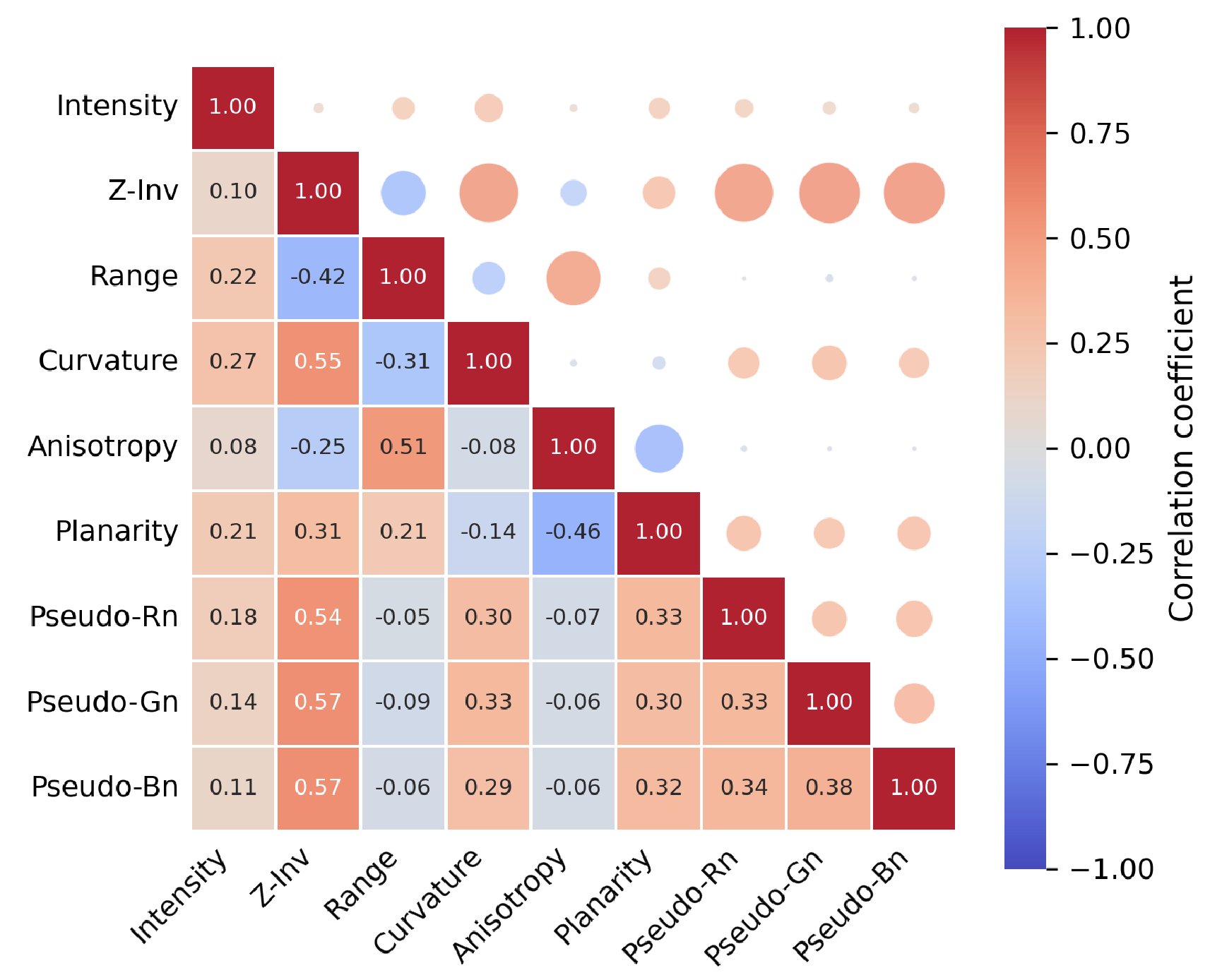}
    \caption{Correlation matrix of the feature maps.}
    \label{fig:mangrove3d-corr-mtx}
\end{figure}

%% file: tex_of_fig_tab/fig_feature_maps_stack_mangrove3d.tex
\begin{figure}[htbp]
    \centering
    \begin{minipage}[b]{0.48\textwidth}
        \begin{subfigure}[b]{\textwidth}
            \centering
            \includegraphics[width=\textwidth]{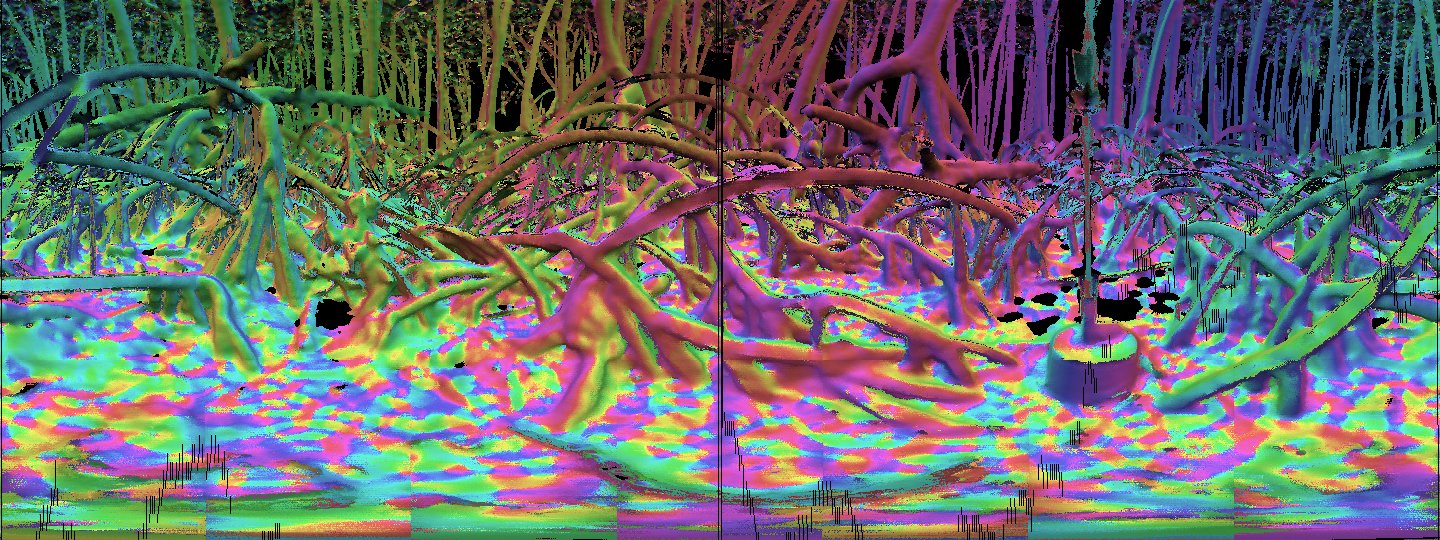}
            \caption*{(a) Pseudo-RGB from normals}
        \end{subfigure}
        \vspace{0.5em}
        \begin{subfigure}[b]{\textwidth}
            \centering
            \includegraphics[width=\textwidth]{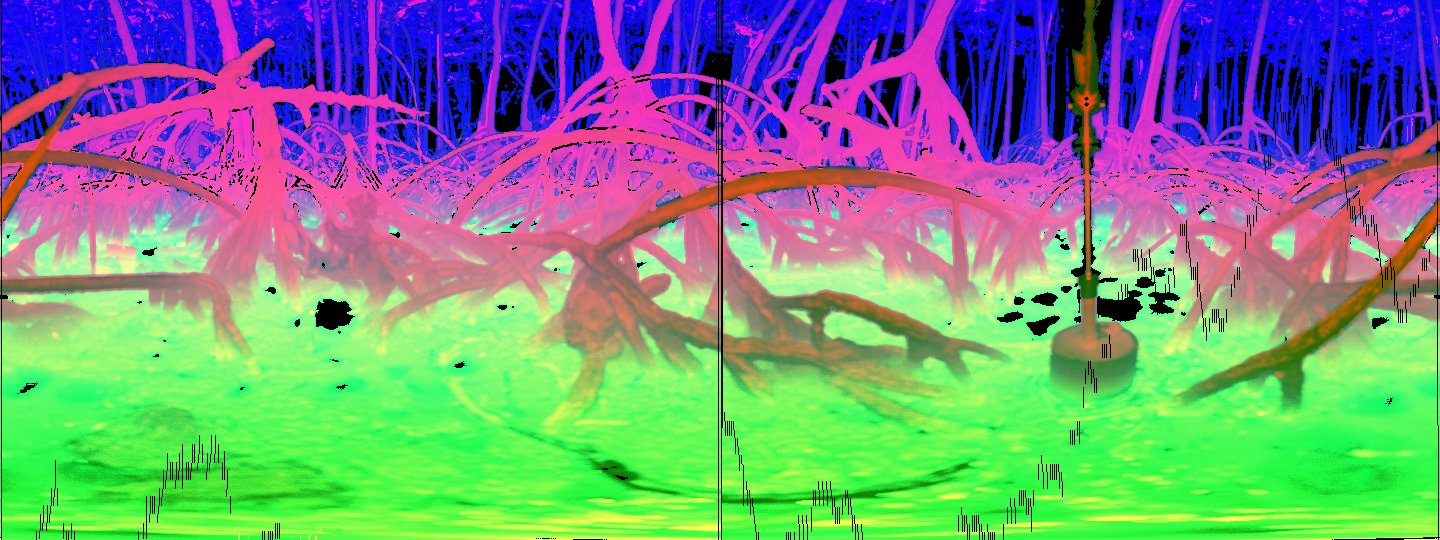}
            \caption*{(b) Stack of Intensity, Z-Inv, and Range}
        \end{subfigure}
        \vspace{0.5em}
        \begin{subfigure}[b]{\textwidth}
            \centering
            \includegraphics[width=\textwidth]{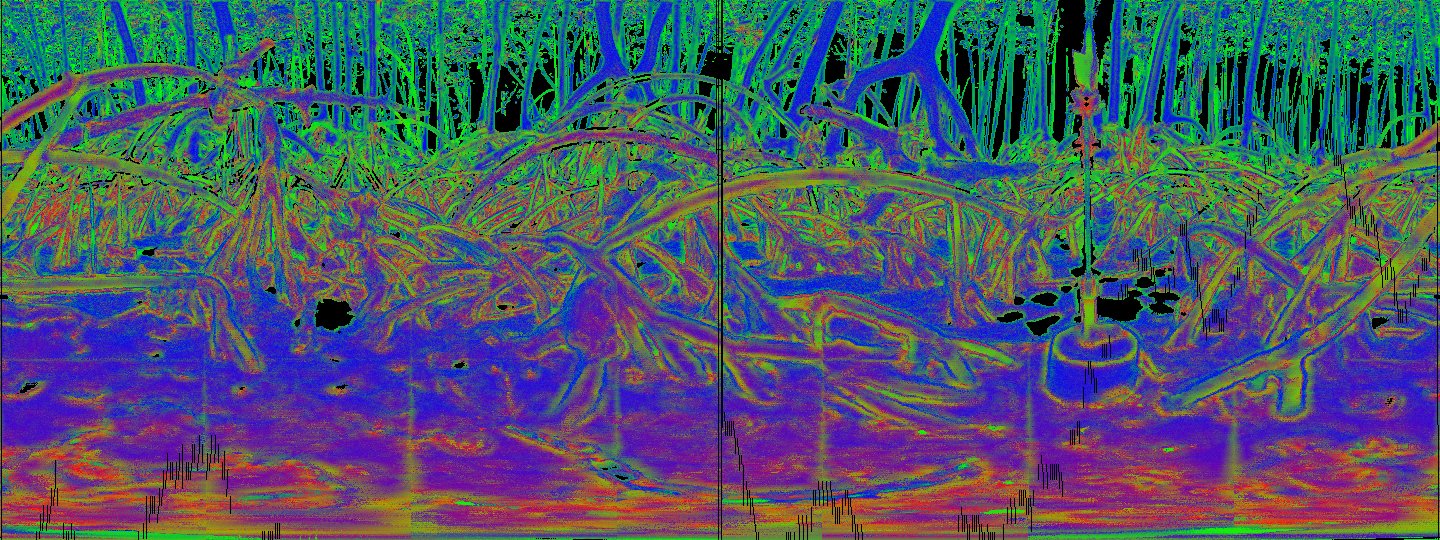}
            \caption*{(c) Stack of curvature, anisotropy, and planarity}
        \end{subfigure}
    \end{minipage}
    \hfill
    \begin{minipage}[b]{0.48\textwidth}
        \begin{subfigure}[b]{\textwidth}
            \centering
            \includegraphics[width=\textwidth]{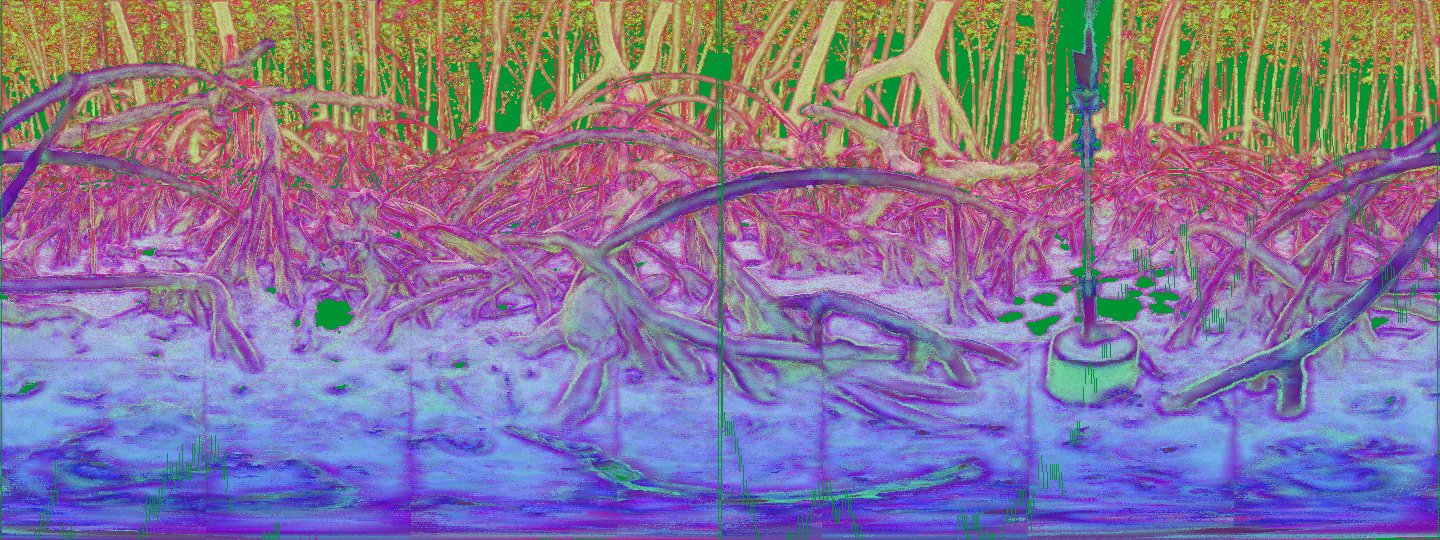}
            \caption*{(d) Stack of the first three components of PCA}
        \end{subfigure}
        \vspace{0.5em}
        \begin{subfigure}[b]{\textwidth}
            \centering
            \includegraphics[width=\textwidth]{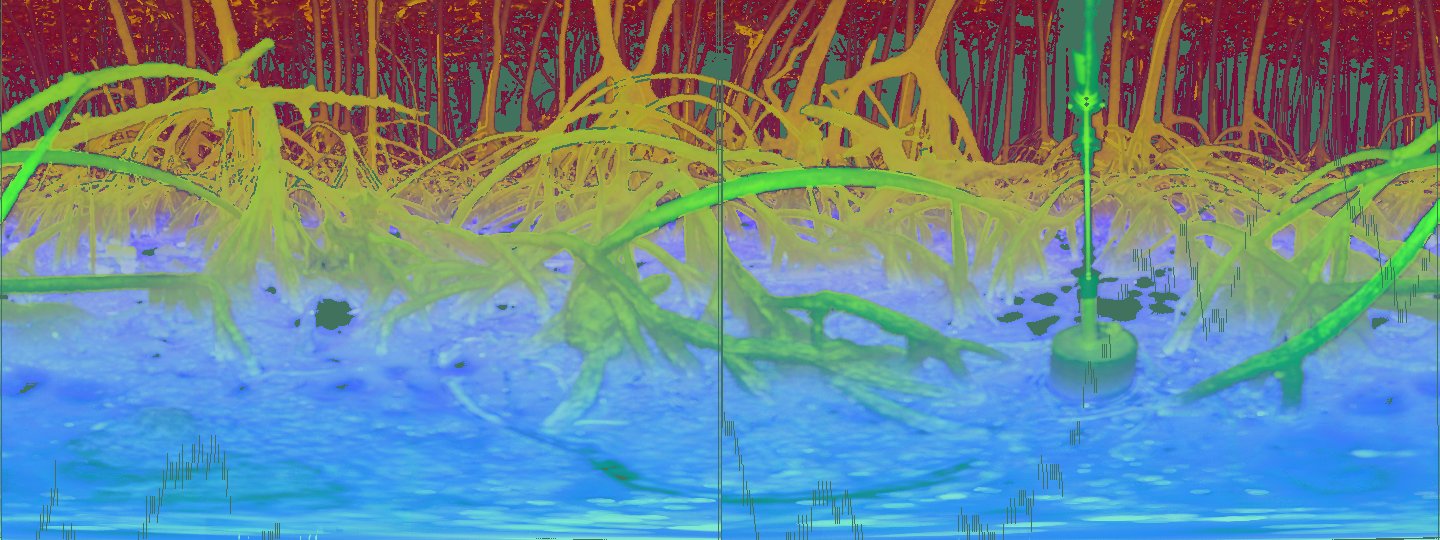}
            \caption*{(e) Stack of the first three components of MNF}
        \end{subfigure}
        \vspace{0.5em}
        \begin{subfigure}[b]{\textwidth}
            \centering
            \includegraphics[width=\textwidth]{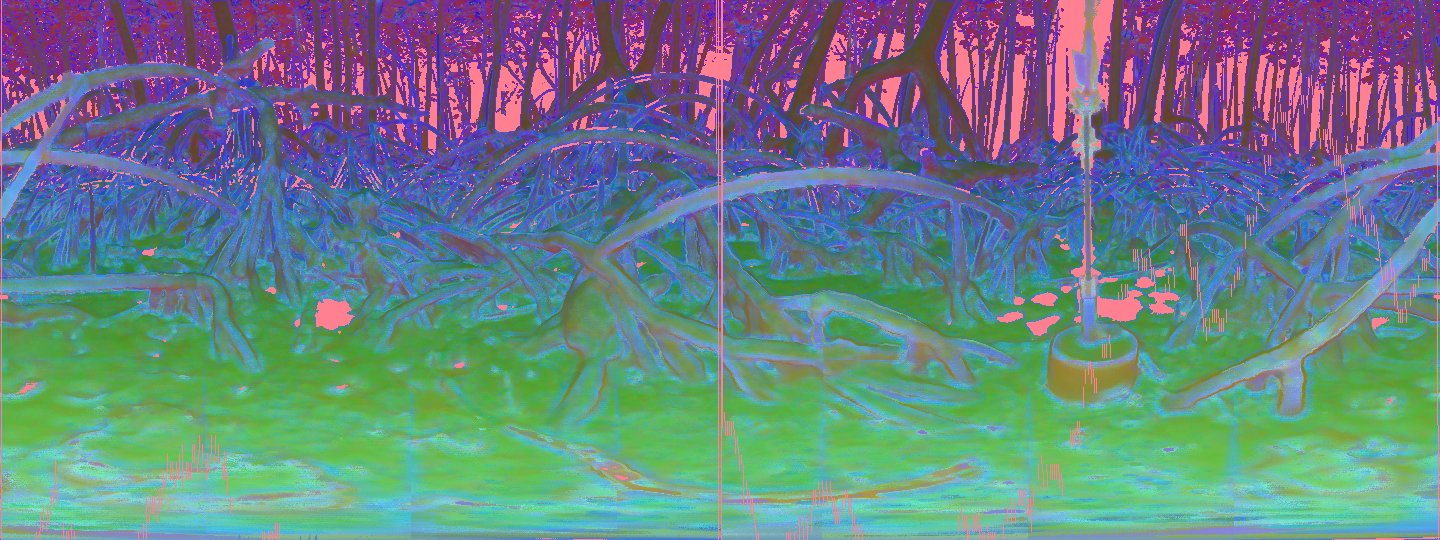}
            \caption*{(f) Stack of the first three components of ICA}
        \end{subfigure}
    \end{minipage}

   \caption{Three-channel feature maps generated from the 2D spherical projection of the terrestrial LiDAR point cloud.}
    \label{fig:feature_maps_stack_mangrove3d}
\end{figure}

%% file: Supple-Mater2.tex
\section{Supplementary Material for Stage 2}
\label{app:stage2}
\subsection{Loss functions}
We trained models using a weighted combination of Dice loss and Cross-Entropy loss:

\begin{equation}\label{eq:loss_combined}
\mathcal{L} = 0.5 \,\mathcal{L}_{\text{Dice}} + 0.5 \,\mathcal{L}_{\text{CrossEntropy}}.
\end{equation}

Dice loss~\cite{milletari2016v,sudre2017generalised}:

\begin{equation}\label{eq:loss_dice}
\mathcal{L}_{\text{Dice}} = 1 - \frac{2 \sum_i p_i g_i}{\sum_i p_i + \sum_i g_i},
\end{equation}

where \(p_i\) and \(g_i\) are the predicted probability and ground-truth label for pixel \(i\).

Cross-Entropy loss~\cite{mao2023cross}:

\begin{equation}\label{eq:loss_ce}
\mathcal{L}_{\text{CrossEntropy}} = - \sum_i \sum_c g_{i,c} \log(p_{i,c}),
\end{equation}

where \(p_{i,c}\) is the predicted probability of class \(c\) at pixel \(i\), and \(g_{i,c}\) is the corresponding ground-truth indicator.

\subsection{Ensemble inference and uncertainty estimation}
For logits \(\mathbf{Z}^{(m)} \in \mathbb{R}^{C \times H \times W}\) from the \(m\)-th model, the ensemble-averaged probability map is:

\begin{equation}\label{eq:ensemble_logits}
\bar{\mathbf{Z}} = \frac{1}{M} \sum_{m=1}^{M} \mathbf{Z}^{(m)}, 
\qquad \bar{\mathbf{P}} = \mathrm{softmax}(\bar{\mathbf{Z}}).
\end{equation}

The predictive entropy of the averaged probability distribution is:

\begin{equation}\label{eq:entropy_predictive}
H[\bar{\mathbf{P}}] = -\sum_{c=1}^{C} \bar{P}_{c,i,j} \log \bar{P}_{c,i,j}.
\end{equation}

The expected entropy across the \(M\) models is:

\begin{equation}\label{eq:entropy_expected}
\mathbb{E}[H[\mathbf{P}]] = \frac{1}{M} \sum_{m=1}^{M}\left(-\sum_{c=1}^{C} P^{(m)}_{c,i,j} \log P^{(m)}_{c,i,j}\right).
\end{equation}

Finally, epistemic uncertainty is estimated as the mutual information~\cite{lakshminarayanan2017simple,kendall2017uncertainties}:

\begin{equation}\label{eq:uncertainty_epistemic}
U_{\text{ep}}(i,j) = H[\bar{\mathbf{P}}] - \mathbb{E}[H[\mathbf{P}]].
\end{equation}

These equations quantify model disagreement at each pixel, providing a principled measure of epistemic uncertainty widely adopted in semantic segmentation.

%% file: Supple-Mater3.tex
\section{Supplementary Material for Stage 3}
\label{app:stage3}
\subsection{Back projection}
Each 3D point $(x,y,z)$ is assigned the label of its corresponding pixel $(i,j)$ from the 2D spherical map:
\begin{equation}\label{eq:backproj_simple}
\ell(x,y,z) = L(i,j).
\end{equation}

In compact form:
\begin{equation}\label{eq:backproj_compact}
\ell(x,y,z) = L\!\Bigl(
   \bigl\lfloor \tfrac{\theta(x,y,z)-\theta_{\min}}{\Delta\theta}\bigr\rfloor,\;
   \bigl\lfloor \tfrac{\phi(x,y,z)-\phi_{\min}}{\Delta\phi}\bigr\rfloor
   \Bigr),
\end{equation}
where $\theta,\phi$ are spherical angles and $(\Delta\theta,\Delta\phi)$ are grid steps.

\subsection{Geometric smoothing}
For a point $i$ at $\mathbf{p}_i$, the $k$ nearest neighbors are
\begin{equation}\label{eq:knn}
\mathcal{N}_k(i) =
   \operatorname*{arg\,top\text{-}k}_{j\neq i}\|\mathbf{p}_i-\mathbf{p}_j\|_2 .
\end{equation}
The label is updated by majority vote:
\begin{equation}\label{eq:knn_vote}
\hat{y}_i = 
   \underset{c\in\{1,\dots,C\}}{\operatorname*{arg\,max}}
   \sum_{j\in\mathcal{N}_k(i)} \mathbf{1}[y_j=c].
\end{equation}

\subsection{Random-Forest relabeling}
A reliable core set is defined as
\begin{equation}\label{eq:rf_core}
\mathcal{R} = \{\,i \mid y_i=\hat{y}_i \;\wedge\; y_i\neq 0 \,\}.
\end{equation}
A balanced Random Forest $f_\theta$ is trained on multiscale features (XYZ, normals, etc.). For suspect points, final labels are:
\begin{equation}\label{eq:rf_update}
y_i^{\text{final}} =
   \begin{cases}
      \arg\max_c p_i(c), & \text{if } \max_c p_i(c)\ge \tau, \\[6pt]
      \hat{y}_i, & \text{otherwise},
   \end{cases}
\end{equation}
with confidence threshold $\tau=0.8$.

\subsection{Virtual spheres}
Virtual spheres re-project 2D feature maps onto synthetic, uniformly sampled spherical grids rather than irregular TLS point distributions. This produces a compact, standardized abstraction of each scan that preserves global structure while dramatically reducing data volume.

At $1^\circ$ angular resolution, for example, an average 800K-point CBL TLS scan compresses to $\sim48K$ points ($\approx1/16$ of the memory) with minimal structural loss. Because the representation is density-neutral, it mitigates distortions from uneven TLS sampling, and resolution can be flexibly tuned to balance fidelity against storage. Importantly, the number of points in a virtual sphere is determined solely by angular resolution and field of view, not by the density of the original point cloud. Thus, while a Semantic3D scan may contain up to 200M points, its virtual sphere at $0.2^\circ$ angular resolution yields only $\sim0.8M$ points—cutting memory demands by a factor of 250 while preserving global structure.

These properties make virtual spheres broadly useful. They enable rapid quality control of scan coverage, occlusion, and segmentation outputs; serve as lightweight metadata summaries for large TLS archives; and support visual benchmarking across sites, campaigns, or ecological conditions. Their clarity and compactness also extend to education and outreach, where visually intuitive overviews of complex forest structures can engage non-expert audiences.

%% file: Supple-Mater4.tex
\section{Animations of the Colorized Point Clouds and Virtual Spheres}
\label{app:animations}
\input{tex_of_fig_tab/tab_animations}

%% file: tex_of_fig_tab/tab_animations.tex
\begin{table}[htbp]
    \centering
    \footnotesize 
    \caption{Figures and corresponding animation links.}
    \label{tab:animations}
    \begin{tabular}{@{}p{6.5cm}p{4cm}@{}} 
        \toprule
        \textbf{Figure} & \textbf{Animation Link} \\
        \midrule
        Figure~\ref{fig:colorized_pcd}: Mangrove3D point clouds & 
        \href{https://drive.google.com/drive/folders/1TzkOQA9D_52S6VbHZMGMOnD5OrqbuIbH?usp=drive_link}{View Animations} \\
        
        Figure~\ref{fig:mangrove_3d_balls}: Mangrove3D virtual spheres & 
        \href{https://drive.google.com/drive/folders/1aE_k9CYtkrEkAnsm6R3XDQ02BiZoJdyr?usp=drive_link}{View Animations} \\
        
        Figure~\ref{fig:forestsemantic_3d_pcd}: ForestSemantic point clouds & 
        \href{https://drive.google.com/drive/folders/1y8sjcAcmwT2QoebL51xwWE_8D8pMql_S?usp=drive_link}{View Animations} \\
        
        Figure~\ref{fig:forestsemantic_balls}: ForestSemantic virtual spheres & 
        \href{https://drive.google.com/drive/folders/1Xc-7xDoTQ7_MY8XLlThHITzy9Oiw10VB?usp=drive_link}{View Animations} \\
        
        Figure~\ref{fig:semantic3d_balls}: Semantic3D virtual spheres & 
        \href{https://drive.google.com/drive/folders/1UAn67j5HrJIK3KQArGLeYRMZ85PPXapD?usp=drive_link}{View Animations} \\
        \bottomrule
    \end{tabular}
\end{table}